\documentclass[letterpaper]{article} % DO NOT CHANGE THIS
\usepackage[preprint]{aaai2027}  % Show authors and omit the AAAI copyright notice.
\usepackage[hyphens]{url}  % DO NOT CHANGE THIS
\usepackage{graphicx} % DO NOT CHANGE THIS
\usepackage{natbib}  % DO NOT CHANGE THIS AND DO NOT ADD ANY OPTIONS TO IT
\usepackage{caption} % DO NOT CHANGE THIS AND DO NOT ADD ANY OPTIONS TO IT
\usepackage{algorithm}
\usepackage{algorithmic}

\usepackage{newfloat}
\usepackage{listings}
\DeclareCaptionStyle{ruled}{labelfont=normalfont,labelsep=colon,strut=off} % DO NOT CHANGE THIS
\floatstyle{ruled}
\newfloat{listing}{tb}{lst}{}
\floatname{listing}{Listing}

\usepackage{booktabs}
\usepackage{multirow}
\usepackage{amsmath}
\usepackage{amssymb}

\title{TurboClear: One-Step Object-Effect Removal via Region-Calibrated Distribution Matching and Fusion}
\author{
    Jiawei Guo\textsuperscript{\rm 1},
    Junxian Li\textsuperscript{\rm 1},
    Yixin Tang\textsuperscript{\rm 1},
    Bingya Zhang\textsuperscript{\rm 2},
    Jiaxin Lu\textsuperscript{\rm 2},\\
    Yulun Zhang\textsuperscript{\rm 1,\normalfont\textdagger},
    Shangchen Zhou\textsuperscript{\rm 3,\normalfont\textdagger}
}
\affiliations{
    \textsuperscript{\rm 1}Shanghai Jiao Tong University\\
    \textsuperscript{\rm 2}Honor Device Co., Ltd\\
    \textsuperscript{\rm 3}Imperial College London\\
    \textsuperscript{\normalfont\textdagger}Corresponding authors: Yulun Zhang, yulun100@gmail.com;
    Shangchen Zhou, shangchenzhou@gmail.com
}

\begin{document}

\maketitle

\begin{abstract}
Recently, diffusion-based removal methods have achieved promising visual quality in removing both target objects and their associated effects. However, they typically rely on multi-step denoising, leading to high inference cost. Directly applying existing one-step distillation methods is also suboptimal, since their global objectives lack explicit region-wise calibration and may weaken the asymmetric edit-and-preserve behavior required by object-effect removal. To address these challenges, we propose TurboClear, a one-step SDXL-based object-effect removal model. During training, we design Region-Calibrated Distribution Matching (RDM) for region-aware distillation to preserve the teacher model's asymmetric edit-and-preserve behavior. Furthermore, we propose Learnable Spatial Fusion (LSF) for lightweight inference-time fusion. Extensive experiments show that TurboClear significantly improves inference efficiency while maintaining competitive visual quality. TurboClear reduces the computational overhead by up to $40.04\times$ compared to ObjectClear, and by up to $665\times$ against the Flux-based method OmniPaint, all while maintaining comparable or better visual removal quality. Code is available at \url{https://github.com/GuoCalix/TurboClear}.
\end{abstract}


\section{Introduction}

Object removal represents a highly specialized and uniquely challenging task within the broader domain of image inpainting and image editing~\cite{meng2021sdedit, yu2021wavefill}. It aims to erase unwanted objects from an image as if they had never appeared, while reconstructing plausible background content. In realistic scenarios, however, the target object often leaves associated visual effects, such as shadows, reflections, and occlusion traces, which may extend beyond the object mask. Therefore, object-effect removal requires not only generating missing content in the affected region, but also preserving the irrelevant background.

This requirement is inherently spatially asymmetric: regions affected by the target object should undergo substantial semantic and structural changes, whereas unaffected regions should remain nearly identity-mapped. Traditional mask-confined inpainting methods~\cite{ekin2024clipaway, rombach2022high, yi2023instinpaint, zhuang2024task, sun2025attentive, li2025rorem} restrict generation within the given mask, requiring users to provide masks that cover all affected pixels. Recent diffusion-based object removal methods~\cite{wei2025omnieraser, winter2024objectdrop, zhu2025georemover, yu2025omnipaint} relax this constraint and improve visual quality, but many of them do not explicitly distinguish object-affected regions from unaffected regions, which may lead to background changes or residual object effects.

\begin{figure}[t]
    \centering
    \includegraphics[width=0.92\columnwidth]{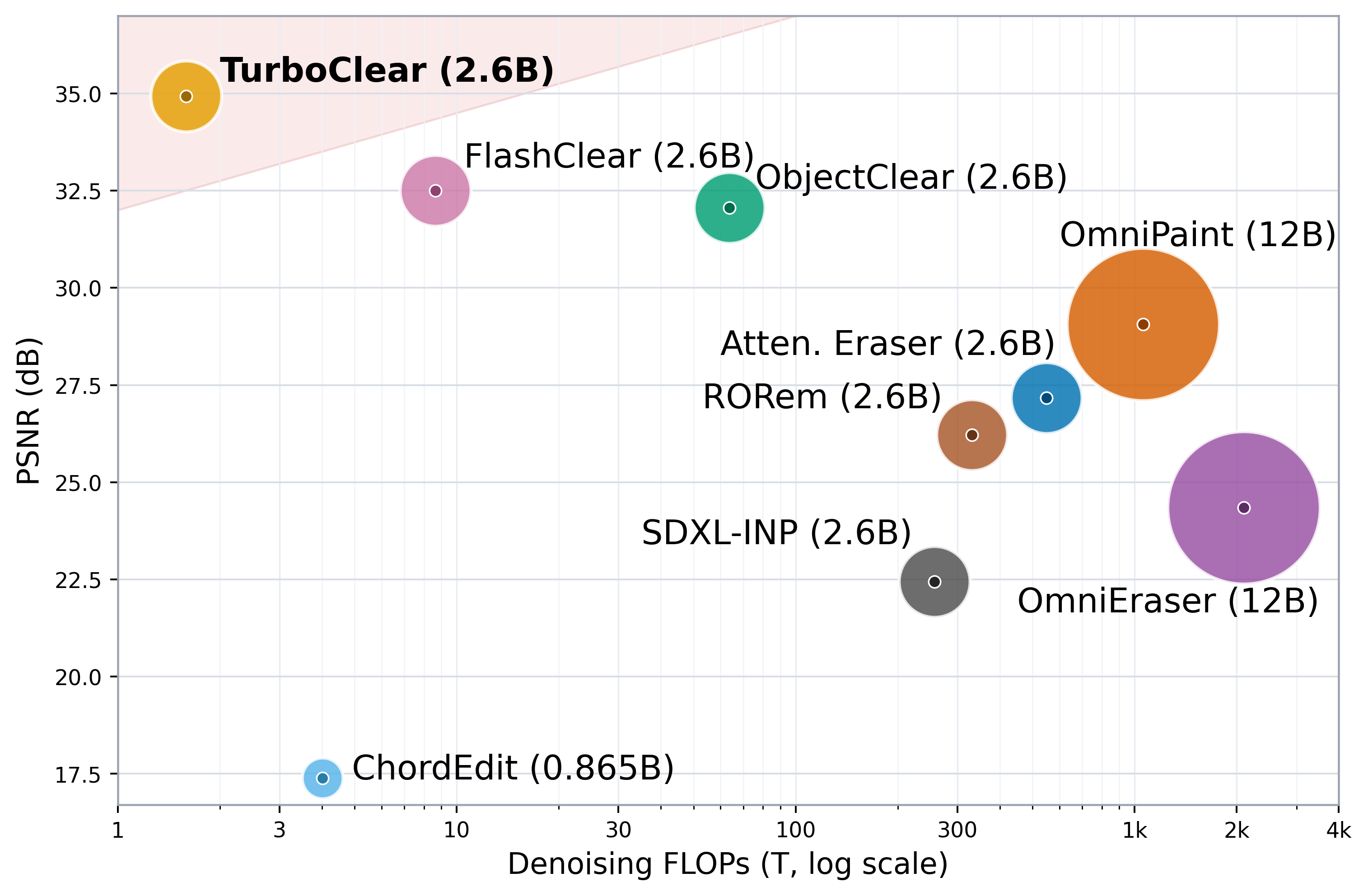}
    \caption{PSNR-Params-FLOPs comparison of mask-  and text-based editing methods. The vertical axis is PSNR (dB), the horizontal axis is FLOPs (computational cost), and bubble size denotes parameter count (memory cost). TurboClear outperforms prior methods with substantially lower FLOPs.}
    \label{fig:psnr_flops_params}
\end{figure}

More advanced and dedicated methods such as ObjectClear~\cite{zhao2025objectclear} encourage such asymmetric behavior through specialized supervision on cross-attention layers and achieve promising removal quality. However, they still inherit the multi-step sampling process of diffusion models. Even with relatively lightweight SDXL-based architectures~\cite{podell2023sdxl}, iterative inference remains computationally expensive, limiting real-time deployment on both servers and edge devices.

A natural solution is to distill multi-step object removal models into a one-step student. However, existing acceleration or one-step distillation methods, such as Consistency Model~\cite{song2023consistency, luo2023latent, lu2024simplifying}, DMD~\cite{yin2024one}, and DMD2~\cite{yin2024improved}, are not specifically designed for object-effect removal. Their objectives lack explicit region-wise calibration, and thus may weaken the spatial asymmetry required by the task, causing either residual effects in edited regions or unnecessary changes in preserved regions.

To address this challenge, we propose TurboClear, a one-step SDXL-based object-effect removal model. During training, we introduce Region-Calibrated Distribution Matching (RDM), which uses object-effect region masks to assign different matching targets to different spatial regions: affected regions are matched toward the generative teacher distribution, while unaffected regions are calibrated toward the ground-truth preservation target. This enables stable one-step distillation while maintaining the asymmetric edit-preserve behavior.

Furthermore, to structurally decouple generation and preservation during inference, we introduce Learnable Spatial Fusion (LSF). Instead of relying on a single prediction stream, TurboClear maintains two asymmetric streams, a removal-oriented generative stream and an identity-preserving reference stream, and learns spatial gates to fuse them. This design preserves background content while effectively removing both the target object and its associated effects. As shown in Fig.~\ref{fig:psnr_flops_params}, our method successfully reduces the computational burden while keeping the remove quality. Our contributions can be summarized as follows:
\begin{itemize}
	\item We propose TurboClear, a one-step SDXL-based object-effect removal model that addresses the efficiency bottleneck of diffusion-based image erasing.
	\item We propose Region-Calibrated Distribution Matching, a distillation objective tailored for object-effect removal, which preserves the spatially asymmetric edit-and-preserve behavior during one-step distillation.
	\item We design Learnable Spatial Fusion, a lightweight inference-time fusion module that adaptively combines removal-oriented and preservation-oriented predictions with negligible computational overhead.
\end{itemize}

% Object removal represents a highly specialized and uniquely challenging task within the broader domain of image inpainting, characterized by an inherent and extreme spatial asymmetry. Unlike general generative tasks, it strictly requires the pristine preservation of the background, while simultaneously demanding the precise localization and generation of physically reasonable textures to replace the user-specified object and its associated visual effects (e.g., shadows and reflections). However, most diffusion-based models~\cite{meng2021sdedit, podell2023sdxl, rombach2022high, corneanu2024latentpaint, liu2024structure, yu2021wavefill, xie2023smartbrush, ekin2024clipaway, zhuang2024task, sun2025attentive, li2025rorem} fine-tuned directly from general generative priors struggle to handle this strict asymmetry. They frequently fail to completely eliminate residual visual effects or synthesize plausible, clean backgrounds in the target areas, often hallucinating unneeded semantic elements. While a few recent specialized multi-step diffusion models~\cite{wei2025omnieraser, winter2024objectdrop, zhu2025georemover, zhao2025objectclear} have largely overcome these quality issues and achieved remarkable visual success, their reliance on iterative sampling introduces prohibitive computational overhead. This inefficiency severely hinders their deployment on large-scale central servers or edge devices, making real-time interactive response largely unattainable.

\section{Related Work}

\textbf{Object Removal} aims to seamlessly eliminate a user-specified object from an image based on an input mask. While diffusion-based methods currently dominate this task, traditional approaches~\cite{ekin2024clipaway, rombach2022high, yi2023instinpaint, zhuang2024task, sun2025attentive, li2025rorem} rely on precise hard masks to strictly dictate the editing region. This imposes rigorous requirements on mask quality and fails to address residual effects—such as shadows or reflections—that extend beyond the mask boundaries. To overcome this, recent methods relax the mask constraints during generation~\cite{suvorov2022resolution,lugmayr2022repaint,jiang2025smarteraser,liu2025erase} or incorporate textual guidance~\cite{nichol2021glide,saharia2022palette,avrahami2022blended,brooks2023instructpix2pix,kawar2023imagic}. However, this brings a new challenge: the model's spatial edit-and-preserve asymmetry becomes difficult to maintain perfectly, which manifests as unintended background alterations and object artifacts. While some recent methods~\cite{zhang2023adding,ju2024brushnet,chen2024anydoor,manukyan2023hd} attempt to address this via plug-in modules, dedicated methods~\cite{zhao2025objectclear,zhu2025georemover,wei2025omnieraser} explicitly incorporate spatial asymmetric constraints into the training phase, either by directly supervising internal network representations or by introducing customized spatial guidance. These methods have substantially improved the removal quality. Nevertheless, all the aforementioned methods are bottlenecked by the multi-step diffusion architecture, introducing substantial computational overhead that limits their industrial applicability.

\noindent\textbf{Step Distillation.} To accelerate diffusion models, existing strategies primarily bifurcate into training-free solvers~\cite{lu2022dpm,zhao2023unipc,liu2022pseudo, kulikov2025flowedit} and training-based step distillation~\cite{ meng2023distillation, zheng2024trajectory, yan2024perflow, ren2024hyper, luhman2021knowledge, heek2024multistep, xu2024accelerating, zhou2024score, gu2023boot, nguyen2023swiftbrush}. In the training-free and text-guided paradigm, recent advancements~\cite{lu2026chordedit} have impressively pushed the boundary to single-step editing. However, without tailored structural constraints, they often struggle to perfectly maintain complex background layouts. On the training-based front, early techniques such as Progressive Distillation~\cite{salimans2022progressive} and Consistency Models~\cite{song2023consistency,luo2023latent,lu2024simplifying} successfully compress the iterative sampling trajectory. To further enhance perceptual quality, Adversarial Distillation~\cite{sauer2024adversarial, lin2024sdxl, xu2024ufogen} has emerged as a dominant paradigm. Dedicated methods~\cite{tang2026flash} tailor adversarial distillation specifically for object removal, achieving high-fidelity results in just four steps. Yet, due to the inherent instability of adversarial objectives, such methods notoriously struggle to converge at the extreme single-step regime. Distribution Matching Distillation~\cite{yin2024one, yin2024improved} introduces advanced distribution matching formulations, enabling robust single-step generation. However, as generic synthesis frameworks, they inherently lack the spatial asymmetric constraints essential for localized edit-and-preserve tasks. Motivated by these insights, we propose Region-Calibrated Distribution Matching (RDM), which adds explicit asymmetric spatial constraints into distribution matching.

\section{Method}

\begin{figure*}[t]
    \centering
    \includegraphics[width=\textwidth]{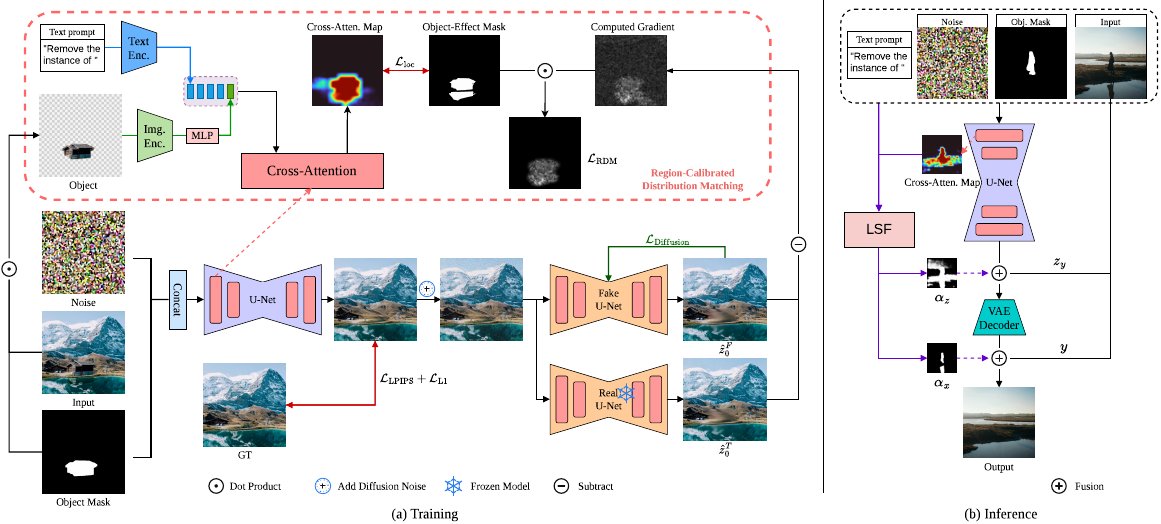}
    \caption{Overview of TurboClear. (a) RDM spatially calibrates the one-step distribution-matching gradient using the object-effect mask, while paired reconstruction and localization objectives stabilize training. The fake UNet is jointly optimized using a standard diffusion loss under a fixed fake-to-student update ratio. (b) With a single UNet inference, LSF predicts latent- and pixel-space gates, $\alpha_z$ and $\alpha_x$, to fuse the removal-oriented prediction with the identity-preserving input stream. The object-effect mask is used only during training; inference requires the input image, object mask, and text prompt.}
    \label{fig:pipeline}
\end{figure*}

\subsection{Preliminary and Overall Architecture}

As illustrated in Fig.~\ref{fig:pipeline}, TurboClear couples region-calibrated one-step distillation with dual-stream spatial fusion. We first formulate one-step object-effect removal in the latent diffusion space. Given an input image $y\in\mathbb{R}^{3\times H\times W}$, an object mask $m_o\in\{0,1\}^{1\times H\times W}$, a text condition $c$, and a paired clean target $x^\star\in\mathbb{R}^{3\times H\times W}$, the goal is to predict an output image that removes both the target object and its visual effects while preserving irrelevant background content~\cite{rombach2022high}. During training, we additionally use an object-effect mask $m_e\in\{0,1\}^{1\times H\times W}$, which covers the target object and its associated effects such as shadows or reflections. Importantly, $m_e$ is only used as privileged supervision during training, while inference uses the deployable object mask $m_o$.

Let $E:\mathbb{R}^{3\times H\times W}\rightarrow\mathbb{R}^{C\times h\times w}$ and $D:\mathbb{R}^{C\times h\times w}\rightarrow\mathbb{R}^{3\times H\times W}$ denote the VAE encoder and decoder~\cite{kingma2013auto}. We define $z^\star=E(x^\star)\in\mathbb{R}^{C\times h\times w}$ and $z_y=E(y)\in\mathbb{R}^{C\times h\times w}$. A latent diffusion model perturbs a clean latent $z_0\in\mathbb{R}^{C\times h\times w}$ by
\begin{equation}
z_t = \sqrt{\bar{\alpha}_t} z_0 + \sqrt{1-\bar{\alpha}_t}\epsilon,
\quad \epsilon \sim \mathcal{N}(0,I),
\end{equation}
where $\bar{\alpha}_t$ is the cumulative noise schedule~\cite{ho2020denoising, song2020denoising}. A denoiser predicts either noise or velocity and can be converted into a clean latent estimate $\hat{z}_0$. In multi-step object removal, a teacher denoiser iteratively applies this process conditioned on $(y,m_o,c)$, which yields high-quality removal but incurs large inference cost.

TurboClear distills such a teacher into a one-step student. Starting from a Gaussian latent $z_T\in\mathbb{R}^{C\times h\times w}$, $z_T\sim\mathcal{N}(0,I)$, the student predicts a clean latent $\hat{z}_0\in\mathbb{R}^{C\times h\times w}$ in a single UNet evaluation,
\begin{equation}
\hat{z}_0 = G_\theta(z_T,t_s,y,m_o,c),
\quad \hat{x}=D(\hat{z}_0),
\end{equation}
where $t_s$ is the one-step sampling timestep. Generic one-step distillation may not be directly well suited to object-effect removal, as such tasks require a non-negligible portion of the spatial content to remain unchanged rather than be regenerated. Therefore, TurboClear addresses this asymmetric distillation problem from two complementary perspectives. First, Region-Calibrated Distribution Matching (RDM) calibrates the distribution matching signal with the object-effect region during student training. Second, Learnable Spatial Fusion (LSF) builds a dual-stream inference architecture~\cite{li2019selective} that adaptively combines a removal-oriented generative stream and an identity-preserving reference stream.

\subsection{Region-Calibrated Distribution Matching}

Generic distribution matching distillation aligns the student distribution with a teacher distribution by contrasting a real teacher score and a fake score model fitted to the current student samples~\cite{yin2024one,yin2024improved}. However, for object-effect removal, applying this signal over the entire image can encourage unnecessary background changes. RDM instead treats distribution matching as a region-calibrated local generation prior.

Given the student prediction $\hat{z}_0\in\mathbb{R}^{C\times h\times w}$, we sample a diffusion timestep $t$ and noise it again as
\begin{equation}
z_t = \sqrt{\bar{\alpha}_t}\hat{z}_0 + \sqrt{1-\bar{\alpha}_t}\epsilon.
\end{equation}
Let $F_T$ be the frozen multi-step teacher denoiser and $F_\psi$ be a fake score denoiser trained on current student samples. Their clean latent predictions $\hat{z}_0^T,\hat{z}_0^F\in\mathbb{R}^{C\times h\times w}$ are
\begin{equation}
\hat{z}_0^T = F_T(z_t,t,y,m_o,c), \quad
\hat{z}_0^F = F_\psi(z_t,t,y,m_o,c).
\end{equation}
The distribution matching direction is estimated by the difference between the teacher-induced residual and the fake-score residual,
\begin{equation}
g_{\rm DMD}
= (\hat{z}_0-\hat{z}_0^T)-(\hat{z}_0-\hat{z}_0^F)
= \hat{z}_0^F-\hat{z}_0^T.
\end{equation}
This signal moves the student sample toward the teacher distribution while correcting for the current generated distribution. In our task, the key issue is not whether such a direction is useful, but where it should be applied.

We downsample $m_e$ to the latent resolution and optionally smooth it into a soft region mask $m_e^\ell\in[0,1]^{1\times h\times w}$. RDM masks and normalizes the distribution matching direction $g_{\rm DMD}\in\mathbb{R}^{C\times h\times w}$ as
\begin{equation}
s = \frac{\|m_e^\ell\odot(\hat{z}_0-\hat{z}_0^T)\|_1}
{C\|m_e^\ell\|_1+\varepsilon},
\end{equation}
\begin{equation}
g_{\rm RDM} = \frac{m_e^\ell\odot g_{\rm DMD}}{s+\varepsilon},
\end{equation}
where $g_{\rm RDM}\in\mathbb{R}^{C\times h\times w}$ and $m_e^\ell$ is broadcast along the channel dimension. The normalization makes the magnitude comparable across samples with different effect-region sizes, while the mask prevents the distribution matching signal from directly optimizing already clean background areas.

Following the stop-gradient formulation of distribution matching, we define a local target
\begin{equation}
\tilde{z}_0 = {\rm sg}(\hat{z}_0 - g_{\rm RDM}),
\end{equation}
and optimize
\begin{equation}
\mathcal{L}_{\rm RDM}
= \frac{1}{2\left(C\|m_e^\ell\|_1+\varepsilon\right)}
\|\sqrt{m_e^\ell}\odot(\hat z_0-\tilde z_0)\|_2^2.
\end{equation}
During joint training, the fake score denoiser (fake UNet) is optimized on noised student samples using the standard diffusion loss under a fixed fake-to-student update ratio, so that $F_\psi$ continuously tracks the evolving student distribution. To stabilize the one-step student before distribution matching, we first warm it up with a perceptual paired objective. The final student objective combines region-calibrated distribution matching with paired reconstruction and localization terms:
\begin{equation}
\begin{split}
\mathcal{L}_{\rm student}
=&\lambda_{\rm rdm}\mathcal{L}_{\rm RDM}
+\lambda_{\rm fg}\|m_e\odot(\hat{x}-x^\star)\|_1\\
&+\lambda_{\rm bg}\|(1-m_e)\odot(\hat{x}-x^\star)\|_1\\
&+\lambda_{\rm p}{\rm LPIPS}(\hat{x},x^\star)
+\lambda_{\rm loc}\mathcal{L}_{\rm loc}.
\end{split}
\end{equation}
Here $\mathcal{L}_{\rm loc}$ encourages the object-token cross-attention to concentrate on the object-effect region, preserving the teacher's edit-and-preserve behavior~\cite{zhao2025objectclear}. Thus, RDM uses the teacher distribution primarily as a completion prior in affected regions, while paired losses keep the unaffected background anchored to the clean target.

\newcommand{\lsflabels}{%
\makebox[0.19\columnwidth][c]{\scriptsize Input w/ Mask} &
\makebox[0.19\columnwidth][c]{\scriptsize Unfused Output} &
\makebox[0.19\columnwidth][c]{\scriptsize Unfused Diff.} &
\makebox[0.19\columnwidth][c]{\scriptsize LSF Output} &
\makebox[0.19\columnwidth][c]{\scriptsize LSF Diff.} \\
}

\begin{figure}[t]
\centering
\setlength{\tabcolsep}{0pt}
\renewcommand{\arraystretch}{1.0}
\begin{tabular}{ccccc}
\includegraphics[width=0.19\columnwidth]{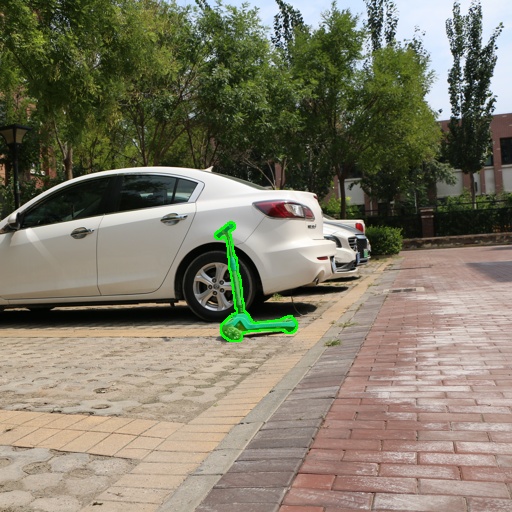} &
\includegraphics[width=0.19\columnwidth]{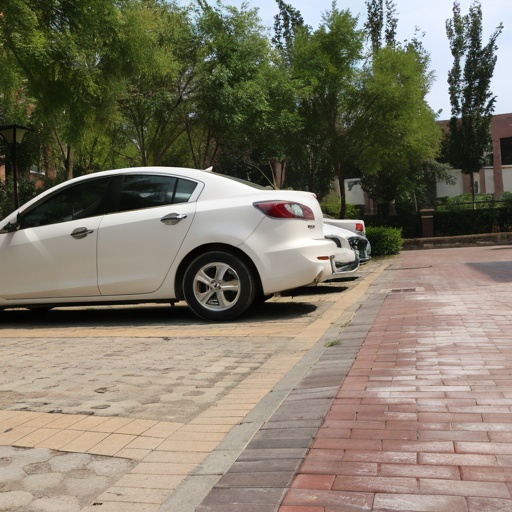} &
\includegraphics[width=0.19\columnwidth]{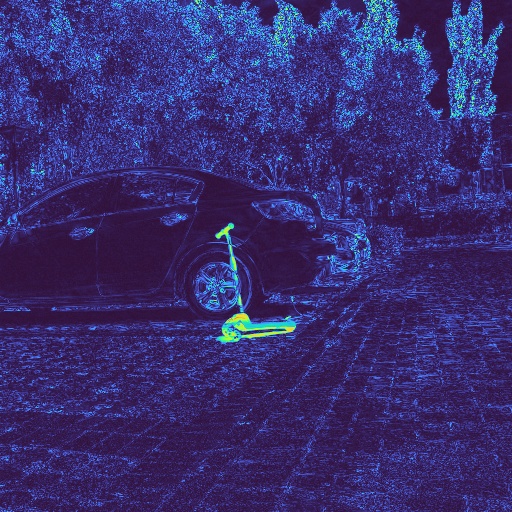} &
\includegraphics[width=0.19\columnwidth]{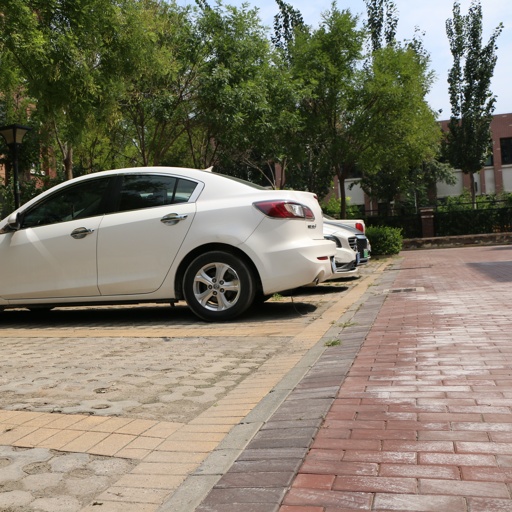} &
\includegraphics[width=0.19\columnwidth]{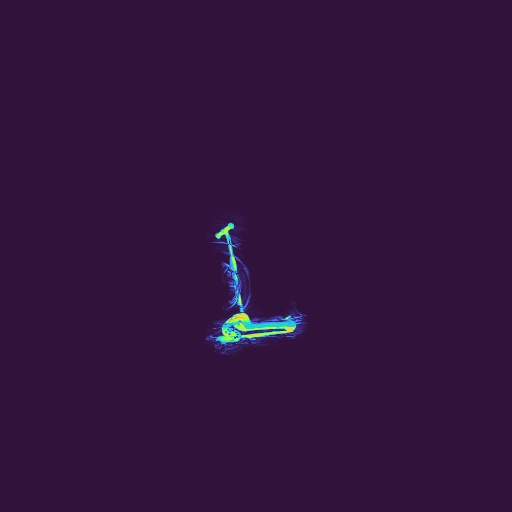} \\
\includegraphics[width=0.19\columnwidth]{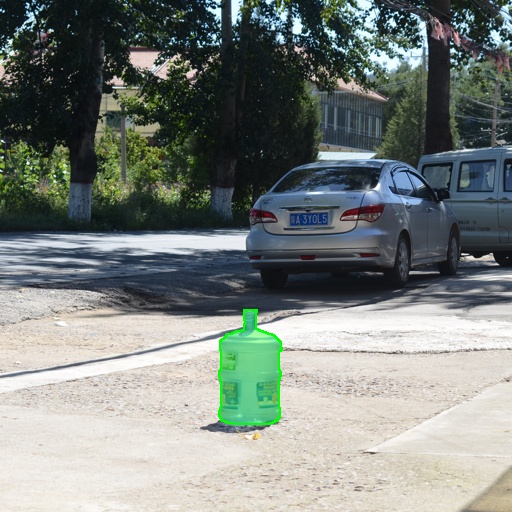} &
\includegraphics[width=0.19\columnwidth]{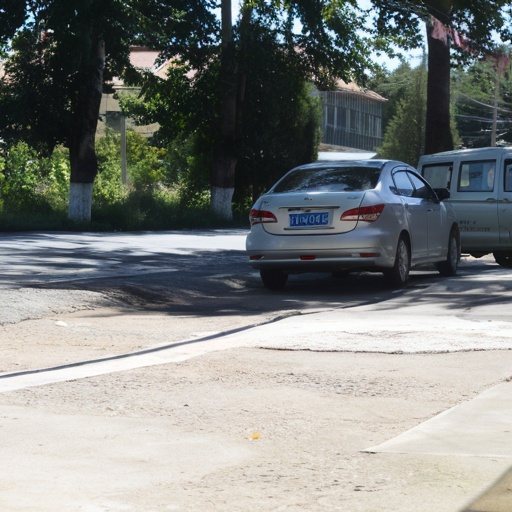} &
\includegraphics[width=0.19\columnwidth]{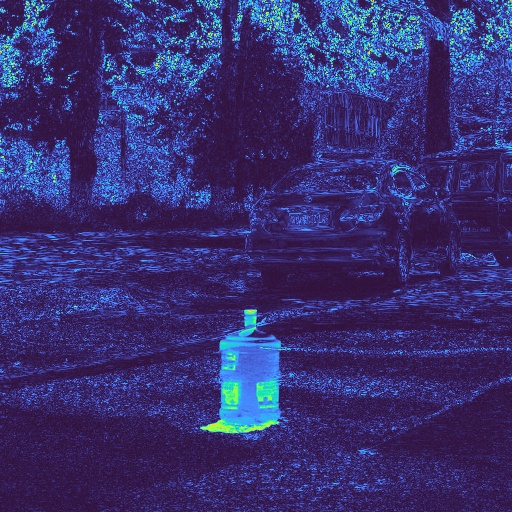} &
\includegraphics[width=0.19\columnwidth]{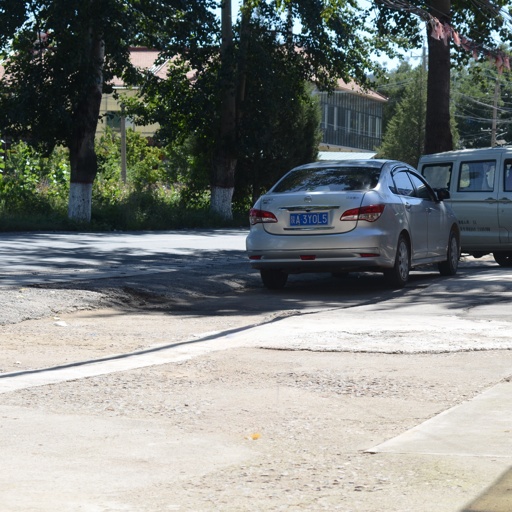} &
\includegraphics[width=0.19\columnwidth]{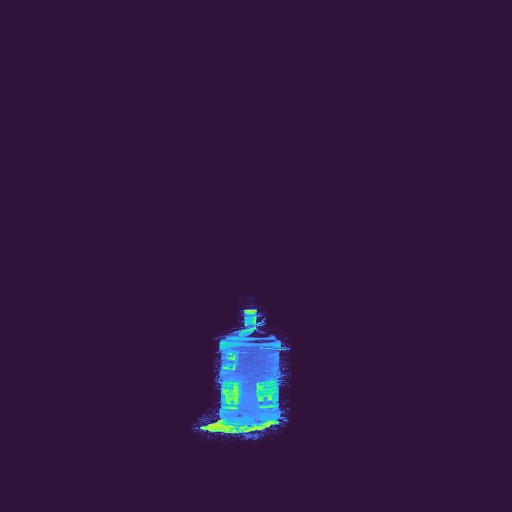} \\
\includegraphics[width=0.19\columnwidth]{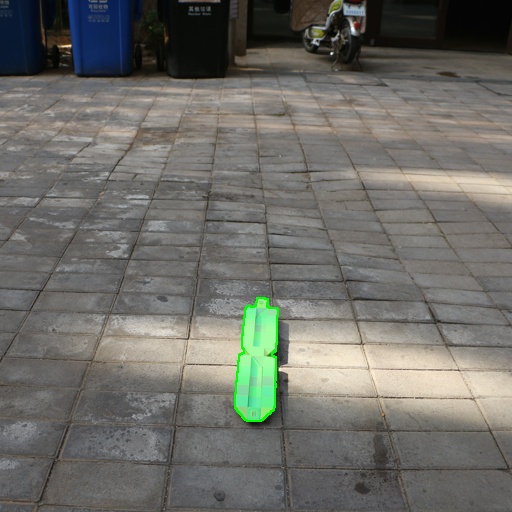} &
\includegraphics[width=0.19\columnwidth]{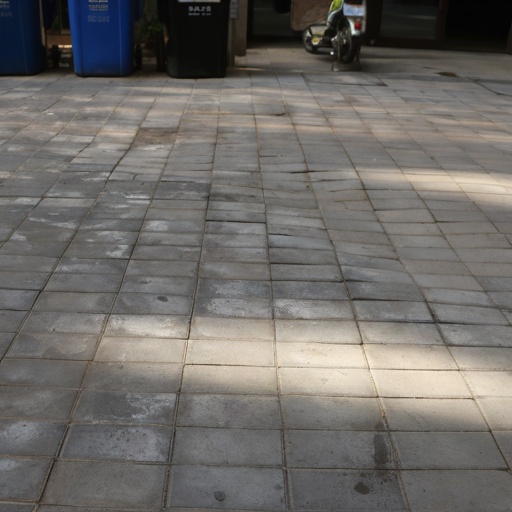} &
\includegraphics[width=0.19\columnwidth]{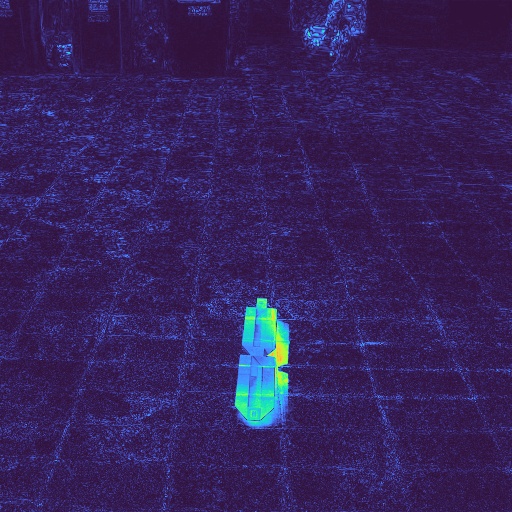} &
\includegraphics[width=0.19\columnwidth]{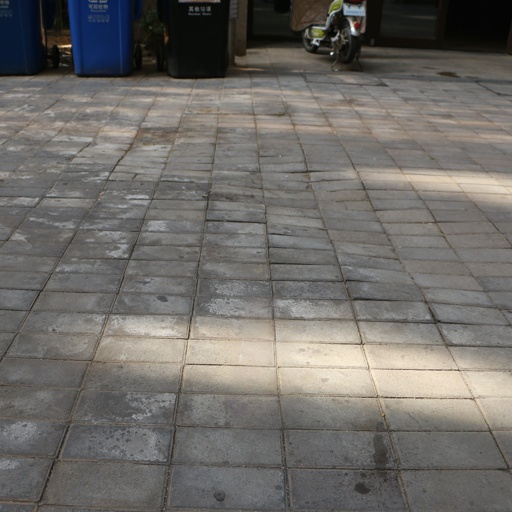} &
\includegraphics[width=0.19\columnwidth]{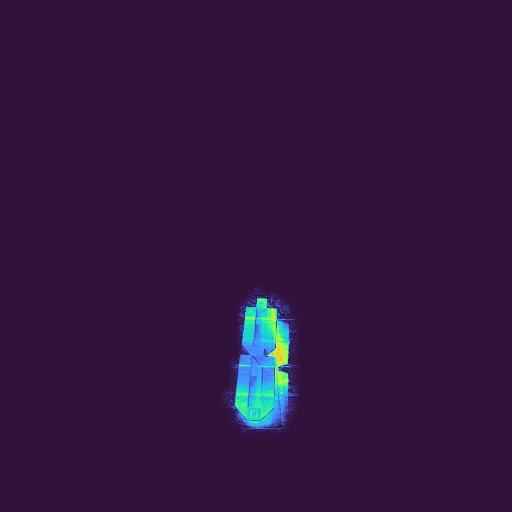} \\
\includegraphics[width=0.19\columnwidth]{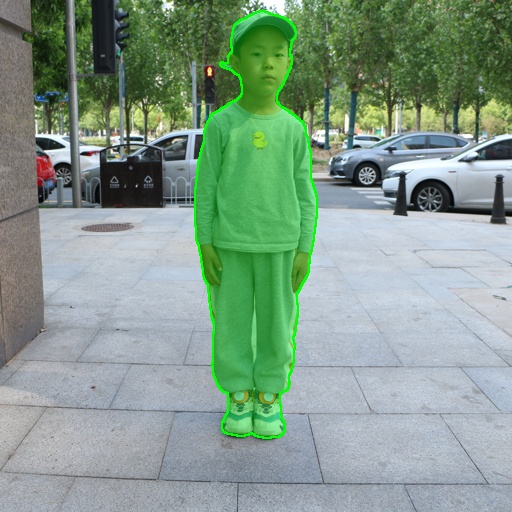} &
\includegraphics[width=0.19\columnwidth]{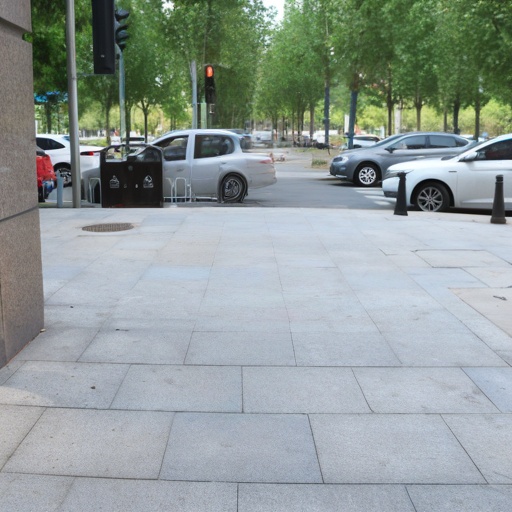} &
\includegraphics[width=0.19\columnwidth]{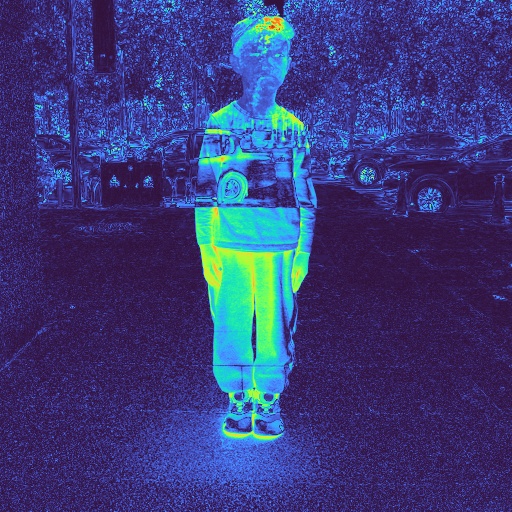} &
\includegraphics[width=0.19\columnwidth]{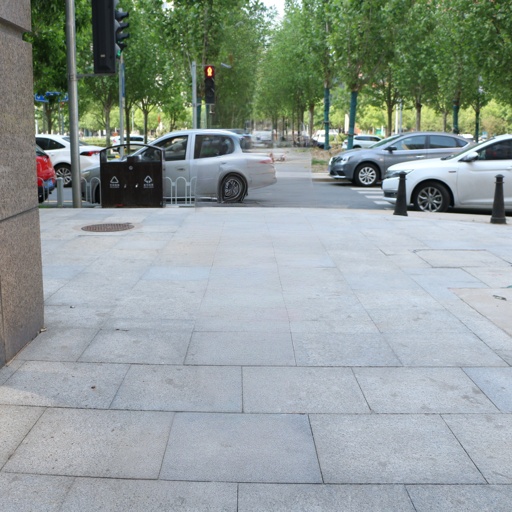} &
\includegraphics[width=0.19\columnwidth]{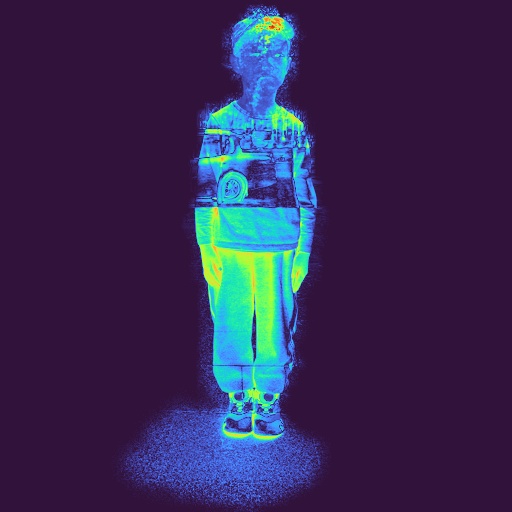} \\
\includegraphics[width=0.19\columnwidth]{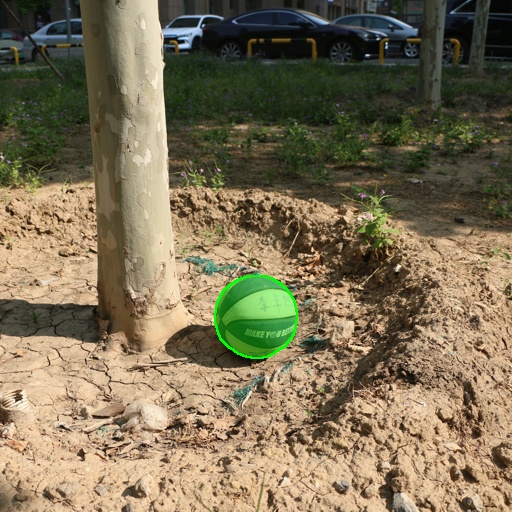} &
\includegraphics[width=0.19\columnwidth]{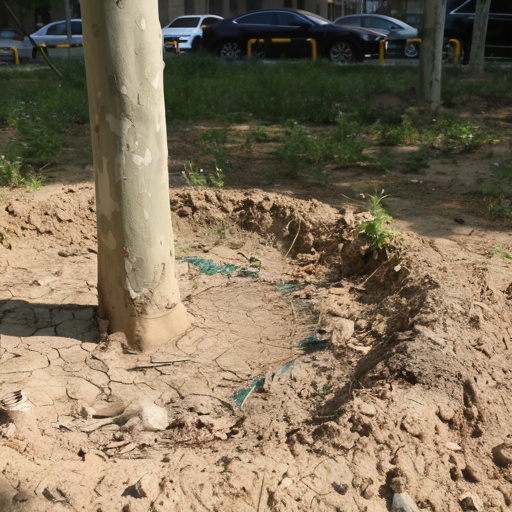} &
\includegraphics[width=0.19\columnwidth]{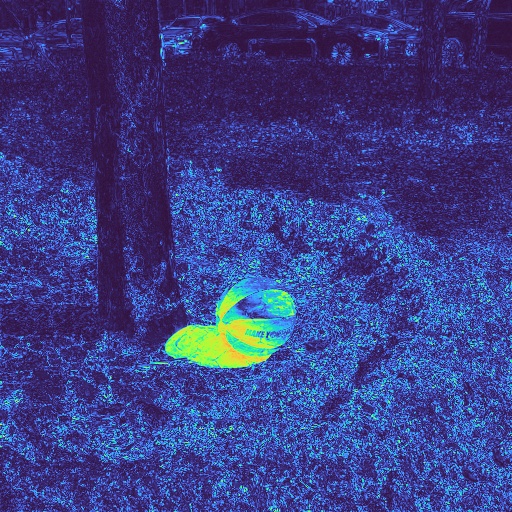} &
\includegraphics[width=0.19\columnwidth]{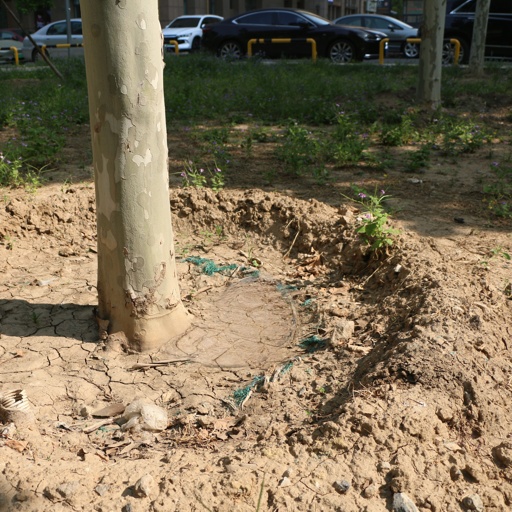} &
\includegraphics[width=0.19\columnwidth]{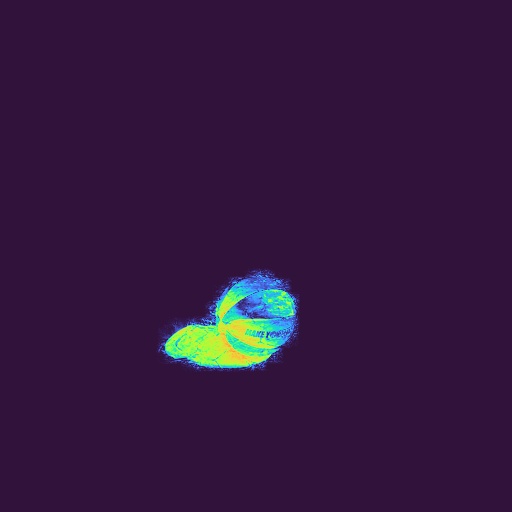} \\
\includegraphics[width=0.19\columnwidth]{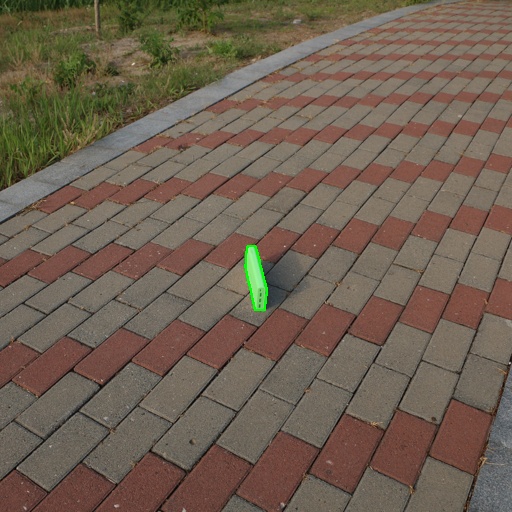} &
\includegraphics[width=0.19\columnwidth]{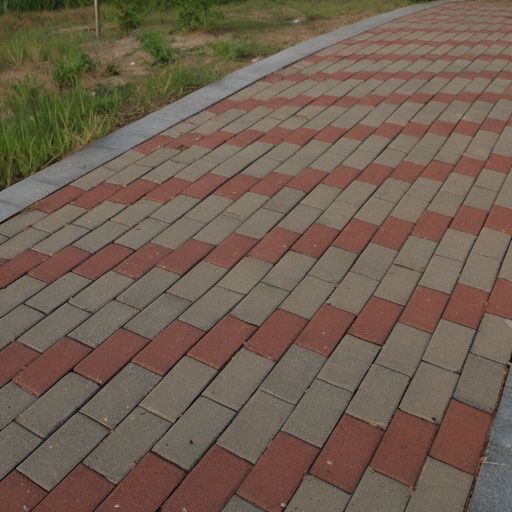} &
\includegraphics[width=0.19\columnwidth]{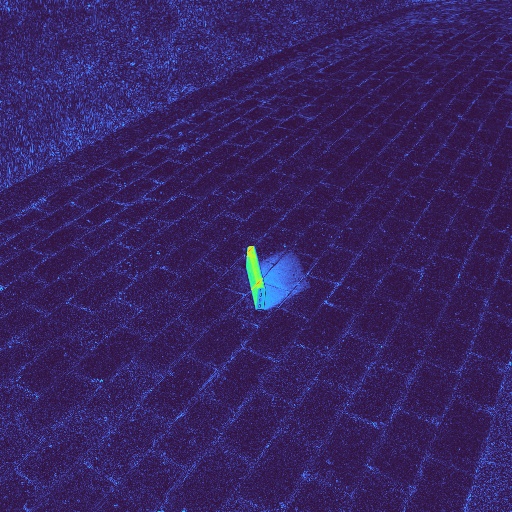} &
\includegraphics[width=0.19\columnwidth]{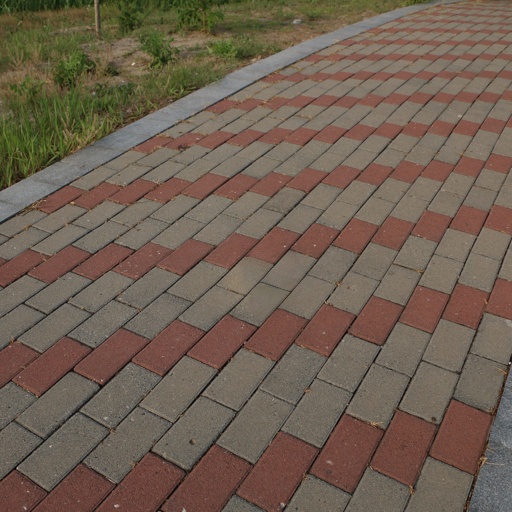} &
\includegraphics[width=0.19\columnwidth]{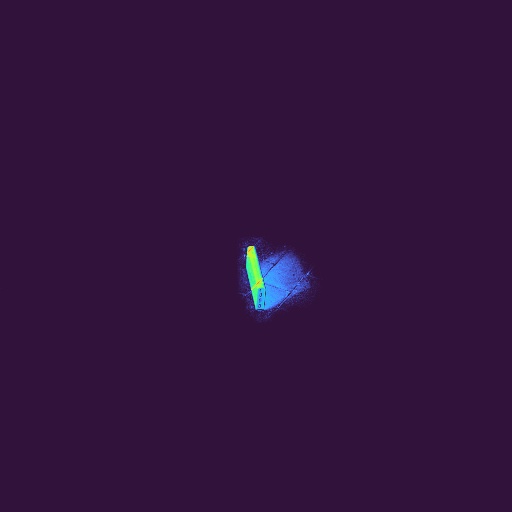} \\
\lsflabels
\end{tabular}
\caption{Visualizing the spatial asymmetry enabled by LSF. The difference maps show the per-pixel mean absolute RGB difference from the input, rendered with the same color scale. Compared with the unfused stream, LSF confines changes more tightly to the object-effect region and preserves the unaffected background. Zoom in to see more details.}
\label{fig:lsf_spatial_asymmetry}
\end{figure}

\newcommand{\quallabels}{%
\makebox[0.089\textwidth][c]{\scriptsize Input w/ mask} &
\makebox[0.089\textwidth][c]{\scriptsize PowerPaint} &
\makebox[0.089\textwidth][c]{\scriptsize DesignEdit} &
\makebox[0.089\textwidth][c]{\scriptsize CLIPAway} &
\makebox[0.089\textwidth][c]{\scriptsize OmniEraser} &
\makebox[0.089\textwidth][c]{\scriptsize\shortstack{Atten. Eraser}} &
\makebox[0.089\textwidth][c]{\scriptsize RORem} &
\makebox[0.089\textwidth][c]{\scriptsize OmniPaint} &
\makebox[0.089\textwidth][c]{\scriptsize ObjectClear} &
\makebox[0.089\textwidth][c]{\scriptsize FlashClear} &
\makebox[0.089\textwidth][c]{\scriptsize\shortstack{TurboClear (ours)}} \\
}

\begin{figure*}[t]
\centering
\setlength{\tabcolsep}{0pt}
\renewcommand{\arraystretch}{1.0}
\begin{tabular}{ccccccccccc}
\includegraphics[width=0.089\textwidth]{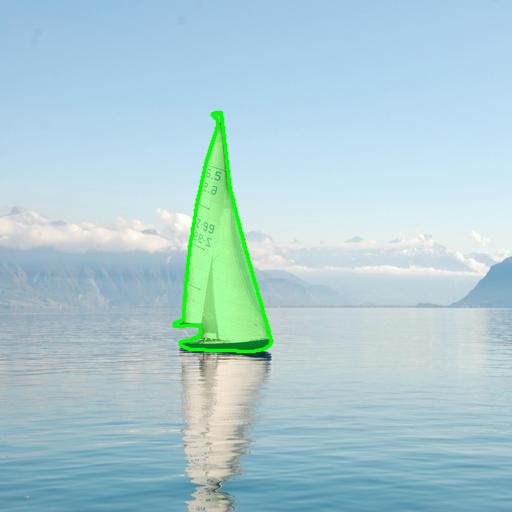} &
\includegraphics[width=0.089\textwidth]{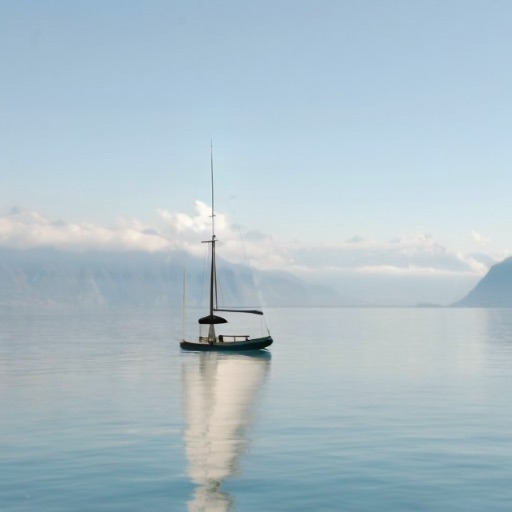} &
\includegraphics[width=0.089\textwidth]{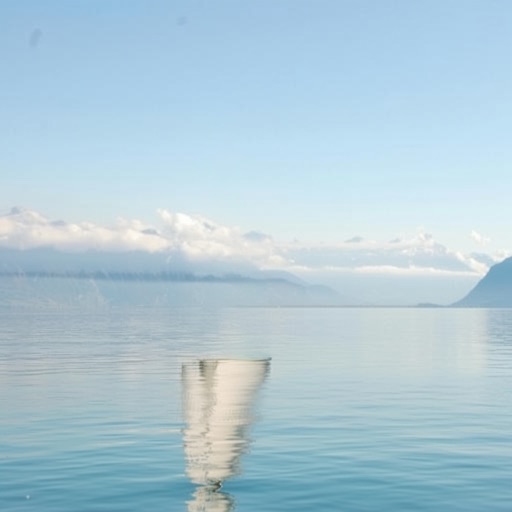} &
\includegraphics[width=0.089\textwidth]{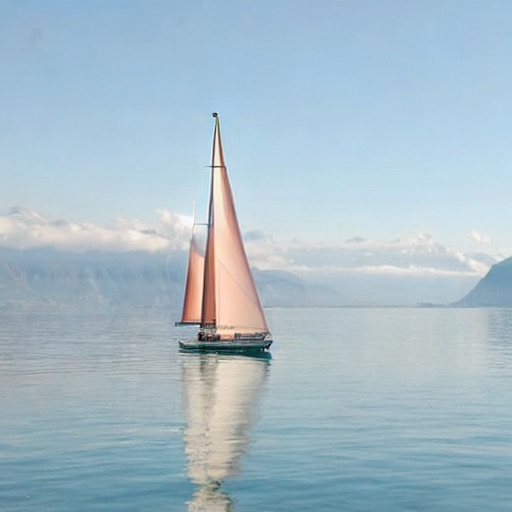} &
\includegraphics[width=0.089\textwidth]{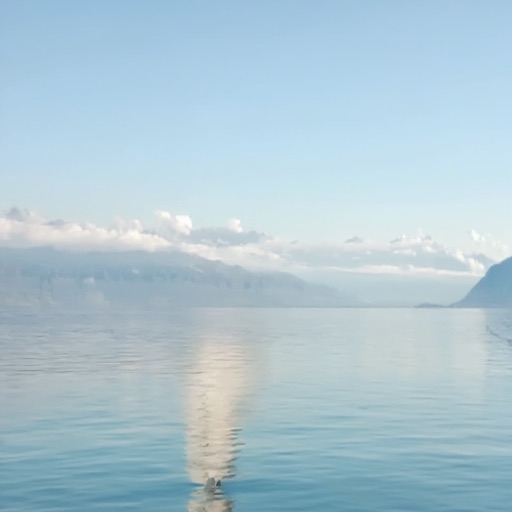} &
\includegraphics[width=0.089\textwidth]{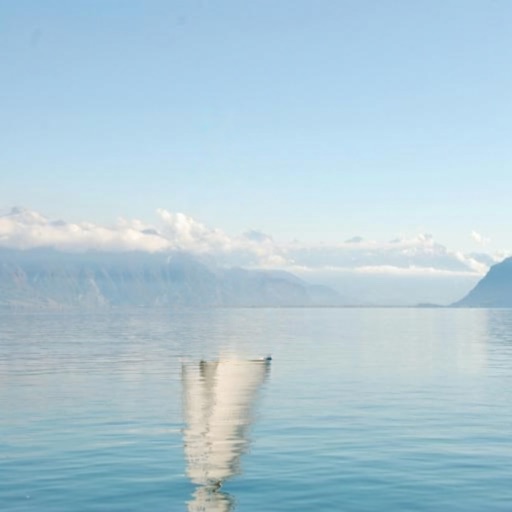} &
\includegraphics[width=0.089\textwidth]{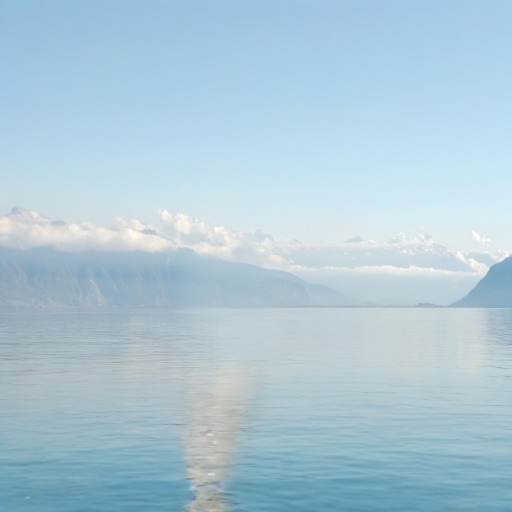} &
\includegraphics[width=0.089\textwidth]{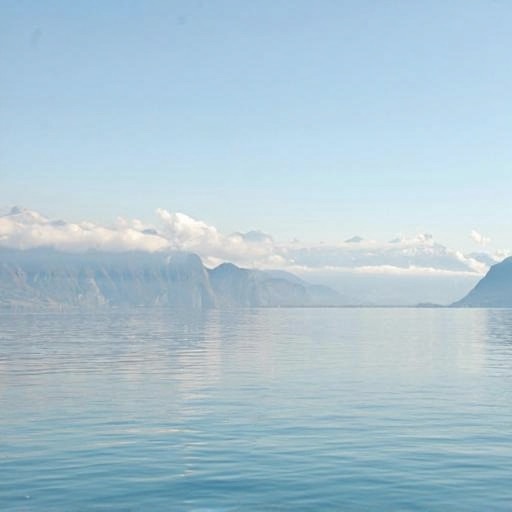} &
\includegraphics[width=0.089\textwidth]{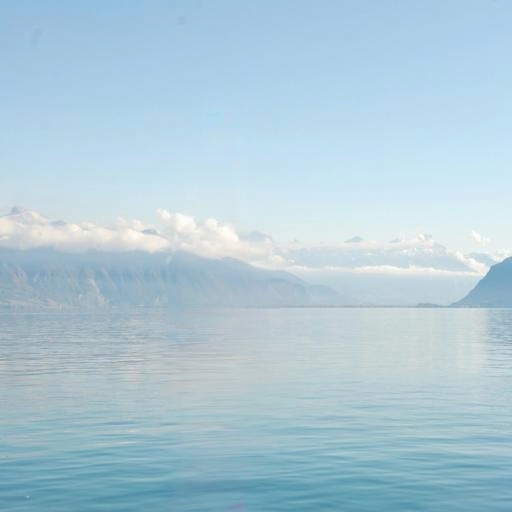} &
\includegraphics[width=0.089\textwidth]{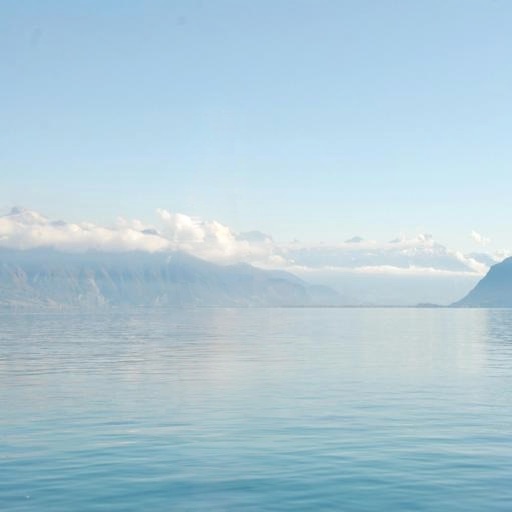} &
\includegraphics[width=0.089\textwidth]{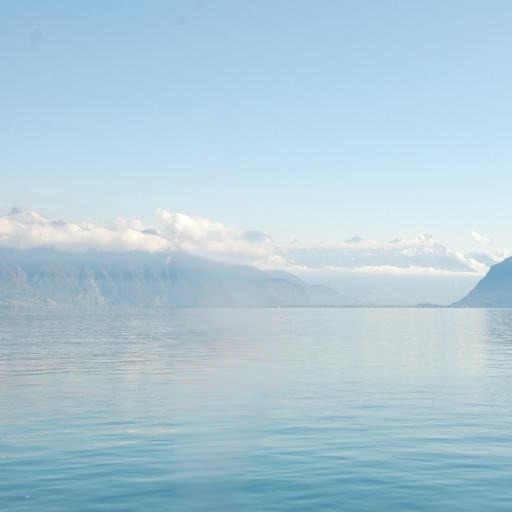} \\
\includegraphics[width=0.089\textwidth]{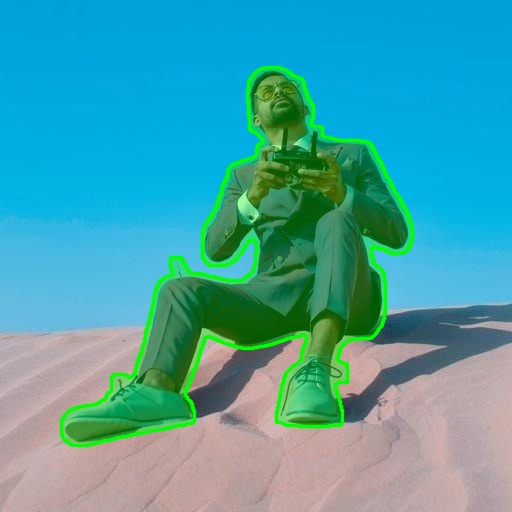} &
\includegraphics[width=0.089\textwidth]{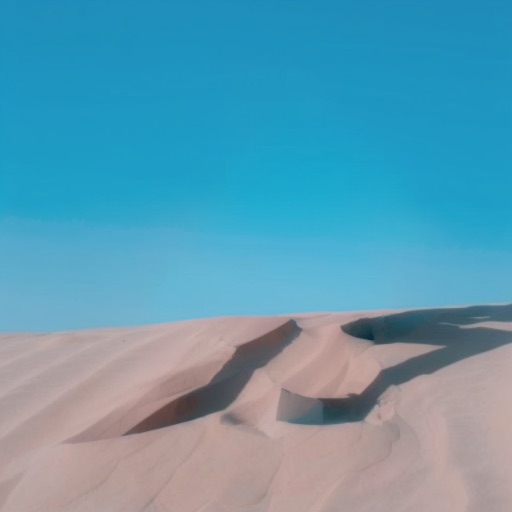} &
\includegraphics[width=0.089\textwidth]{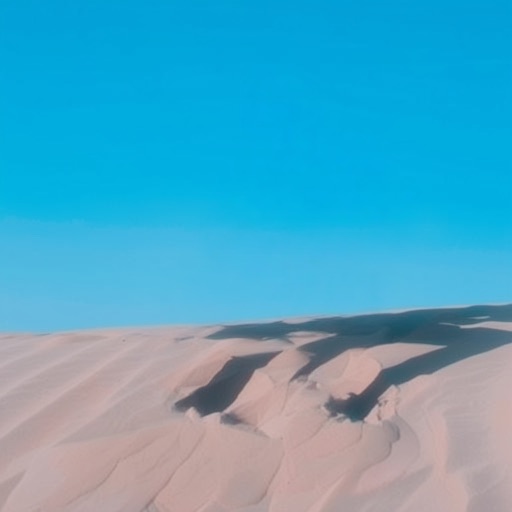} &
\includegraphics[width=0.089\textwidth]{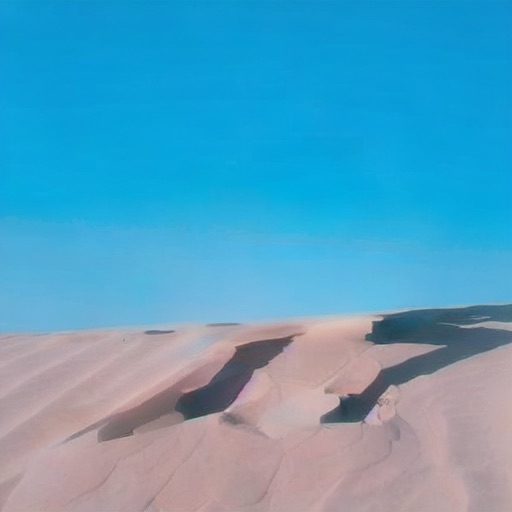} &
\includegraphics[width=0.089\textwidth]{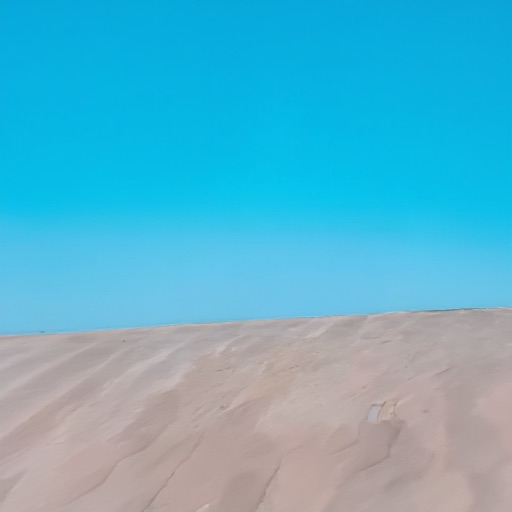} &
\includegraphics[width=0.089\textwidth]{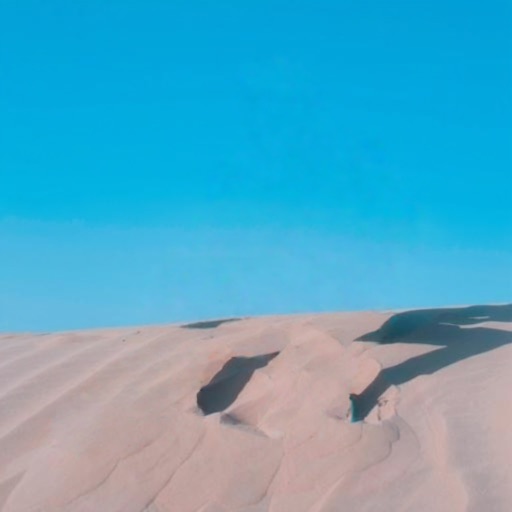} &
\includegraphics[width=0.089\textwidth]{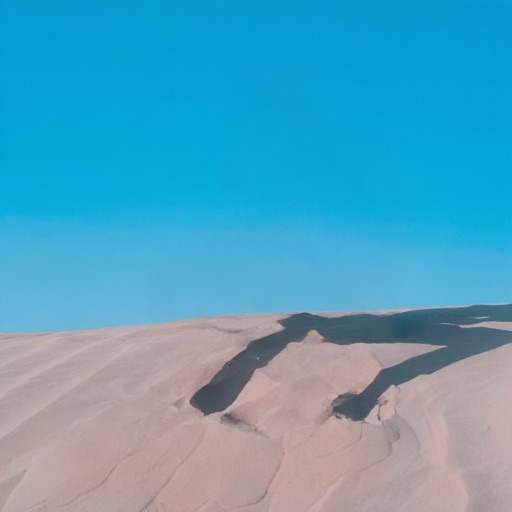} &
\includegraphics[width=0.089\textwidth]{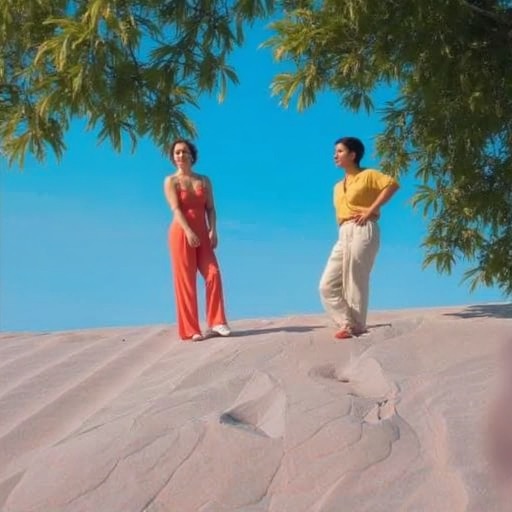} &
\includegraphics[width=0.089\textwidth]{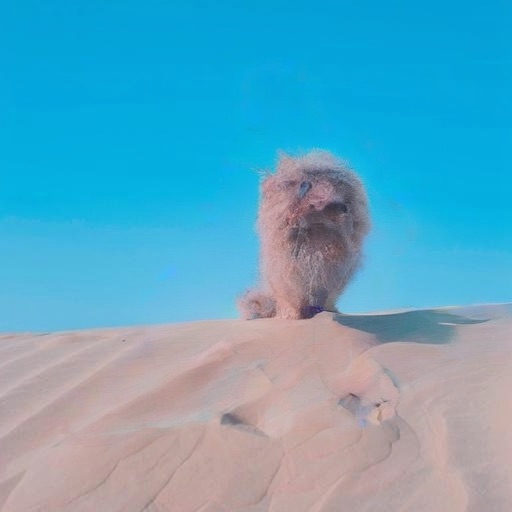} &
\includegraphics[width=0.089\textwidth]{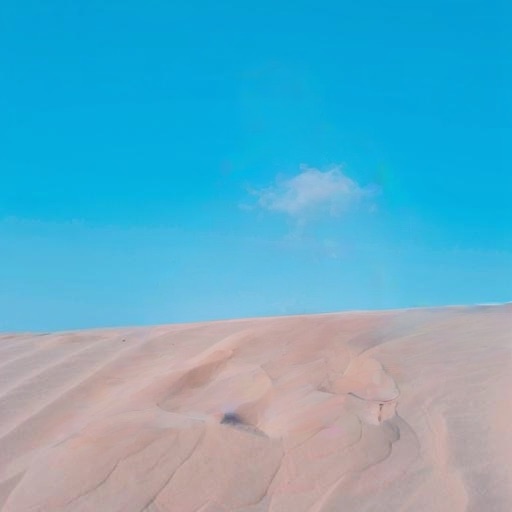} &
\includegraphics[width=0.089\textwidth]{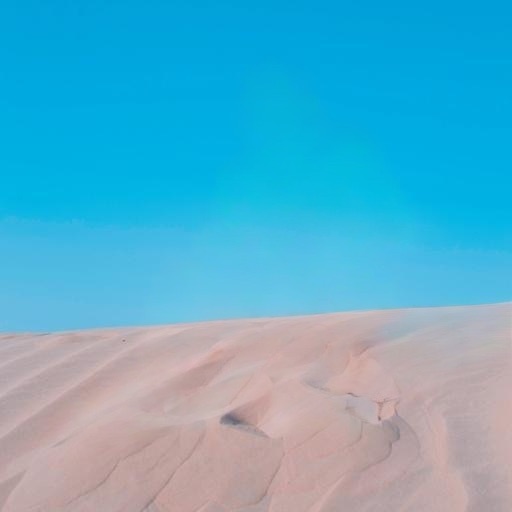} \\
\includegraphics[width=0.089\textwidth]{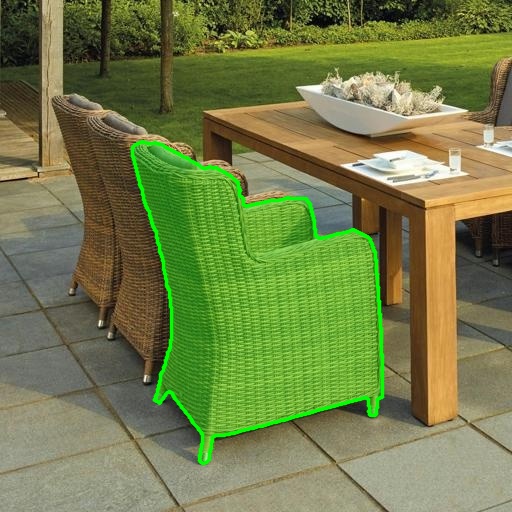} &
\includegraphics[width=0.089\textwidth]{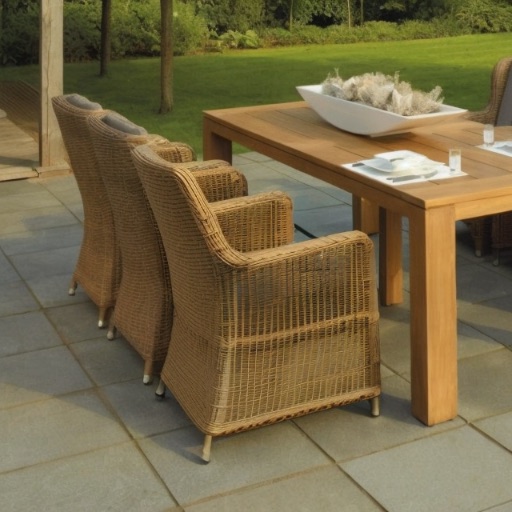} &
\includegraphics[width=0.089\textwidth]{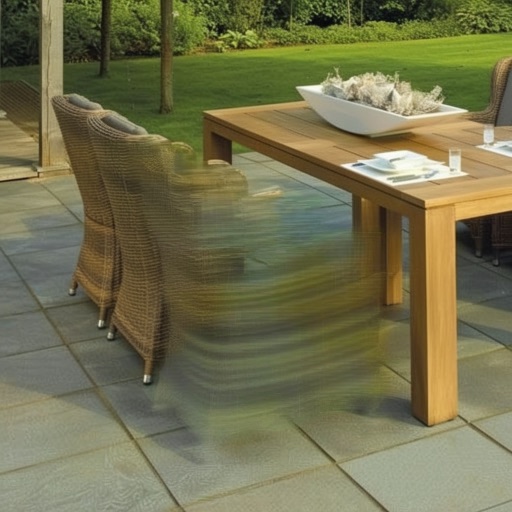} &
\includegraphics[width=0.089\textwidth]{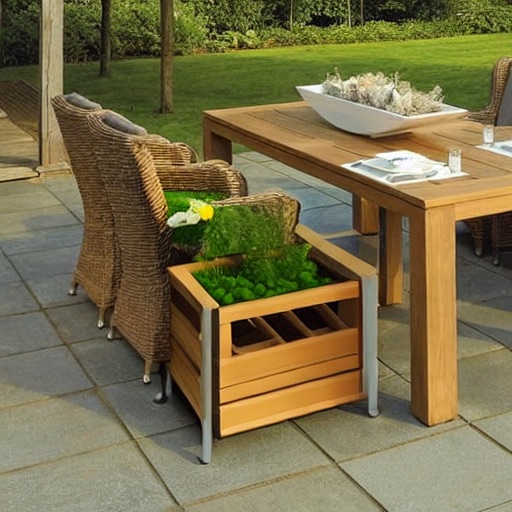} &
\includegraphics[width=0.089\textwidth]{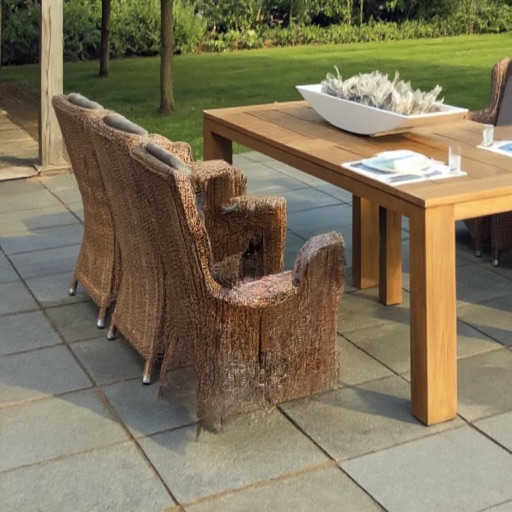} &
\includegraphics[width=0.089\textwidth]{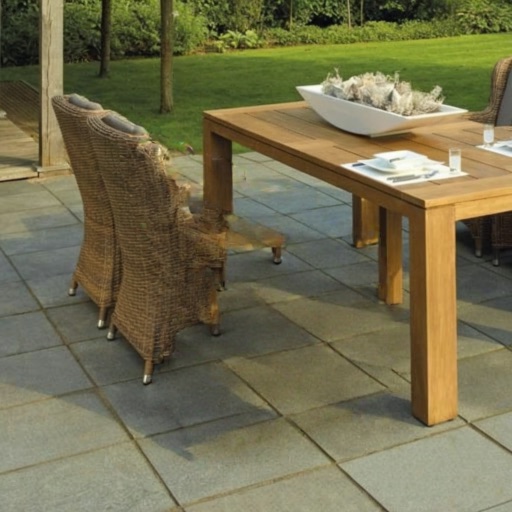} &
\includegraphics[width=0.089\textwidth]{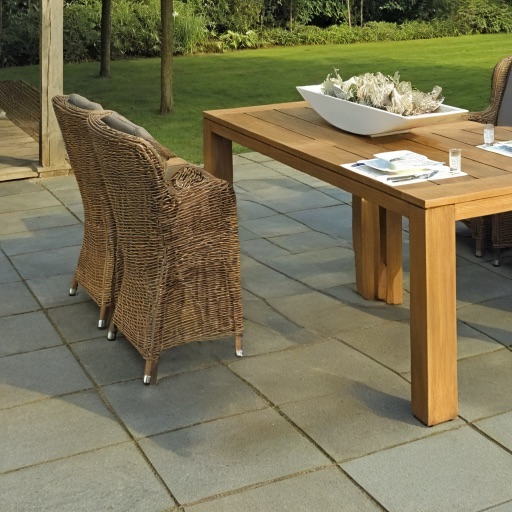} &
\includegraphics[width=0.089\textwidth]{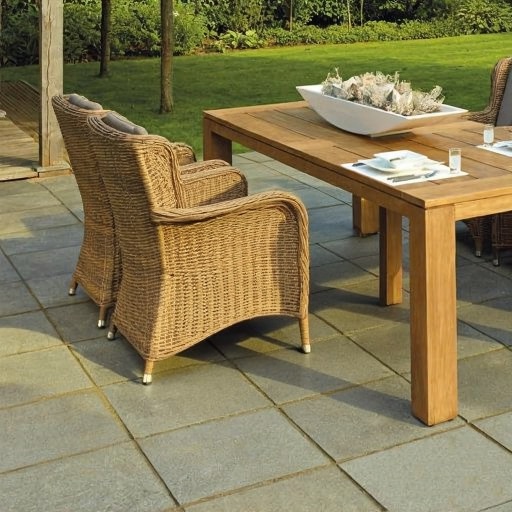} &
\includegraphics[width=0.089\textwidth]{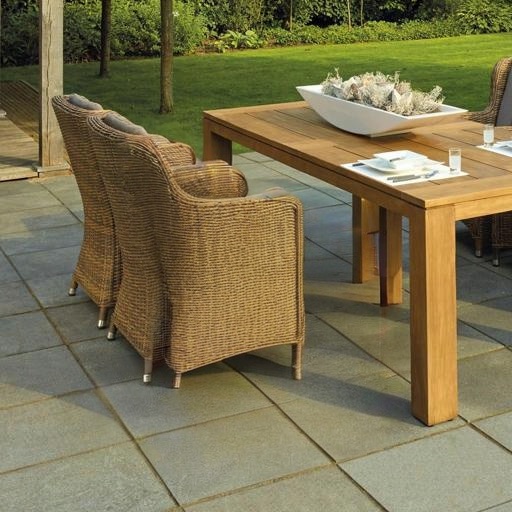} &
\includegraphics[width=0.089\textwidth]{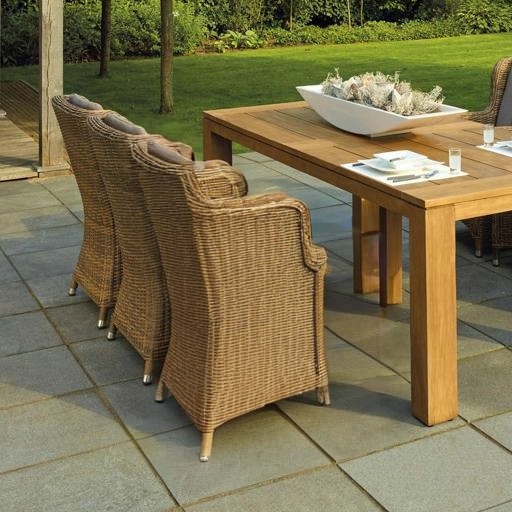} &
\includegraphics[width=0.089\textwidth]{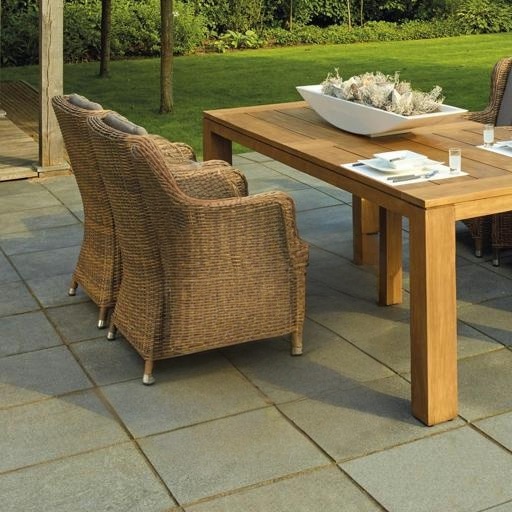} \\
\includegraphics[width=0.089\textwidth]{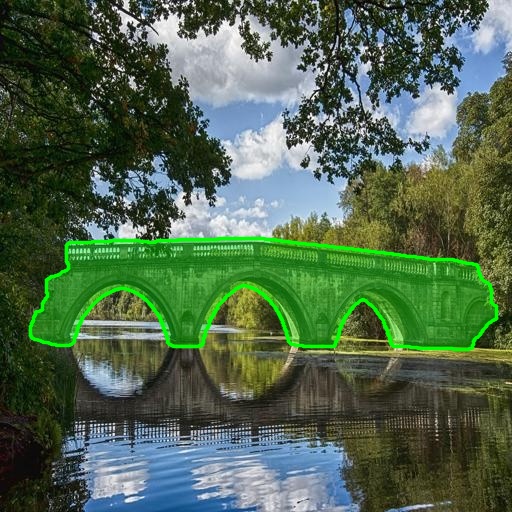} &
\includegraphics[width=0.089\textwidth]{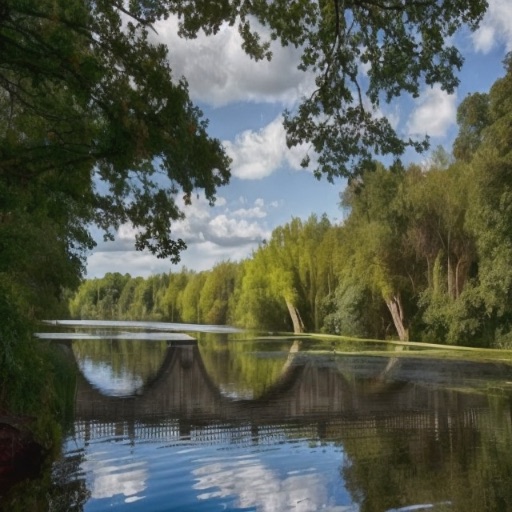} &
\includegraphics[width=0.089\textwidth]{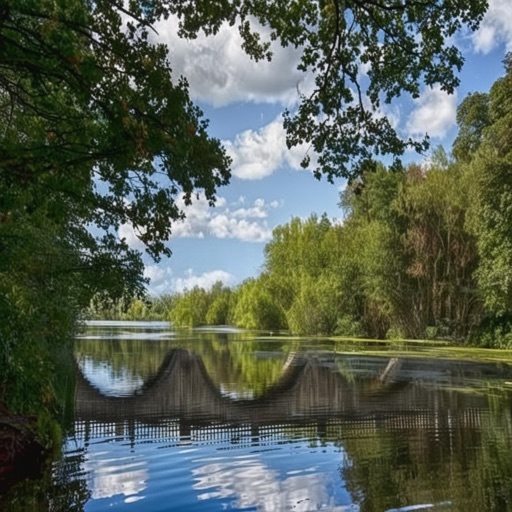} &
\includegraphics[width=0.089\textwidth]{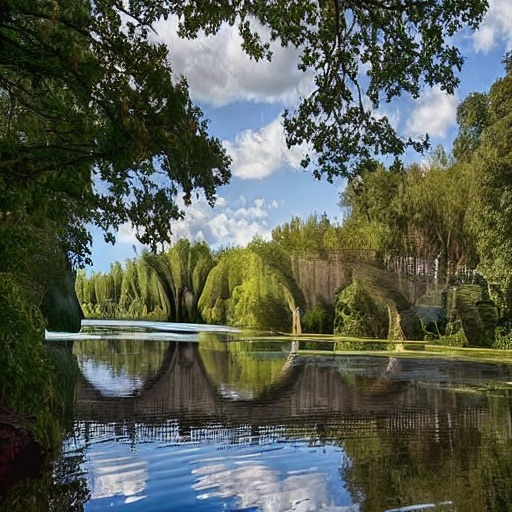} &
\includegraphics[width=0.089\textwidth]{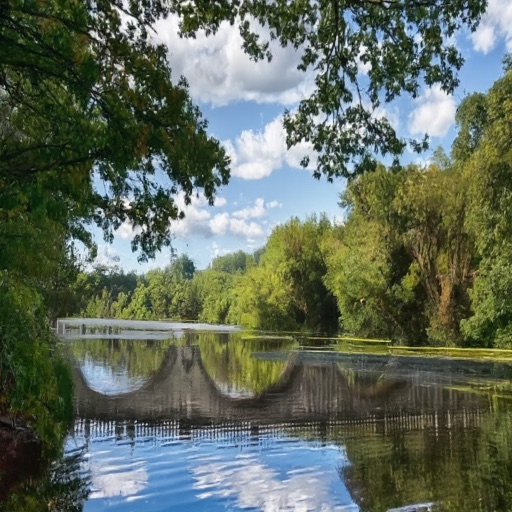} &
\includegraphics[width=0.089\textwidth]{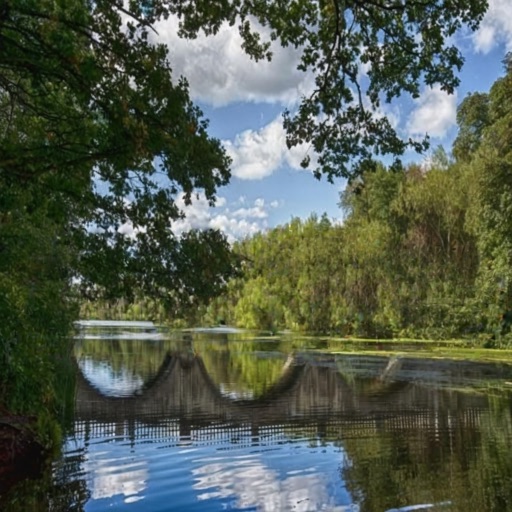} &
\includegraphics[width=0.089\textwidth]{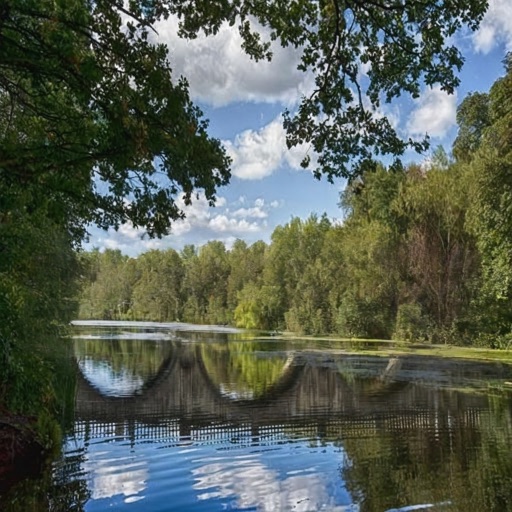} &
\includegraphics[width=0.089\textwidth]{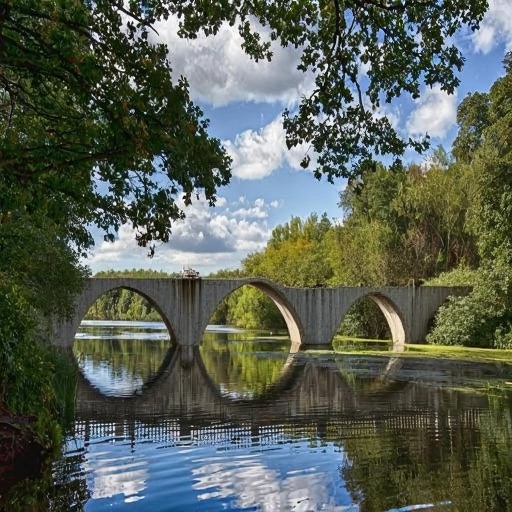} &
\includegraphics[width=0.089\textwidth]{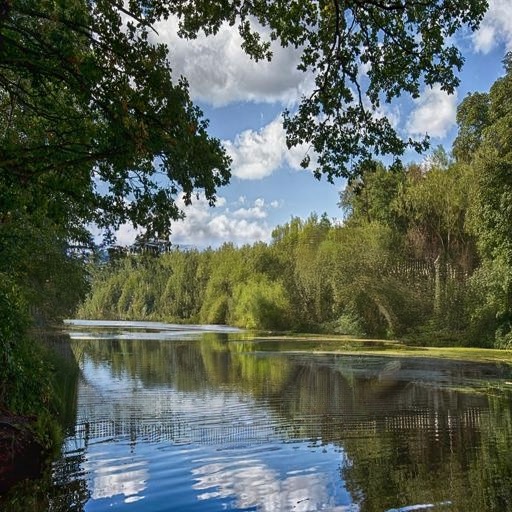} &
\includegraphics[width=0.089\textwidth]{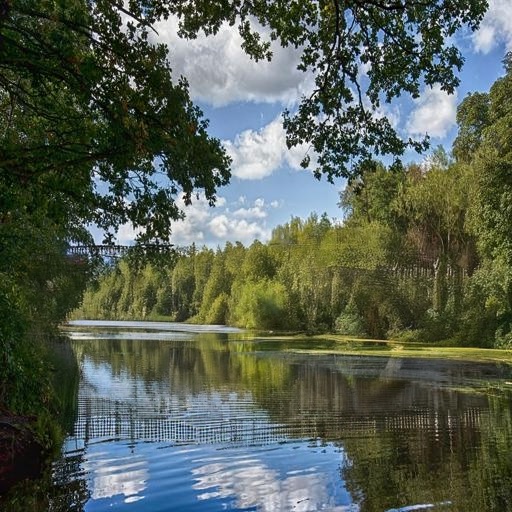} &
\includegraphics[width=0.089\textwidth]{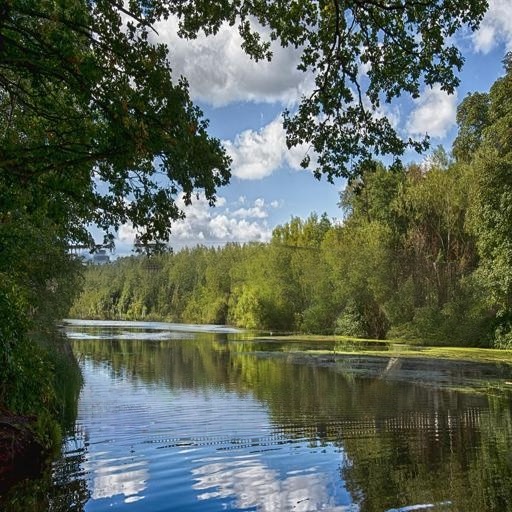} \\
\includegraphics[width=0.089\textwidth]{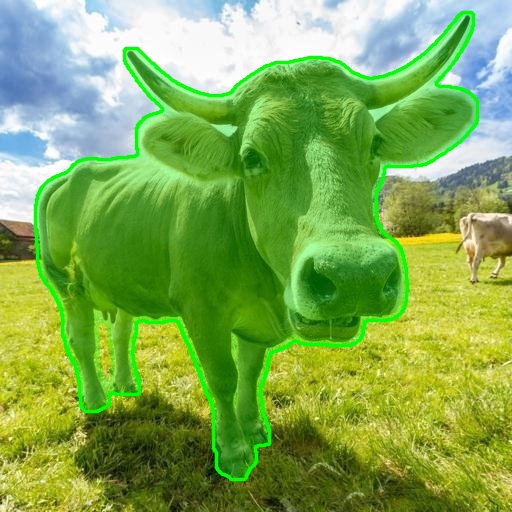} &
\includegraphics[width=0.089\textwidth]{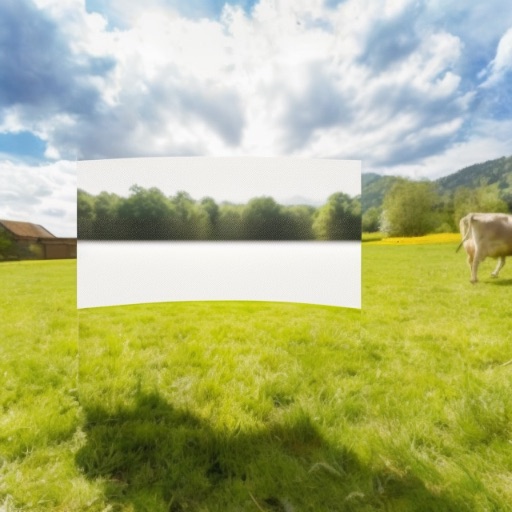} &
\includegraphics[width=0.089\textwidth]{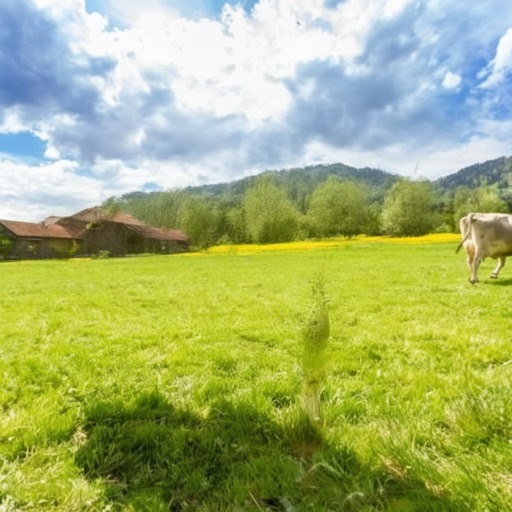} &
\includegraphics[width=0.089\textwidth]{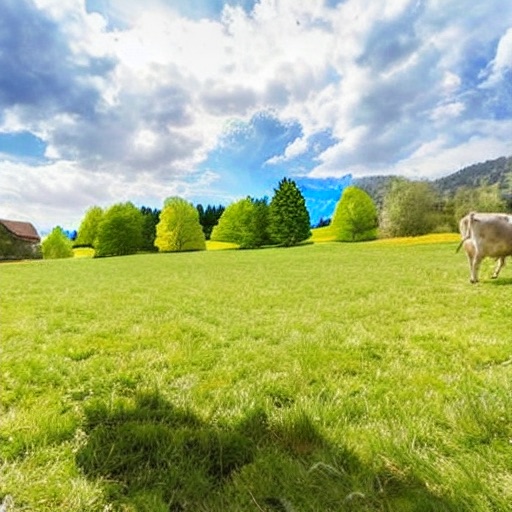} &
\includegraphics[width=0.089\textwidth]{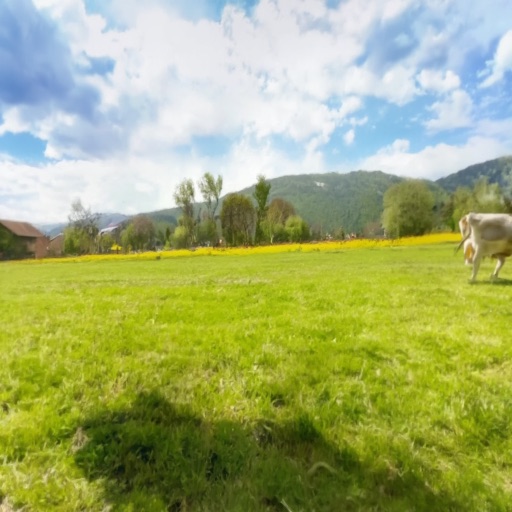} &
\includegraphics[width=0.089\textwidth]{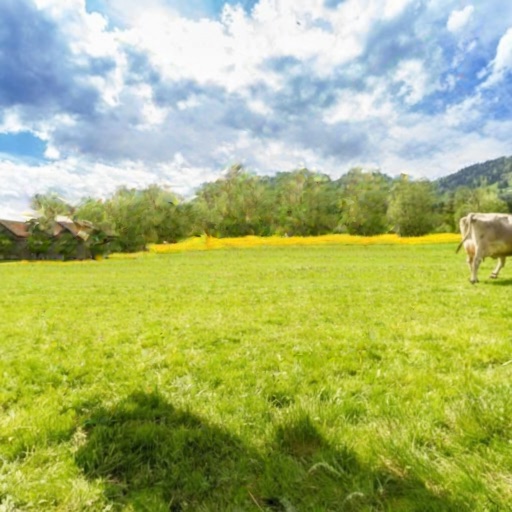} &
\includegraphics[width=0.089\textwidth]{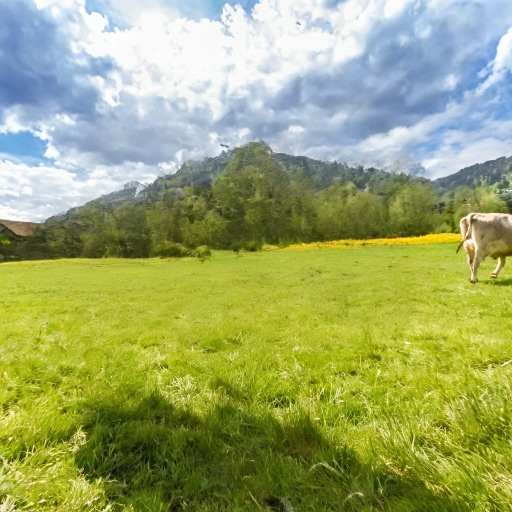} &
\includegraphics[width=0.089\textwidth]{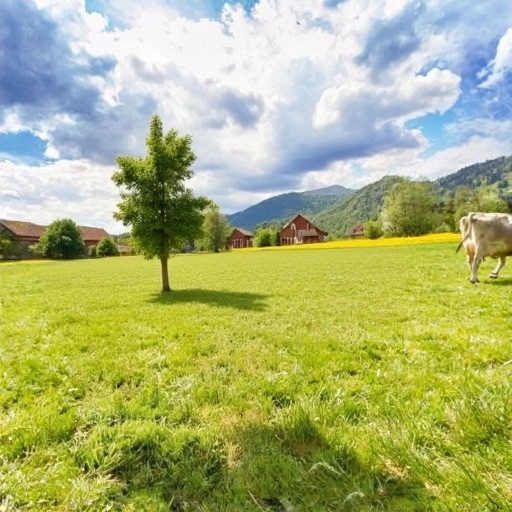} &
\includegraphics[width=0.089\textwidth]{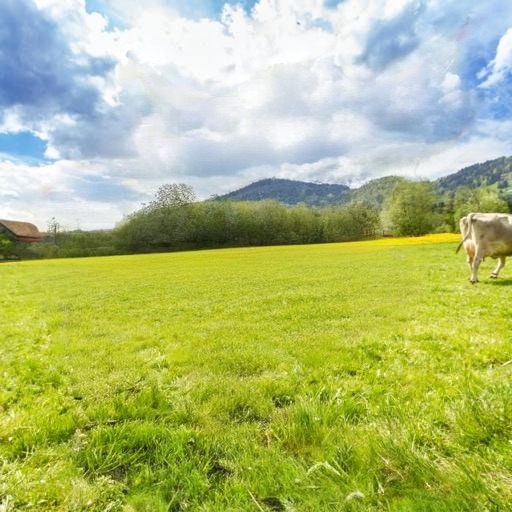} &
\includegraphics[width=0.089\textwidth]{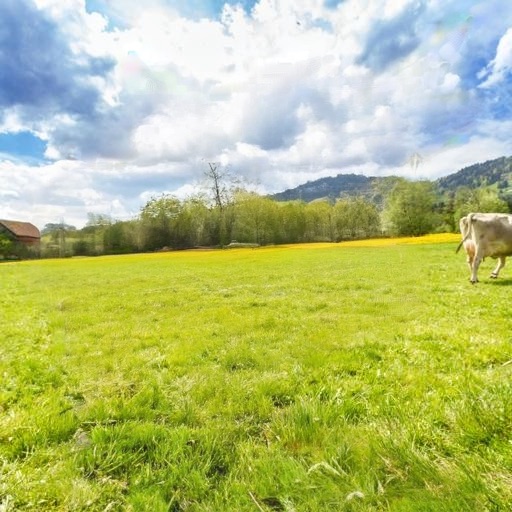} &
\includegraphics[width=0.089\textwidth]{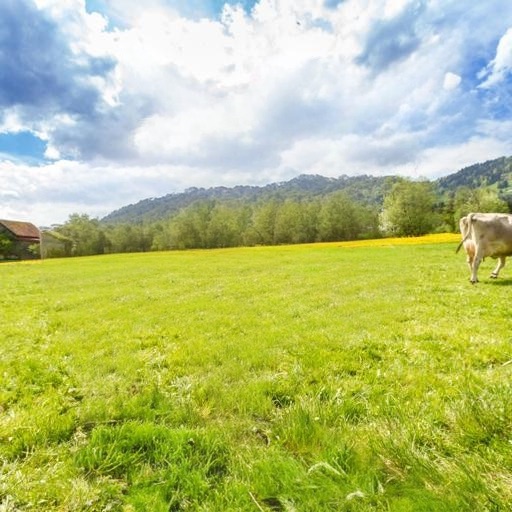} \\
\quallabels
\end{tabular}
\caption{Qualitative comparison on five representative OBER-Wild samples without ground-truth targets. TurboClear removes object effects while preserving unaffected background regions, and its spatially asymmetric behavior limits unnecessary generation outside the removal area. Zoom in to see more details.}
\label{fig:qualitative_comparison}
\end{figure*}

\subsection{Learnable Spatial Fusion}

Although RDM teaches the one-step student to perform asymmetric removal, a single generated stream still has to solve two conflicting goals: synthesize new content in affected regions and preserve identity elsewhere. As shown in Fig.~\ref{fig:lsf_spatial_asymmetry}, LSF makes this asymmetry explicit at inference time through two streams. The removal stream is the one-step UNet prediction $\hat{z}_0\in\mathbb{R}^{C\times h\times w}$, which contains the generated clean content. The preservation stream is the reference latent $z_y\in\mathbb{R}^{C\times h\times w}$ and the original image $y\in\mathbb{R}^{3\times H\times W}$, which provide identity information for unchanged regions.

We first extract a spatial prior from the object-token cross-attention of the one-step UNet. Let $P\in[0,1]^{N_h\times h\times w\times L}$ denote the cross-attention tensor, where $N_h$ is the number of attention heads and $L$ is the text-token length. For the object token $k_o$, we construct a normalized saliency prior
\begin{equation}
A = \frac{1}{N_h}\sum_{r=1}^{N_h}{\rm Norm}_{u,v}(P_{r,u,v,k_o}),
\quad A\in[0,1]^{1\times h\times w},
\end{equation}
where ${\rm Norm}_{u,v}$ denotes per-image spatial min-max normalization. This prior indicates where the UNet relies on the object condition, but it is not necessarily the optimal fusion coefficient. To see this, consider a spatial location $p$ and a local L2 reconstruction risk. The optimal pixel-space mixing coefficient satisfies
\begin{equation}
\alpha^\star(p)=
\mathop{\arg\min}_{\alpha\in[0,1]}
\|\alpha \hat{x}(p)+(1-\alpha)y(p)-x^\star(p)\|_2^2 .
\end{equation}
Thus, the best fusion decision depends on the local relation among the generated stream $\hat{x}$, the reference stream $y$, and the clean target $x^\star$, rather than on the attention prior alone.

Therefore, LSF learns a calibration function over both streams:
\begin{equation}
F = [\hat{z}_0,\ z_y,\ |\hat{z}_0-z_y|,\ m_o^\ell,\ A]
\in\mathbb{R}^{(3C+2)\times h\times w},
\end{equation}
where $m_o^\ell\in\{0,1\}^{1\times h\times w}$ is the object mask at latent resolution. A lightweight convolutional head predicts residual logit corrections $\Delta_z,\Delta_x\in\mathbb{R}^{1\times h\times w}$ as $\Delta_z,\Delta_x=\Phi_\phi(F)$. With $q(A)=\log(A/(1-A))$ after numerical clipping, the latent and pixel gates are
\begin{equation}
\alpha_z = \sigma(a_z q(A)+b_z+\Delta_z),
\quad
\alpha_x = \sigma(a_x q(A)+b_x+\Delta_x).
\end{equation}
The zero-initialized residual head makes the initial fusion close to the attention prior, while training learns when to trust the generated removal stream and when to copy from the preservation stream.

The latent fusion and final image fusion~\cite{levin2008closed} are defined as
\begin{equation}
z_f = (1-\alpha_z)\odot z_y + \alpha_z\odot \hat{z}_0,
\end{equation}
\begin{equation}
x_f = \alpha_x^\uparrow\odot D(z_f) + (1-\alpha_x^\uparrow)\odot y,
\end{equation}
where $\alpha_z,\alpha_x\in[0,1]^{1\times h\times w}$, $z_f\in\mathbb{R}^{C\times h\times w}$, $x_f\in\mathbb{R}^{3\times H\times W}$, and $\alpha_x^\uparrow\in[0,1]^{1\times H\times W}$ is bilinearly upsampled to image resolution. This dual-stream formulation directly matches the task requirement: affected regions should prefer the removal stream, whereas safe background regions should prefer the original image stream.

During LSF training, the one-step student is frozen and only the fusion head is optimized. The effect mask is again used only as supervision:
\begin{equation}
\begin{split}
\mathcal{L}_{\rm LSF}
=&\lambda_m\|m_e\odot(x_f-x^\star)\|_1\\
&+\lambda_b\|(1-m_e)\odot(x_f-x^\star)\|_1\\
&+\lambda_p{\rm LPIPS}(x_f,x^\star)
+\lambda_\alpha\mathcal{L}_{\alpha}.
\end{split}
\end{equation}
Let $m_e^-,m_e^+\in[0,1]^{1\times H\times W}$ be eroded and dilated effect masks. The gate regularization is
\begin{equation}
\begin{split}
\mathcal{L}_{\alpha}
&=
\frac{\|(1-\alpha_x^\uparrow)\odot m_e^-\|_1}
{\|m_e^-\|_1+\varepsilon}\\
&+
\frac{\|\alpha_x^\uparrow\odot(1-m_e^+)\|_1}
{\|1-m_e^+\|_1+\varepsilon}
+{\rm TV}(\alpha_x^\uparrow).
\end{split}
\end{equation}
It encourages high generation weights in the core affected region, low generation weights in safe background regions, and smooth spatial transitions.

\begin{table*}[t]
\centering
\setlength{\tabcolsep}{2.6pt}
\renewcommand{\arraystretch}{1.05}
\begin{tabular}{lccccc ccccc}
\toprule
 & \multicolumn{5}{c}{OBER-Test (512 $\times$ 512)} & \multicolumn{5}{c}{RORD-Val (960 $\times$ 540)} \\
\cmidrule(lr){2-6}\cmidrule(lr){7-11}
Method & FLOPs$\downarrow$ & LPIPS$\downarrow$ & LPIPS-L$\downarrow$ & PSNR$\uparrow$ & PSNR-M$\uparrow$
& FLOPs$\downarrow$ & LPIPS$\downarrow$ & LPIPS-L$\downarrow$ & PSNR$\uparrow$ & PSNR-M$\uparrow$ \\
\midrule
SDXL-INP & 256.9 & 0.1310 & 0.4409 & 22.44 & 12.26 & 256.9 & 0.1808 & 0.3995 & 20.23 & 12.26 \\
PowerPaint & 122.5 & 0.1583 & 0.3489 & 23.02 & 15.36 & 122.5 & 0.1903 & 0.2983 & 21.77 & 16.79 \\
GeoRemover & 3897.4 & 0.1454 & 0.1859 & 24.68 & 21.66 & 3897 & 0.1200 & 0.2094 & 24.49 & 19.41 \\
DesignEdit & 1727 & 0.1302 & 0.2548 & 26.39 & 20.61 & 1727 & 0.1937 & 0.3101 & 23.28 & 19.68 \\
CLIPAway & 80.37 & 0.1328 & 0.3614 & 22.22 & 14.44 & 80.37 & 0.1620 & 0.3015 & 21.11 & 15.63 \\
OmniEraser & 2097 & 0.2102 & 0.2630 & 24.35 & 21.29 & 2097 & 0.2289 & 0.3030 & 22.11 & 18.64 \\
Attentive Eraser & 549.2 & 0.0809 & 0.2436 & 27.17 & 20.85 & 549.2 & 0.1399 & 0.3014 & 24.10 & 17.84 \\
RORem & 331.3 & 0.0979 & 0.2391 & 26.22 & 19.14 & 331.3 & 0.1496 & 0.2534 & 24.02 & 18.58 \\
OmniPaint & 1057 & 0.0521 & \textbf{0.1299} & 29.06 & 23.57 & 2015 & 0.1178 & 0.2380 & 22.75 & 17.63 \\
ObjectClear & 63.63 & 0.0380 & 0.1540 & 32.06 & 23.34 & 86.60 & 0.0717 & 0.2235 & 27.35 & 19.40 \\
FlashClear & \underline{8.65} & \underline{0.0360} & \underline{0.1423} & \underline{32.50} & \underline{23.58} & \underline{15.61} & \underline{0.0684} & \underline{0.2073} & \underline{27.60} & \textbf{19.98} \\
TurboClear (ours) & \textbf{1.59} & \textbf{0.0286} & 0.1443 & \textbf{34.93} & \textbf{24.35} & \textbf{3.21} & \textbf{0.0627} & \textbf{0.2067} & \textbf{28.29} & \underline{19.85} \\
\bottomrule
\end{tabular}
\caption{Quantitative comparison on OBER-Test and RORD-Val. FLOPs are measured in tera-FLOPs (T). LPIPS-L and PSNR-M denote local LPIPS and masked PSNR, respectively. Best and second-best results are highlighted in bold and underlined.}
\label{tab:sota_quantitative}
\end{table*}

\section{Experiment}

\subsection{Experiment Settings}
\textbf{Implementation details.} TurboClear is built on the SDXL inpainting architecture. The one-step student is initialized from ObjectClear~\cite{zhao2025objectclear}, which also serves as the frozen multi-step teacher in RDM. Following the practice of distribution matching distillation~\cite{yin2024improved}, we first warm up the student with a perceptual paired objective and then optimize it with RDM for 25K iterations. After the one-step student is fixed, we train the LSF module for 10K iterations, so that the fusion head learns only the spatial calibration between the removal stream and the preservation stream. Training is conducted on 8 NVIDIA A800 GPUs with a total batch size of 16 under bfloat16 mixed precision. Unless otherwise specified, inference uses a fixed one-step scheduler with classifier-free guidance disabled and is evaluated on a single NVIDIA A800 GPU.

\textbf{Evaluation protocol.} We evaluate TurboClear quantitatively on two benchmarks with different resolutions. OBER-Test~\cite{zhao2025objectclear} contains 163 object-effect removal samples at $512\times512$ resolution and serves as the main benchmark. We further evaluate on the 343-sample RORD-Val~\cite{sagong2022rord} split used by ObjectClear~\cite{zhao2025objectclear} at $960\times540$ resolution to test high-resolution generalization. For qualitative evaluation, we additionally use the challenging OBER-Wild~\cite{zhao2025objectclear} set, which does not provide ground-truth targets. All methods are evaluated with the same input masks and prompts when applicable.

\textbf{Metrics.} For efficiency, we report theoretical denoising FLOPs as the primary cost metric, which avoids device-dependent latency variations and system-level implementation differences. We also report synchronized generation-path latency on a single A800 GPU for practical reference. For removal quality, we follow prior object removal evaluation~\cite{tang2026flash} and report PSNR, masked PSNR, LPIPS~\cite{zhang2018unreasonable}, and local LPIPS. PSNR measures global fidelity, masked PSNR focuses on the object-mask region, while LPIPS and local LPIPS evaluate perceptual similarity globally and locally.

\subsection{Comparison with SOTA methods}
We compare TurboClear with representative inpainting, object removal, and efficient removal baselines, including SDXL-INP~\cite{podell2023sdxl}, PowerPaint~\cite{zhuang2024task}, GeoRemover~\cite{zhu2025georemover}, DesignEdit~\cite{jia2024designedit}, CLIPAway~\cite{ekin2024clipaway}, OmniEraser~\cite{wei2025omnieraser}, Attentive Eraser~\cite{sun2025attentive}, RORem~\cite{li2025rorem}, OmniPaint~\cite{yu2025omnipaint}, ObjectClear~\cite{zhao2025objectclear}, and FlashClear~\cite{tang2026flash}.

\textbf{Quantitative evaluation.} Table~\ref{tab:sota_quantitative} compares TurboClear with prior state-of-the-art methods on OBER-Test at the native resolution and RORD-Val at a higher resolution. TurboClear outperforms previous methods on nearly all reported quality metrics at both resolutions, demonstrating effective object-effect removal while preserving unmasked background content. It achieves this quality with substantially lower computational cost: across the two resolutions, TurboClear reduces denoising computation by roughly $27$--$40\times$ relative to the ObjectClear and by roughly $628$--$665\times$ relative to the Flux-based SOTA OmniPaint. These results show that single-step distillation can retain the model capability and the spatial asymmetry between generation in affected regions and preservation elsewhere.

\textbf{Qualitative comparison.} We further assess the visual quality of TurboClear on OBER-Wild~\cite{zhao2025objectclear}, a challenging collection without ground-truth targets. Since no GT is available, this comparison focuses on visual inspection. As shown in Fig.~\ref{fig:qualitative_comparison}, compared with earlier methods, TurboClear removes the targeted object effects more completely while preserving the surrounding background more faithfully. Compared with OmniPaint~\cite{yu2025omnipaint}, TurboClear exhibits stronger spatial asymmetry: it suppresses unnecessary generation in unaffected regions while encouraging removal where needed. Compared with ObjectClear~\cite{zhao2025objectclear} and FlashClear~\cite{tang2026flash}, TurboClear maintains comparable visual removal quality while substantially reducing the denoising cost. In summary, our approach effectively preserves spatial asymmetry and model capability during the single-step distillation process.

\begin{figure}[t]
\centering
\setlength{\tabcolsep}{0pt}
\renewcommand{\arraystretch}{0}
\begin{tabular}{cccc}
\includegraphics[width=0.24\columnwidth]{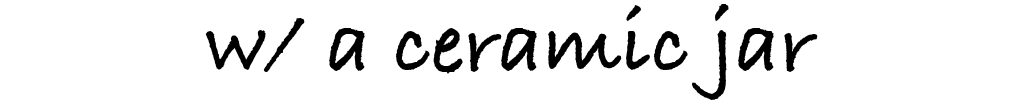} &
\includegraphics[width=0.24\columnwidth]{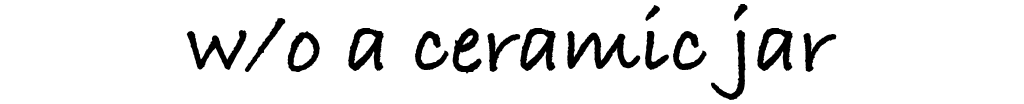} & & \\
\includegraphics[width=0.24\columnwidth]{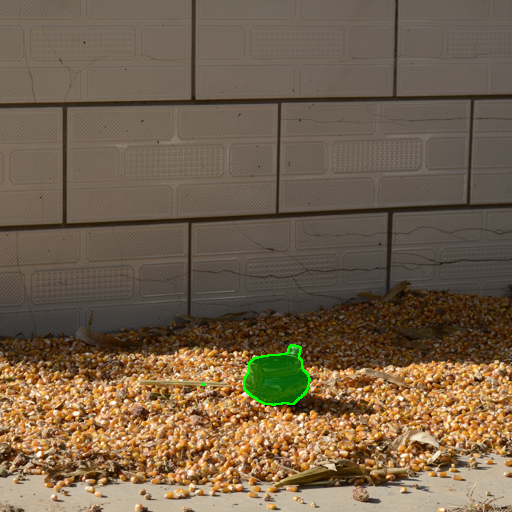} &
\includegraphics[width=0.24\columnwidth]{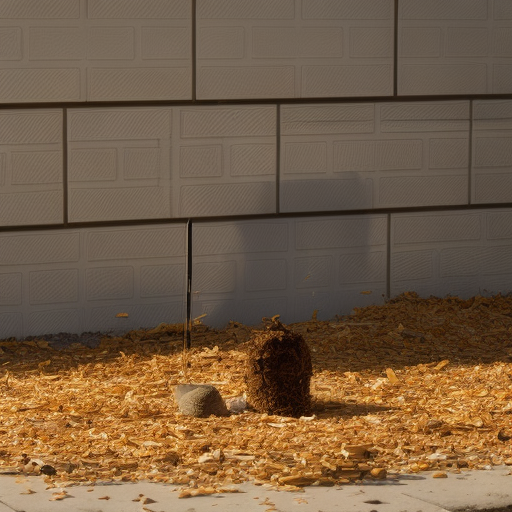} &
\includegraphics[width=0.24\columnwidth]{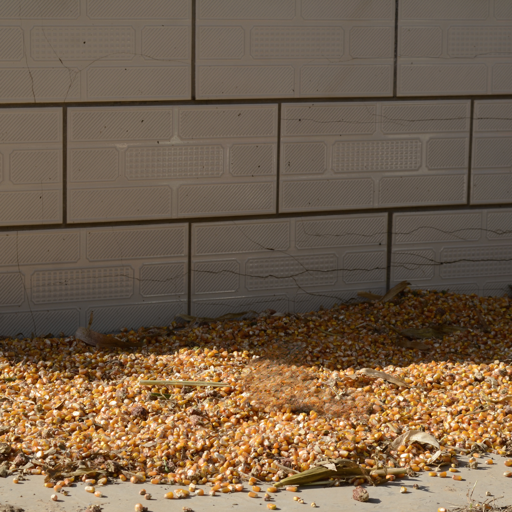} &
\includegraphics[width=0.24\columnwidth]{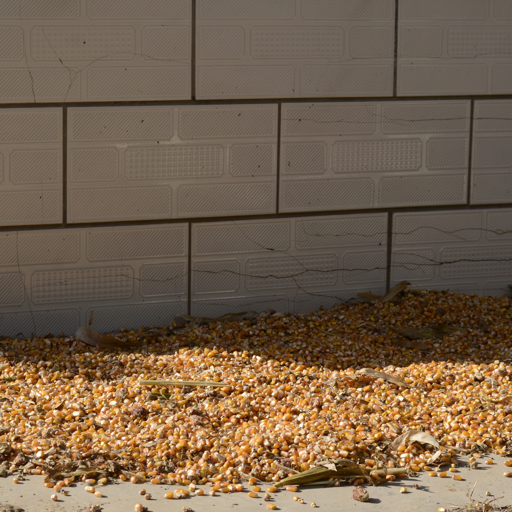} \\
\includegraphics[width=0.24\columnwidth]{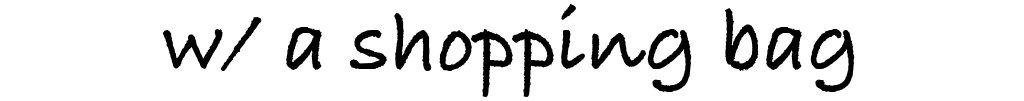} &
\includegraphics[width=0.24\columnwidth]{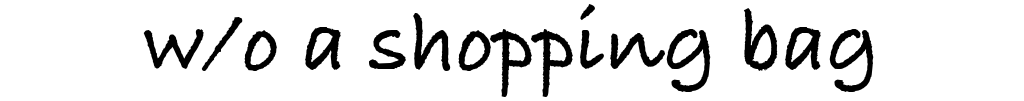} & & \\
\includegraphics[width=0.24\columnwidth]{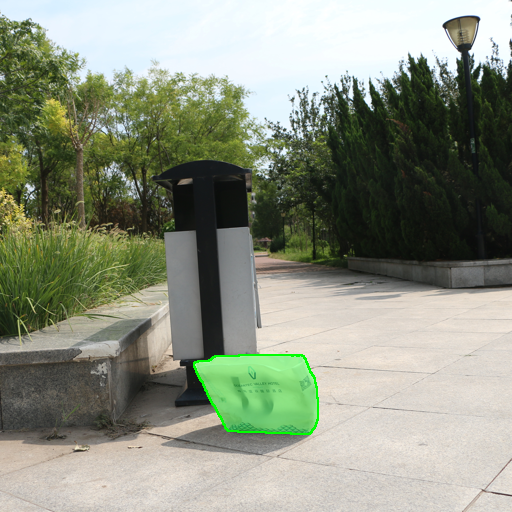} &
\includegraphics[width=0.24\columnwidth]{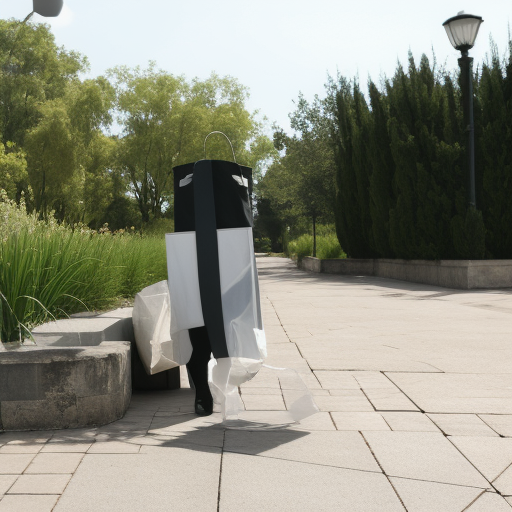} &
\includegraphics[width=0.24\columnwidth]{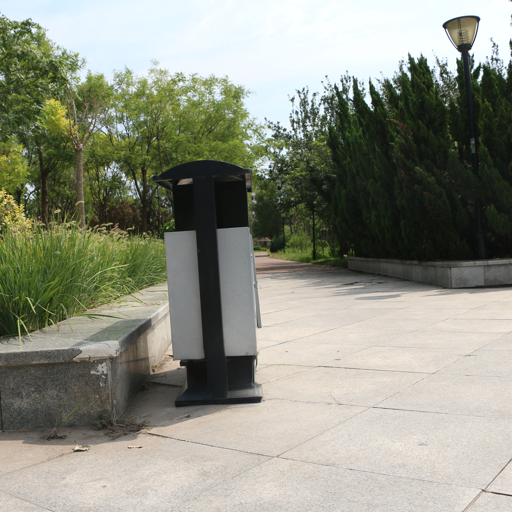} &
\includegraphics[width=0.24\columnwidth]{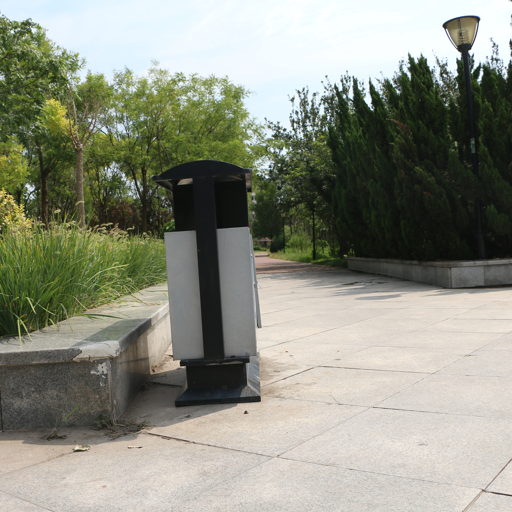} \\
\includegraphics[width=0.24\columnwidth]{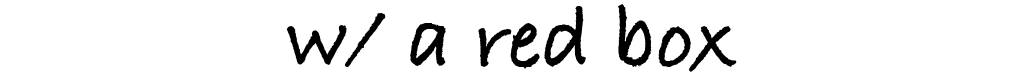} &
\includegraphics[width=0.24\columnwidth]{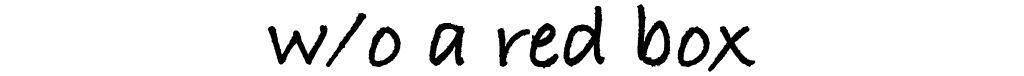} & & \\
\includegraphics[width=0.24\columnwidth]{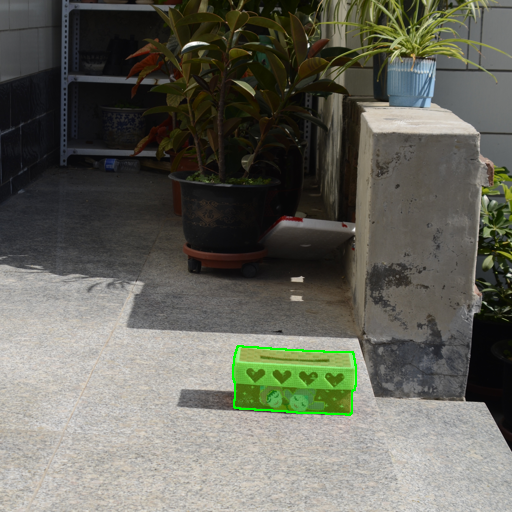} &
\includegraphics[width=0.24\columnwidth]{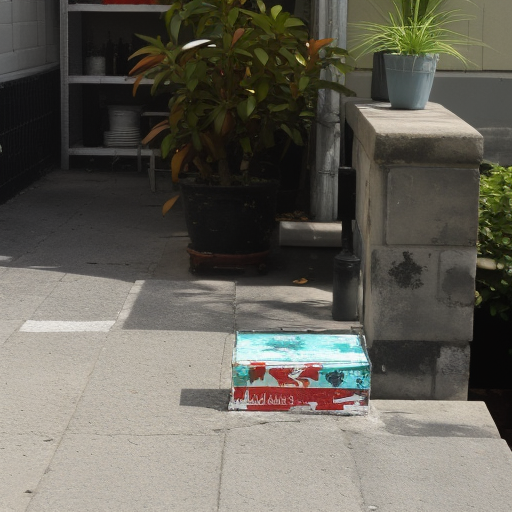} &
\includegraphics[width=0.24\columnwidth]{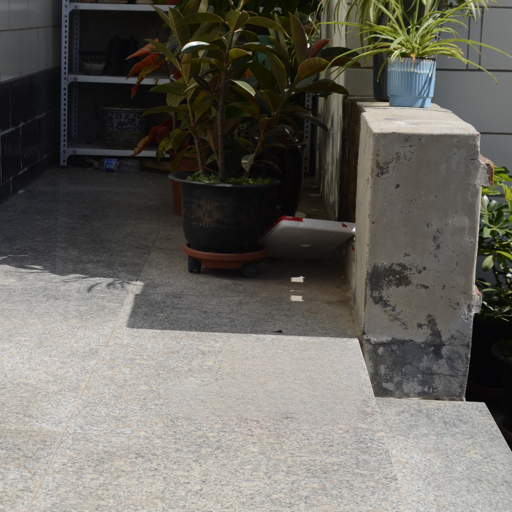} &
\includegraphics[width=0.24\columnwidth]{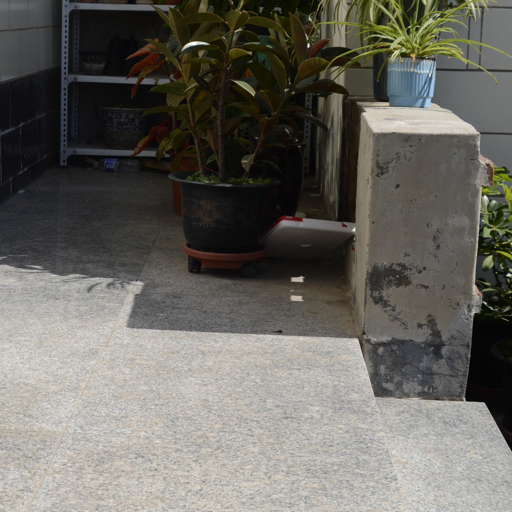} \\
\includegraphics[width=0.24\columnwidth]{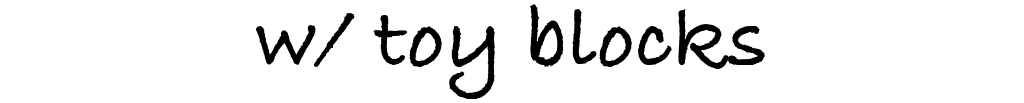} &
\includegraphics[width=0.24\columnwidth]{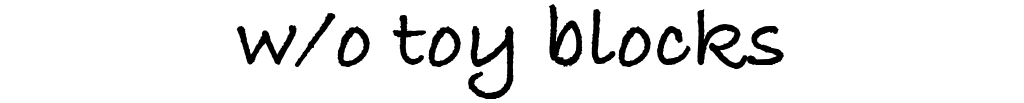} & & \\
\includegraphics[width=0.24\columnwidth]{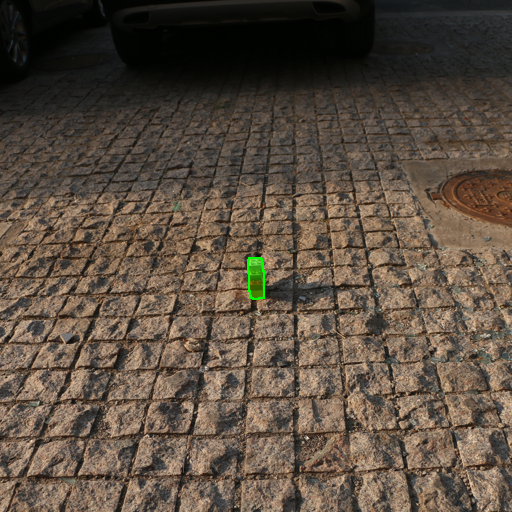} &
\includegraphics[width=0.24\columnwidth]{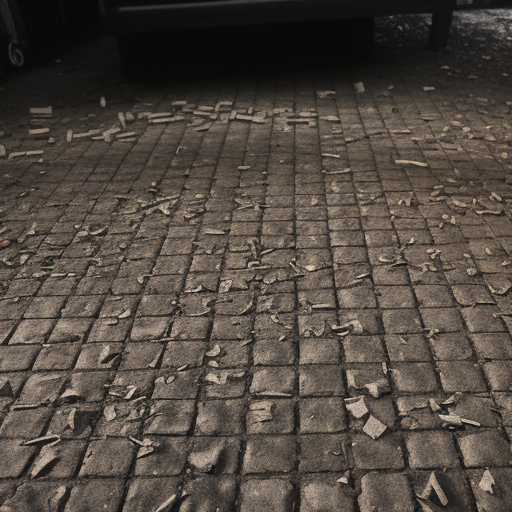} &
\includegraphics[width=0.24\columnwidth]{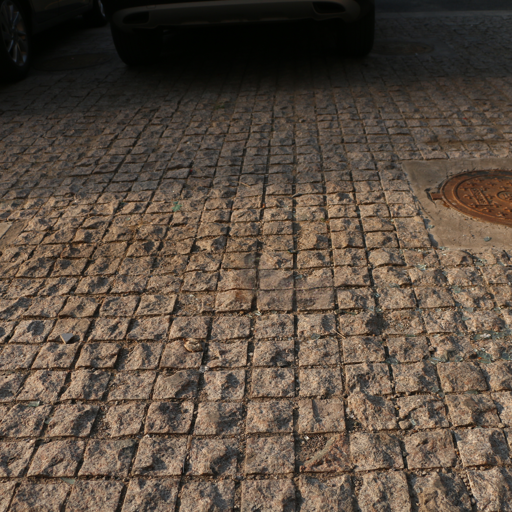} &
\includegraphics[width=0.24\columnwidth]{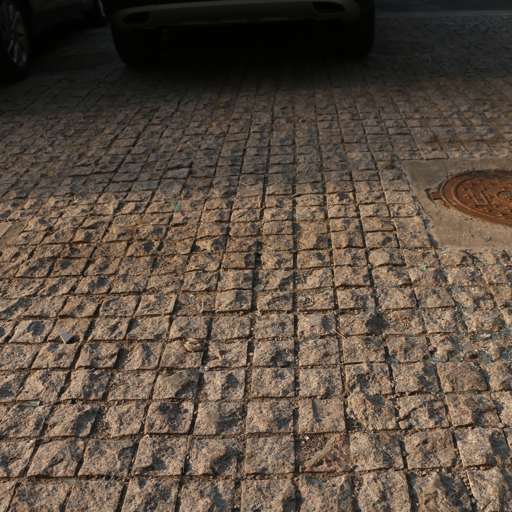} \\[0.2cm]
Input w/ mask & ChordEdit & TurboClear & GT \\
\end{tabular}
\caption{Qualitative comparison with ChordEdit on OBER-Test. Source and target prompts are shown above the Input with mask and ChordEdit columns. TurboClear achieves a cleaner removal of the masked object and its effects, while better preserving the background.}
\label{fig:text_based_comparison}
\end{figure}

\subsection{Comparison with Text-Based Methods}

ChordEdit~\cite{lu2026chordedit} is a recent state-of-the-art training-free method for one-step text-guided image editing. Unlike mask-conditioned object removal methods, ChordEdit requires a source-target prompt pair for every input. Since OBER-Test does not provide such text annotations, we use GPT-5.6~\cite{openai2026gpt56} as a vision-language model to identify the masked object and generate a pair in the form of ``with [object]'' and ``without [object].'' We then refine all 163 pairs by correcting object identities and attributes, and adding necessary spatial qualifiers, ensuring one-to-one alignment with OBER-Test. Subsequently, we evaluated the image quality metrics, denoising FLOPs, and latency.

For a fair latency comparison, both methods are measured with CUDA synchronization after two warm-up runs, excluding input preprocessing, image/text condition encoding, and disk I/O. The measured path contains each method's editing backbone and one VAE decode; for TurboClear, it additionally includes attention extraction and LSF. 

\begin{table}[t]
\centering
\small
\setlength{\tabcolsep}{0.9pt}
\renewcommand{\arraystretch}{1.2}
\begin{tabular}{lcccccc}
\toprule
Method & FLOPs$\downarrow$ & LPIPS$\downarrow$ & LPIPS-L$\downarrow$ & PSNR$\uparrow$ & PSNR-M$\uparrow$ & Lat.$\downarrow$ \\
\midrule
ChordEdit  & 4.022 & 0.3864 & 0.3893 & 17.39 & 18.28 & 0.1310 \\
TurboClear & \textbf{1.589} & \textbf{0.0286} & \textbf{0.1443} & \textbf{34.93} & \textbf{24.35} & \textbf{0.0412} \\
\bottomrule
\end{tabular}
\caption{Comparison with the text-based one-step editor ChordEdit on OBER-Test. FLOPs are measured in tera-FLOPs and latency in seconds.}
\label{tab:text_based_comparison}
\end{table}

As shown in Table~\ref{tab:text_based_comparison}, TurboClear substantially outperforms ChordEdit on all paired fidelity metrics with fewer inference FLOPs and latency. The qualitative examples in Fig.~\ref{fig:text_based_comparison} further show that explicit mask conditioning better localizes the removal and preserves the surrounding scene.

\subsection{Ablation Study}

\textbf{Distillation and inference strategy.} Table~\ref{tab:strategy_ablation} compares different one-step distillation strategies. Generic LCM~\cite{luo2023latent}, DMD2~\cite{yin2024improved} and adversarial RAD~\cite{tang2026flash} objectives are less effective for object-effect removal, whereas the RDM improves both global and local fidelity over DMD2. Adding LSF yields a further substantial gain in PSNR and LPIPS than naive fusion like AGF~\cite{zhao2025objectclear}. This confirms that region-calibrated training and spatial fusion address complementary aspects of removal and preservation. Figure~\ref{fig:lsf_spatial_asymmetry} provides a direct visual comparison: LSF confines changes more tightly to the object-effect region and preserves the unaffected background.

\begin{table}[t]
\centering
\small
\setlength{\tabcolsep}{3.0pt}
\renewcommand{\arraystretch}{1.05}
\begin{tabular}{lcccc}
\toprule
Method & LPIPS$\downarrow$ & LPIPS-L$\downarrow$ & PSNR$\uparrow$ & PSNR-M$\uparrow$ \\
\midrule
LCM        & 0.1203 & 0.2042 & 26.56 & 23.55 \\
DMD2       & 0.0897 & 0.1642 & 27.85 & 23.98 \\
DMD2 + GAN & 0.0842 & 0.1531 & 27.95 & 23.87 \\
RAD        & 0.1108 & 0.1889 & 26.60 & 22.89 \\
RDM & 0.0802 & \underline{0.1489} & 28.47 & \textbf{24.40} \\
RDM + AGF & \underline{0.0442} & 0.1591 & \underline{31.71} & 24.25 \\
RDM + LSF (ours) & \textbf{0.0286} & \textbf{0.1443} & \textbf{34.93} & \underline{24.35} \\
\bottomrule
\end{tabular}
\caption{Ablation of one-step distillation strategies. Best and second-best results are highlighted in bold and underlined.}
\label{tab:strategy_ablation}
\end{table}

\textbf{Advantages of LSF.} While ObjectClear's attention-guided fusion (AGF)~\cite{zhao2025objectclear} relies on multi-step attention refinement that localizes poorly in one-step models—often reintroducing artifacts—LSF learns spatial gates via task-specific asymmetric supervision. This explicitly enforces an edit-and-preserve objective. As shown in Figs.~\ref{fig:lsf_spatial_asymmetry} and \ref{fig:lsf_vs_agf}, LSF strictly confines changes to the target region, avoiding erroneous fusion and preserving background consistency.

\newcommand{\agflsflabels}{%
\makebox[0.19\columnwidth][c]{\scriptsize Input w/ Mask} &
\makebox[0.19\columnwidth][c]{\scriptsize AGF} &
\makebox[0.19\columnwidth][c]{\scriptsize AGF Diff.} &
\makebox[0.19\columnwidth][c]{\scriptsize LSF} &
\makebox[0.19\columnwidth][c]{\scriptsize LSF Diff.} \\
}

\begin{figure}[t]
\centering
\setlength{\tabcolsep}{0pt}
\renewcommand{\arraystretch}{1.0}
\begin{tabular}{ccccc}
\includegraphics[width=0.19\columnwidth]{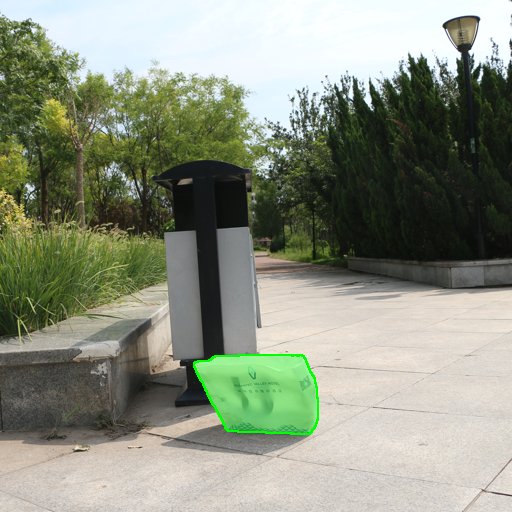} &
\includegraphics[width=0.19\columnwidth]{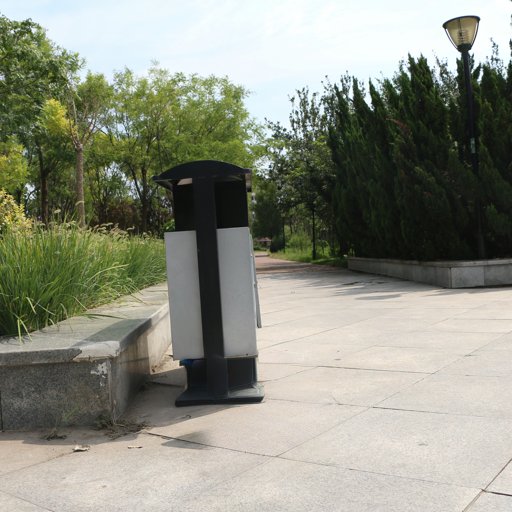} &
\includegraphics[width=0.19\columnwidth]{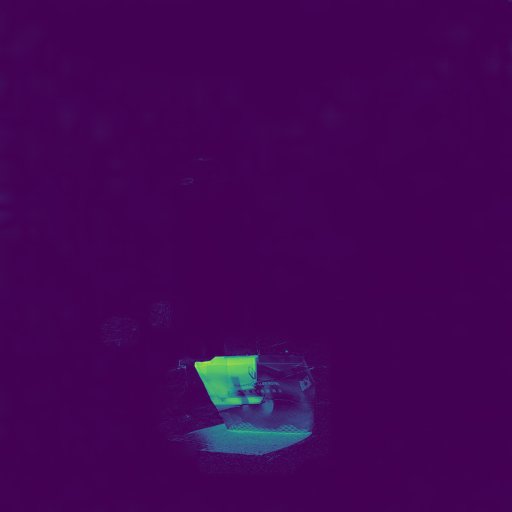} &
\includegraphics[width=0.19\columnwidth]{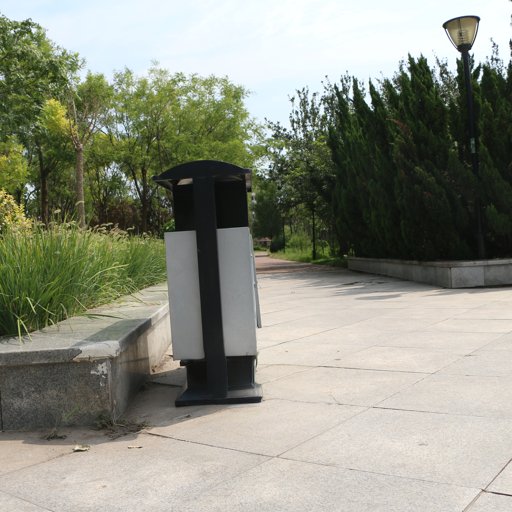} &
\includegraphics[width=0.19\columnwidth]{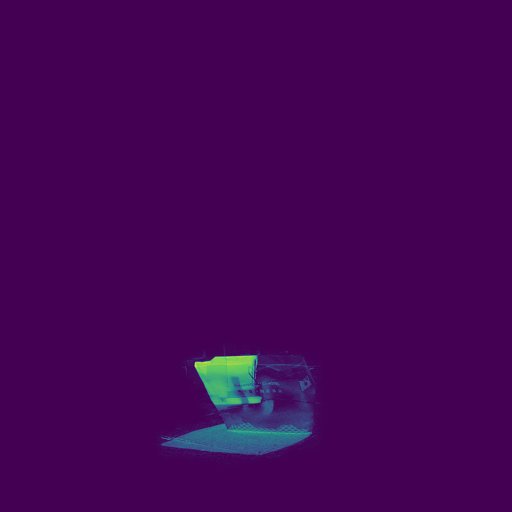} \\
\includegraphics[width=0.19\columnwidth]{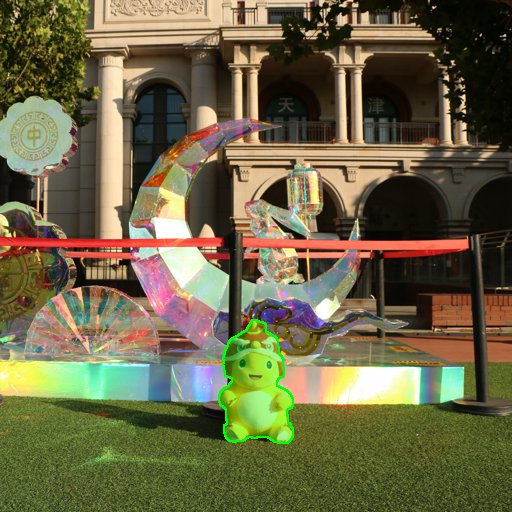} &
\includegraphics[width=0.19\columnwidth]{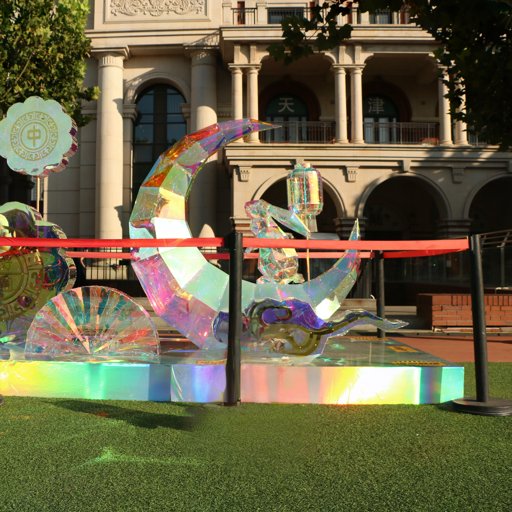} &
\includegraphics[width=0.19\columnwidth]{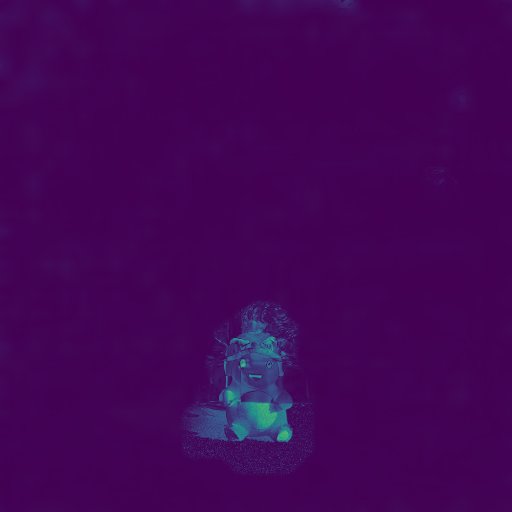} &
\includegraphics[width=0.19\columnwidth]{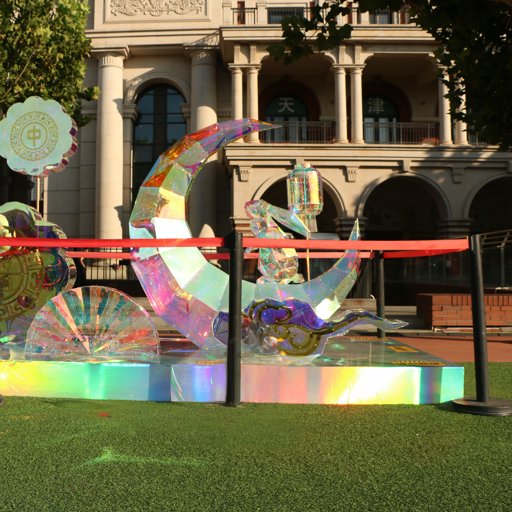} &
\includegraphics[width=0.19\columnwidth]{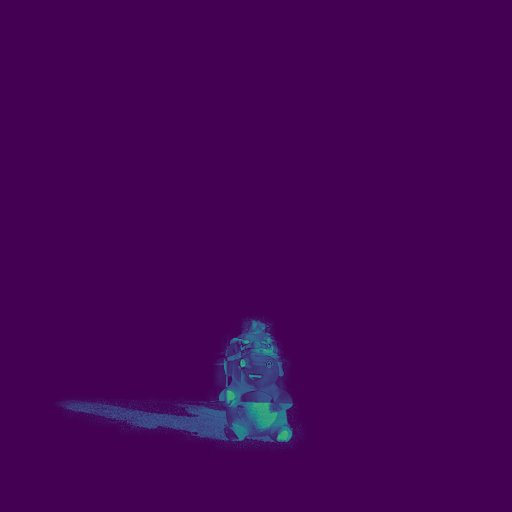} \\
\agflsflabels
\end{tabular}
\caption{Visual comparison of AGF and LSF implemented with RDM. Difference maps show RGB difference from the input. LSF localizes regeneration more precisely, avoiding residual artifacts while preserving the background.}
\label{fig:lsf_vs_agf}
\end{figure}

\section{Conclusion}

We proposed TurboClear, a one-step SDXL model for object-effect removal that explicitly preserves the task's spatial asymmetry: affected regions are regenerated while unaffected content remains unchanged. Region-Calibrated Distribution Matching and Learnable Spatial Fusion retain this behavior with one UNet evaluation and a lightweight fusion head. Across native- and high-resolution benchmarks, TurboClear delivers state-of-the-art removal and background fidelity at substantially lower computational cost, demonstrating that single-step distillation can preserve both model capability and spatially selective generation.

Our controlled ablations further clarify the complementary roles of the two components. RDM improves the one-step student by calibrating its generative supervision according to the object-effect region, reducing the conflict between removal and preservation. With the distilled student fixed, LSF provides a more precise alternative to attention-guided fusion by learning where to retain the input and where to use the generated prediction. Together, they improve both local removal quality and global background consistency rather than trading one objective for the other.

\clearpage
\bibliography{aaai2027}


\clearpage
% Appendix content from AAAI-27/SupplementaryMaterial.tex.
\setcounter{secnumdepth}{2}
\raggedbottom
\appendix
\section{Implementation Details}

\paragraph{Denoising FLOPs measurement.}
We measure theoretical denoising FLOPs by instrumenting the computation graph executed by the denoising backbone and accumulating its operations over all sampling steps. Each compared method is evaluated using its default inference configuration, including the recommended number of sampling steps, classifier-free guidance (CFG), and image resizing strategy. Formally, for a sampler with $N$ denoising steps, we compute
\begin{equation}
\mathcal{F}_{\rm denoise}
=\sum_{k=1}^{N}
\operatorname{FLOPs}\!\left(G_k;
B_k,C_k,H_k,W_k\right),
\end{equation}
where $G_k$ denotes the denoising computation graph executed at step $k$, and $(B_k,C_k,H_k,W_k)$ is its effective input shape. CFG increases the effective batch size by evaluating conditional and unconditional branches, while different resizing rules produce different latent spatial dimensions. We include operations executed by the default denoising path and exclude condition encoders, VAE encoding/decoding, the lightweight fusion head, evaluation networks, and post-processing.

\paragraph{Quality metric definitions.}
All predictions and targets are converted from the model range $[-1,1]$ to $[0,1]$ before computing the quality metrics. Let $M_o$ denote the binarized input object mask. The evaluator thresholds this mask at $0.5$ and broadcasts it over the three RGB channels. We compute masked PSNR as
\begin{equation}
\begin{aligned}
\operatorname{MSE}_{M_o}
\;&=
\frac{\sum_{c,i,j}M_{o,ij}(\hat{x}_{cij}-x^{\star}_{cij})^2}
{\sum_{c,i,j}M_{o,ij}+\varepsilon},\\
\operatorname{PSNR\text{-}M}
\;&=-10\log_{10}\!\left(\max(\operatorname{MSE}_{M_o},10^{-12})\right).
\end{aligned}
\end{equation}
Here $\varepsilon=10^{-8}$.
We call this metric PSNR-mask (reported as PSNR-M in the tables). It measures fidelity inside the input object mask. We additionally report PSNR-BG over its complement $1-M_o$ to quantify preservation of the unmasked background.

For LPIPS-Local (denoted LPIPS-L), we compute $d_{\rm LPIPS}(\hat{x}|_{\mathcal{B}(M_o)},x^\star|_{\mathcal{B}(M_o)})$, where $\mathcal{B}(M_o)$ is the object-mask bounding box padded symmetrically to a minimum side length of $64$ pixels and clipped to the image boundary. The padded crop includes contextual background, and we use the default AlexNet LPIPS network; lower values indicate better perceptual fidelity. DISTS-Local uses the same crop.

\section{More Training Details}

\paragraph{Data and preprocessing.}
We train on the OBER training split. Images are cropped around the object, bicubically resized to $512\times512$, and mapped to $[-1,1]$; masks are binarized and resized with nearest-neighbor interpolation. The fixed prompt is \texttt{remove the instance of object}. The object mask conditions the SDXL inpainting UNet and image-prompt encoder, while the object-effect mask supervises region losses, DMD masking, and attention localization. Training uses object-centered crops, horizontal flips with probability $0.5$, and random mask dilation/erosion; color, rotation, and blank-mask augmentation are disabled. The DMD and fusion stages use batch size $2$ per process, and warm-up uses batch size $6$ per process. All random seeds are set to 231 during training and inference whenever possible.

\paragraph{One-step warm-up and initialization.}
Before RDM, we warm up the full student UNet for $1{,}000$ steps using a VGG-based LPIPS objective. We use AdamW with learning rate $1\times10^{-5}$, betas $(0.9,0.999)$, batch size $6$ per process, fixed one-step DDIM timestep $399$, adopting the timestep shift technique from OpenDMD and Pixart-Sigma, and $512\times512$ inputs. The warm-up objective is
\begin{equation}
\mathcal{L}_{\rm warm}=1.0\,\mathcal{L}_{\rm LPIPS}+0.01\,\mathcal{L}_{\rm loc};
\end{equation}
all other loss terms are disabled.

\paragraph{RDM distillation stage.}
We optimize the full student UNet for $25$K steps with the frozen ObjectClear SDXL inpainting UNet as teacher. One-step DDIM uses fixed timestep $399$; DMD samples timesteps uniformly from $[20,980]$, uses unit CFG for the teacher and fake score model, and updates the fake score model five times per student update. The object-effect mask is resized to latent resolution, dilated with a radius-$2$ max-pool, blurred with a radius-$1$ average-pool, and used for normalized masked DMD gradients.

The active generator objective is
\begin{equation}
\begin{aligned}
\mathcal{L}_{G}
&= 0.1\,\mathcal{L}_{\rm RDM}
 + 0.1\,\mathcal{L}_{\rm mask}
 + 0.1\,\mathcal{L}_{\rm bg} \\
&\quad
 + 1.0\,\mathcal{L}_{\rm LPIPS}
 + 0.01\,\mathcal{L}_{\rm loc}.
\end{aligned}
\end{equation}
Here $\mathcal{L}_{\rm mask}$ and $\mathcal{L}_{\rm bg}$ are effect-region and background masked image L1 losses. The localization term uses the five central cross-attention block groups with object-token index $5$. The fake score model uses unit-weight epsilon-prediction MSE. AdamW uses learning rates $5\times10^{-7}$ for the student and fake score model, with betas $(0.9,0.999)$.

\paragraph{Learnable Spatial Fusion stage.}
After freezing the one-step student, we train the fusion head for $10$K steps. The head takes the predicted latent, input-image latent, their absolute difference, the object-mask latent, and the attention prior as input; it has hidden width $32$, three convolutional layers, and logit clipping $\epsilon=10^{-4}$. The object mask is used at inference, while the object-effect mask provides supervision. The fusion losses use weights $(1.0,0.5,0.2)$ for effect-region L1, background L1, and global LPIPS, and $(0.05,0.05,0.01)$ for foreground-alpha, background-alpha, and total-variation regularization. The effect-mask core and safe background use erosion radius $8$ and dilation radius $16$, respectively. AdamW uses learning rate $10^{-4}$ and betas $(0.9,0.999)$ with bfloat16 training. At inference, the learned latent and pixel alpha maps blend the generated result with the input-image stream.

\section{FLOPs and Latency Comparison}

Wall-clock latency is sensitive to differences among codebases as well as their I/O pipelines, kernel implementations, and system-level optimizations. It therefore provides only limited guidance for practical industrial deployment and is not used as the primary efficiency metric in the main paper; instead, we use denoising FLOPs for the main comparison. For completeness, Table~\ref{tab:efficiency_comparison} reports both FLOPs and latency. All methods use their default inference configurations.

We place CUDA synchronization immediately before and after the measured inference path. For each method, this path contains the editing backbone and one VAE decoding pass. For TurboClear, it additionally contains the LSF module. The reported values are average latency per image in seconds.

\begin{table*}[t]
\centering
\small
\setlength{\tabcolsep}{7pt}
\begin{tabular}{lcccc}
\toprule
& \multicolumn{2}{c}{OBER-Test} & \multicolumn{2}{c}{RORD-Val} \\
\cmidrule(lr){2-3}\cmidrule(lr){4-5}
Strategy & FLOPs (T) $\downarrow$ & Latency (s) $\downarrow$ & FLOPs (T) $\downarrow$ & Latency (s) $\downarrow$ \\
\midrule
OmniPaint & 1058 & 20.34 & 2016 & 32.03 \\
ObjectClear & \underline{63.63} & \underline{2.290} & \underline{86.60} & \underline{6.077} \\
TurboClear (ours) & \textbf{1.589} & \textbf{0.04117} & \textbf{3.207} & \textbf{0.07094} \\
\bottomrule
\end{tabular}
\caption{FLOPs and per-image latency on OBER-Test and RORD-Val. Lower is better; the best and second-best results are highlighted in bold and underlined, respectively.}
\label{tab:efficiency_comparison}
\end{table*}

On OBER-Test, TurboClear is $55.62\times$ faster than its teacher model, ObjectClear, and $494.10\times$ faster than the Flux-based OmniPaint. On RORD-Val, the corresponding speedups are $85.67\times$ and $451.59\times$, respectively.

\section{Additional Perceptual and Reference-Free Evaluation}

\paragraph{Protocol.}
The main paper follows prior object-removal evaluation and reports PSNR, PSNR-M, LPIPS, and LPIPS-L. To assess complementary aspects of quality and reduce dependence on metrics related to our reconstruction and perceptual training objectives, we additionally report PSNR-BG, DISTS~\cite{ding2022dists}, DISTS-Local, MUSIQ~\cite{ke2021musiq}, CLIP-IQA~\cite{wang2023clipiqa}, CFD~\cite{yu2025omnipaint}, and FID$_{163}$~\cite{heusel2017gans}. PSNR-BG measures fidelity outside the object mask. DISTS evaluates full-reference structural and textural similarity globally and on the local object-mask bounding-box crop. MUSIQ and CLIP-IQA are no-reference perceptual quality estimators, while CFD is a task-oriented no-reference measure of context consistency and object hallucination. None of DISTS, MUSIQ, CLIP-IQA, CFD, or FID is used as a TurboClear training objective.

All methods are evaluated on the same 163 OBER-Test samples at their native $512\times512$ resolution, with exact filename pairing and no evaluation-time resizing. The classifier-free guidance (CFG) scale is fixed to $1.0$ for all evaluated methods. ObjectClear uses its default AGF setting. OmniPaint is run with its default configuration and is marked N/A in the Fusion column because this fusion categorization is not applicable to its different FLUX-based architecture. The object mask is binarized at $0.5$, and the local crop is expanded to at least $64\times64$ pixels without resizing. We report dataset means and paired 95\% bootstrap confidence intervals (CIs) using 10,000 resamples with seed 231. The same sampled image indices are used for both methods in every paired replicate. These CIs quantify variation across test images, not variation across training or inference seeds. Since FID is unreliable with only 163 samples, FID$_{163}$ is included solely as an auxiliary distributional statistic.

To disentangle the effects of distillation and fusion, we evaluate the DMD2, DMD2+GAN, and RDM students both before fusion and with the same attention-guided fusion (AGF) module. TurboClear denotes the complete RDM+LSF configuration. Thus, comparisons within the same Fusion column isolate the distillation strategy, whereas RDM with No fusion, AGF, and LSF isolates the fusion strategy.

\begin{table*}[t]
\centering
\scriptsize
\setlength{\tabcolsep}{2.1pt}
\renewcommand{\arraystretch}{1.05}
\begin{tabular}{lllccccccc}
\toprule
Type & Method & Fusion & PSNR-BG$\uparrow$ & DISTS$\downarrow$ & DISTS-L$\downarrow$
& MUSIQ$\uparrow$ & CLIP-IQA$\uparrow$ & CFD$\downarrow$ & FID$_{163}^{*}\downarrow$ \\
\midrule
\multirow{2}{*}{\shortstack[l]{Multi-step\\methods}}
& ObjectClear & AGF & \underline{35.54} & \underline{0.02172} & 0.1426
& 67.23 & \underline{0.5135} & 0.2644 & \textbf{19.04} \\
& OmniPaint & N/A & 30.01 & 0.04111 & \textbf{0.1298}
& \textbf{67.85} & 0.4680 & 0.2462 & 23.30 \\
\midrule
\multirow{7}{*}{\shortstack[l]{Single-step\\variants}}
& DMD2 & No & 28.47 & 0.06340 & 0.1621 & 66.65 & 0.4710 & 0.2215 & 34.34 \\
& DMD2 + GAN & No & 28.65 & 0.05940 & 0.1503 & 66.90 & 0.4692 & 0.2442 & 32.27 \\
& RDM & No & 29.18 & 0.05790 & 0.1513 & 66.88 & 0.4666 & \textbf{0.1948} & 29.84 \\
\cmidrule(lr){2-10}
& DMD2 & AGF & 33.37 & 0.03612 & 0.1658 & 66.57 & 0.4961 & 0.2365 & 36.49 \\
& DMD2 + GAN & AGF & 33.24 & 0.03584 & 0.1586 & 66.70 & 0.4992 & 0.2591 & 36.93 \\
& RDM & AGF & 33.47 & 0.03313 & 0.1563 & 66.76 & 0.5042 & \underline{0.2214} & 36.33 \\
\cmidrule(lr){2-10}
& TurboClear (RDM) & LSF & \textbf{38.93} & \textbf{0.01677} & \underline{0.1371}
& \underline{67.40} & \textbf{0.5156} & 0.2222 & \underline{19.16} \\
\bottomrule
\end{tabular}
\caption{Complementary quality evaluation on the 163-image OBER-Test set. Fusion is listed separately: No denotes no auxiliary fusion, AGF denotes attention-guided fusion, and LSF denotes our learned spatial fusion. ObjectClear uses AGF by default; N/A indicates that this categorization is not applicable to OmniPaint's different FLUX-based architecture, which is evaluated with its default configuration. DISTS-L uses the object-mask bounding-box crop, while PSNR-BG evaluates the object-mask complement. MUSIQ, CLIP-IQA, and CFD are no-reference metrics; $^{*}$FID$_{163}$ is an auxiliary small-sample statistic. Best and second-best values across all listed configurations are highlighted in bold and underlined.}
\label{tab:additional_quality}
\end{table*}

\begin{table*}[t]
\centering
{\scriptsize
\setlength{\tabcolsep}{2.1pt}
\renewcommand{\arraystretch}{1.14}
\begin{tabular}{lcccccc}
\toprule
Paired comparison & PSNR-BG & DISTS & DISTS-L & MUSIQ & CLIP-IQA & CFD \\
\midrule
TurboClear vs. ObjectClear
& \shortstack{+3.392\\{[3.016, 3.783]}}
& \shortstack{+0.004953\\{[0.003576, 0.006000]}}
& \shortstack{+0.005487\\{[0.002928, 0.008075]}}
& \shortstack{+0.1694\\{[0.04663, 0.2982]}}
& \shortstack{+0.002084\\{[-0.001383, 0.005719]}}
& \shortstack{+0.04219\\{[0.009484, 0.07737]}} \\
TurboClear vs. OmniPaint
& \shortstack{+8.925\\{[8.167, 9.716]}}
& \shortstack{+0.02434\\{[0.02178, 0.02746]}}
& \shortstack{-0.007328\\{[-0.01332, -0.0008963]}}
& \shortstack{-0.4549\\{[-0.6882, -0.2354]}}
& \shortstack{+0.04758\\{[0.03831, 0.05649]}}
& \shortstack{+0.02403\\{[-0.008607, 0.05817]}} \\
TurboClear vs. DMD2 + AGF
& \shortstack{+5.560\\{[5.087, 6.044]}}
& \shortstack{+0.01935\\{[0.01631, 0.02264]}}
& \shortstack{+0.02864\\{[0.02511, 0.03233]}}
& \shortstack{+0.8279\\{[0.5880, 1.084]}}
& \shortstack{+0.01950\\{[0.01416, 0.02490]}}
& \shortstack{+0.01432\\{[-0.01447, 0.04479]}} \\
TurboClear vs. DMD2 + GAN + AGF
& \shortstack{+5.689\\{[5.192, 6.203]}}
& \shortstack{+0.01907\\{[0.01576, 0.02268]}}
& \shortstack{+0.02146\\{[0.01823, 0.02492]}}
& \shortstack{+0.6971\\{[0.4374, 0.9824]}}
& \shortstack{+0.01634\\{[0.01060, 0.02246]}}
& \shortstack{+0.03692\\{[0.009892, 0.06660]}} \\
\midrule
RDM vs. DMD2 (No fusion)
& \shortstack{+0.7082\\{[0.6161, 0.8050]}}
& \shortstack{+0.005508\\{[0.004018, 0.007074]}}
& \shortstack{+0.01080\\{[0.008432, 0.01313]}}
& \shortstack{+0.2353\\{[0.09500, 0.3765]}}
& \shortstack{-0.004423\\{[-0.009883, 0.0009757]}}
& \shortstack{+0.02673\\{[-0.0002980, 0.05533]}} \\
RDM vs. DMD2 + GAN (No fusion)
& \shortstack{+0.5348\\{[0.4661, 0.6103]}}
& \shortstack{+0.001507\\{[0.0004251, 0.002604]}}
& \shortstack{-0.001044\\{[-0.002981, 0.0008304]}}
& \shortstack{-0.01518\\{[-0.1495, 0.1201]}}
& \shortstack{-0.002646\\{[-0.007385, 0.002040]}}
& \shortstack{+0.04941\\{[0.01991, 0.08152]}} \\
RDM vs. DMD2 (AGF)
& \shortstack{+0.09558\\{[-0.02972, 0.2146]}}
& \shortstack{+0.002994\\{[0.002026, 0.003968]}}
& \shortstack{+0.009485\\{[0.007412, 0.01156]}}
& \shortstack{+0.1859\\{[0.04980, 0.3285]}}
& \shortstack{+0.008101\\{[0.004903, 0.01157]}}
& \shortstack{+0.01506\\{[-0.01496, 0.04647]}} \\
RDM vs. DMD2 + GAN (AGF)
& \shortstack{+0.2244\\{[0.1582, 0.2888]}}
& \shortstack{+0.002714\\{[0.001957, 0.003501]}}
& \shortstack{+0.002306\\{[0.0004796, 0.004108]}}
& \shortstack{+0.05505\\{[-0.05902, 0.1723]}}
& \shortstack{+0.004942\\{[0.001592, 0.008446]}}
& \shortstack{+0.03767\\{[0.001419, 0.07489]}} \\
RDM (LSF) vs. RDM (AGF)
& \shortstack{+5.464\\{[4.959, 5.979]}}
& \shortstack{+0.01636\\{[0.01324, 0.01983]}}
& \shortstack{+0.01916\\{[0.01609, 0.02236]}}
& \shortstack{+0.6420\\{[0.4065, 0.8928]}}
& \shortstack{+0.01140\\{[0.007000, 0.01595]}}
& \shortstack{-0.0007436\\{[-0.03266, 0.03125]}} \\
\bottomrule
\end{tabular}
}
\caption{Paired favorable mean differences with 95\% bootstrap CIs in brackets. Positive values favor the first configuration in each comparison after accounting for the metric direction; an interval containing zero is not conclusive. The upper block compares the complete TurboClear pipeline with multi-step methods and competitive AGF-based one-step variants. The lower block holds fusion fixed to isolate distillation, or holds RDM fixed to isolate fusion.}
\label{tab:additional_quality_ci}
\end{table*}

\paragraph{Results.}
Table~\ref{tab:additional_quality} shows that TurboClear achieves the best PSNR-BG, global DISTS, and CLIP-IQA, while ranking second on DISTS-Local, MUSIQ, and the auxiliary FID$_{163}$. The unfused RDM student obtains the lowest CFD, but at substantially lower background-fidelity and perceptual scores. RDM+AGF and RDM+LSF have nearly identical CFD, and their paired interval in Table~\ref{tab:additional_quality_ci} contains zero.

The controlled comparisons in Table~\ref{tab:additional_quality_ci} separate the two contributions. Without fusion, RDM improves PSNR-BG and global DISTS over both DMD2 variants; its advantages over DMD2 also extend to DISTS-Local and MUSIQ, while its advantage over DMD2+GAN extends to CFD. Under the shared AGF setting, RDM improves global DISTS, DISTS-Local, and CLIP-IQA over both alternatives. It additionally improves MUSIQ over DMD2, and PSNR-BG and CFD over DMD2+GAN; the remaining intervals contain zero. Most importantly, with RDM fixed, LSF improves PSNR-BG, DISTS, DISTS-Local, MUSIQ, and CLIP-IQA over AGF, while maintaining comparable CFD. Compared with DMD2+GAN under the same AGF setting, all six paired intervals favor the complete TurboClear pipeline. These complementary results support both region-calibrated distillation and learned spatial fusion without relying on a single reconstruction metric.

\section{Analysis for LSF}

We formalize why the optimal fusion gate cannot, in general, be recovered from attention alone. At a pixel $p$ (omitted below), let $d=\hat{x}-y$ be the difference between the generated and reference streams and $r=x^\star-y$ the desired correction. The local mixing risk is
\begin{equation}
\ell(\alpha)=\|\alpha\hat{x}+(1-\alpha)y-x^\star\|_2^2
=\|\alpha d-r\|_2^2,\qquad \alpha\in[0,1].
\end{equation}
For $d\neq0$, projection of the unconstrained minimizer onto the feasible interval gives the unique optimum
\begin{equation}
\alpha^\star
=\Pi_{[0,1]}\!\left(\frac{\langle d,r\rangle}{\|d\|_2^2}\right).
\end{equation}
Hence, an attention map can determine the optimal gate only if it is a sufficient statistic for the local relation among $\hat{x}$, $y$, and $x^\star$. A one-step attention map need not satisfy this condition.

The consequence of a gating error is quantitative. By expanding the quadratic and using the first-order optimality condition $(\alpha-\alpha^\star)\ell'(\alpha^\star)\geq0$ for the constrained optimum,
\begin{equation}
\begin{aligned}
\ell(\alpha)-\ell(\alpha^\star)
&=\|d\|_2^2(\alpha-\alpha^\star)^2 \\
&\quad +(\alpha-\alpha^\star)\ell'(\alpha^\star) \\
&\geq\|d\|_2^2(\alpha-\alpha^\star)^2.
\end{aligned}
\end{equation}
Thus, an undersized gate in an affected region copies reference content back and can reintroduce removed effects, whereas an oversized gate in the background causes unnecessary changes.

Finally, let $A$ denote the attention prior and let $Z$ collect the richer local features used by LSF, including $A$, both streams, their difference, and the object mask. Let $\mathcal{G}_A$ be the class of attention-only gates and $\mathcal{H}_Z$ the LSF gate class. Since $\mathcal{H}_Z$ contains every lifted attention-only rule $Z\mapsto g(A)$, its optimal population risk satisfies
\begin{equation}
\mathcal{R}_{\rm LSF}^\star
=\inf_{h\in\mathcal{H}_Z}\mathbb{E}[\ell(h(Z))]
\leq
\inf_{g\in\mathcal{G}_A}\mathbb{E}[\ell(g(A))]
=\mathcal{R}_{\rm AGF}^\star.
\end{equation}
The inequality is strict whenever $\alpha^\star$ is not measurable from $A$ alone, the additional features in $Z$ resolve part of this ambiguity, and $\|d\|_2>0$ on a set of nonzero probability; the excess-risk bound above then prevents an attention-only gate from attaining the LSF optimum. This illustrates a function-class advantage for LSF, while its asymmetric supervision drives the learned gate toward that better attainable solution.

\section{User Study}

We conduct an anonymous, double-blind user study to compare the perceptual removal quality of TurboClear and its teacher, ObjectClear. Twenty participants each evaluate 30 randomly sampled pairs, comprising 15 samples from OBER-Test and 15 from RORD-Val, for a total of 600 pairwise judgments. Each trial displays the masked input, where the target object is highlighted in green, together with two removal results labeled only as Method A and Method B. Method identities are concealed from both the participants and the researchers administering the study, and the server independently randomizes the left--right assignment for every trial. Participants select the result that appears more natural with fewer object remnants, artifacts, and background-structure errors, or indicate that the two results are similar.

\begin{table}[t]
\centering
\small
\setlength{\tabcolsep}{3.2pt}
\begin{tabular}{lrrrr}
\toprule
Dataset & Better & Similar & Worse & Sim./Better \\
\midrule
OBER-Test & 20.7\% & 53.3\% & 26.0\% & \textbf{74.0\%} \\
RORD-Val & 23.3\% & 40.7\% & 36.0\% & \textbf{64.0\%} \\
Overall & 22.0\% & 47.0\% & 31.0\% & \textbf{69.0\%} \\
\bottomrule
\end{tabular}
\caption{Anonymous user-study results comparing TurboClear with ObjectClear. Better, Similar, and Worse are reported from the perspective of TurboClear.}
\label{tab:user_study}
\end{table}

As shown in Table~\ref{tab:user_study}, despite its substantial acceleration, TurboClear matches or outperforms ObjectClear in 69.0\% of the pairwise evaluations, including 74.0\% on OBER-Test and 64.0\% on RORD-Val. These results indicate that TurboClear preserves comparable perceptual removal quality in the majority of evaluations.

\section{More Results}

As shown in Fig.~\ref{fig:more_results_a} and Fig.~\ref{fig:more_results_b}, we provide additional qualitative comparisons on OBER-Wild, which has no ground-truth targets. The method columns follow the order used in the main paper, and the labels are placed below each image grid.

\newcommand{\suppquallabels}{%
\makebox[0.089\textwidth][c]{\scriptsize Input} &
\makebox[0.089\textwidth][c]{\scriptsize PowerPaint} &
\makebox[0.089\textwidth][c]{\scriptsize DesignEdit} &
\makebox[0.089\textwidth][c]{\scriptsize CLIPAway} &
\makebox[0.089\textwidth][c]{\scriptsize OmniEraser} &
\makebox[0.089\textwidth][c]{\scriptsize\shortstack{Attentive\\ Eraser}} &
\makebox[0.089\textwidth][c]{\scriptsize RORem} &
\makebox[0.089\textwidth][c]{\scriptsize OmniPaint} &
\makebox[0.089\textwidth][c]{\scriptsize ObjectClear} &
\makebox[0.089\textwidth][c]{\scriptsize FlashClear} &
\makebox[0.089\textwidth][c]{\scriptsize\shortstack{TurboClear\\ (ours)}} \\
}

\begin{figure*}[t]
\centering
\setlength{\tabcolsep}{0pt}
\renewcommand{\arraystretch}{1.0}
\begin{tabular}{ccccccccccc}
\includegraphics[width=0.089\textwidth]{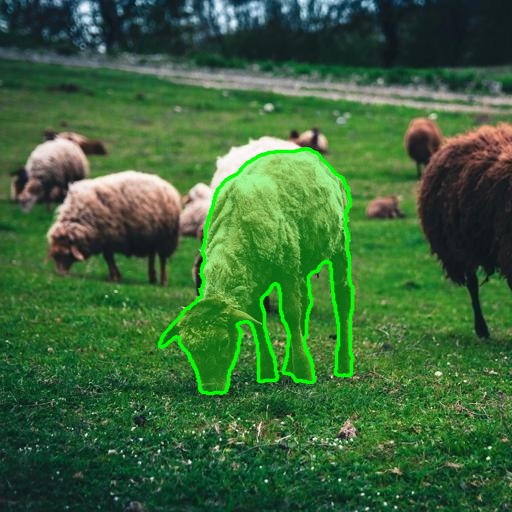} &
\includegraphics[width=0.089\textwidth]{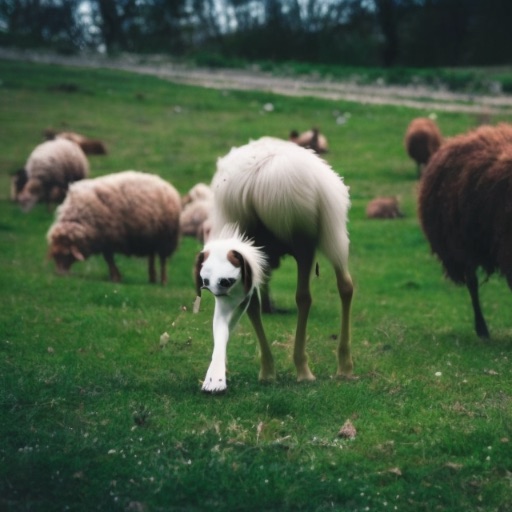} &
\includegraphics[width=0.089\textwidth]{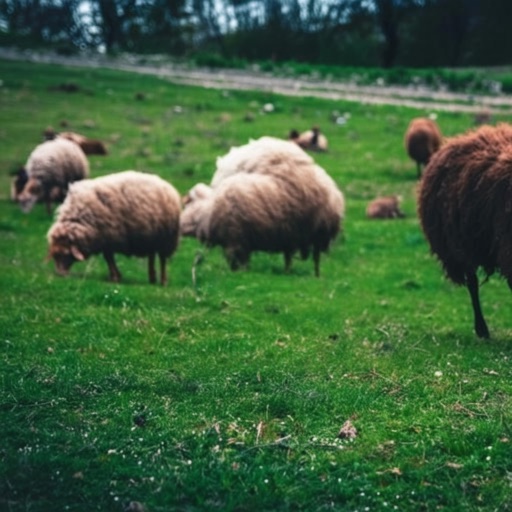} &
\includegraphics[width=0.089\textwidth]{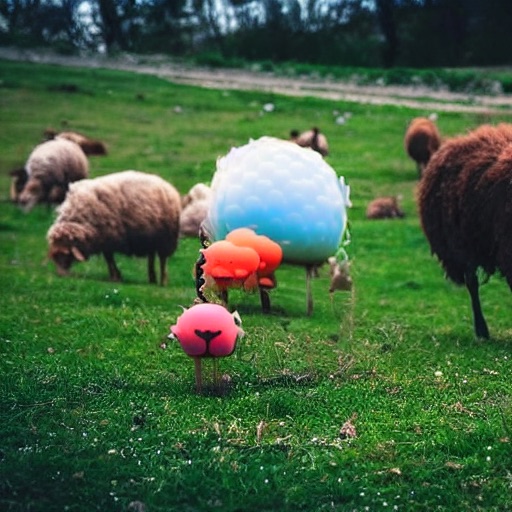} &
\includegraphics[width=0.089\textwidth]{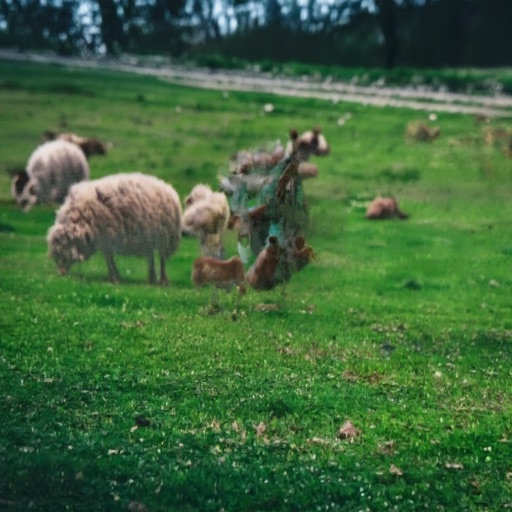} &
\includegraphics[width=0.089\textwidth]{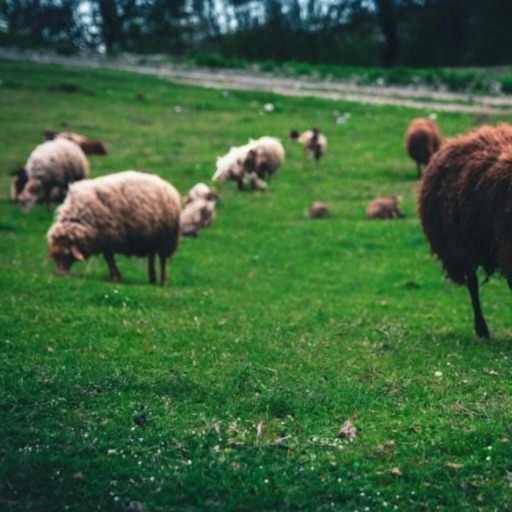} &
\includegraphics[width=0.089\textwidth]{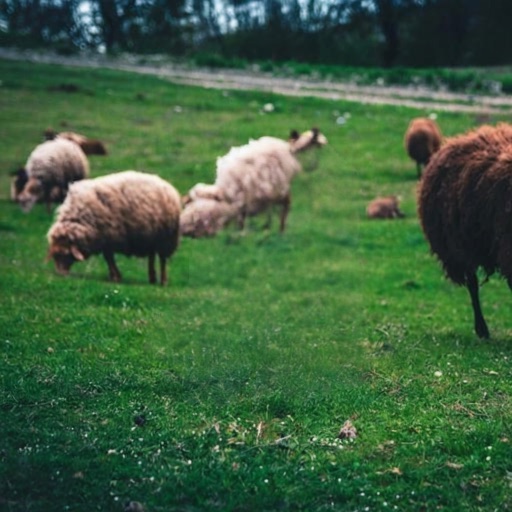} &
\includegraphics[width=0.089\textwidth]{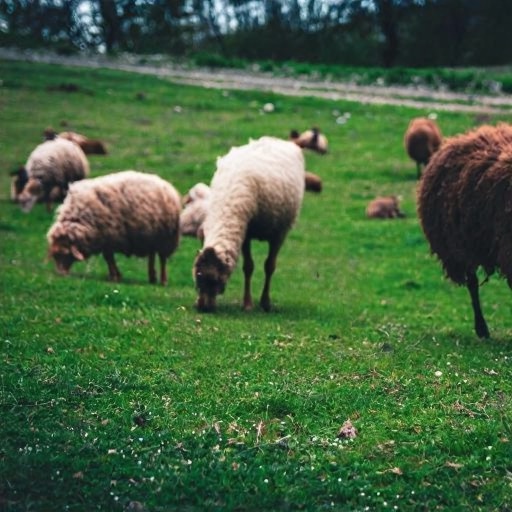} &
\includegraphics[width=0.089\textwidth]{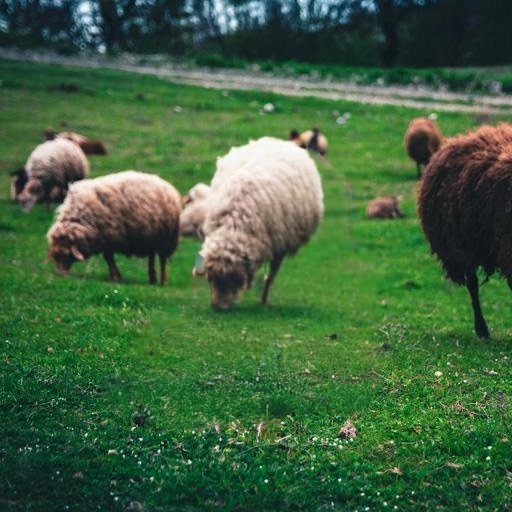} &
\includegraphics[width=0.089\textwidth]{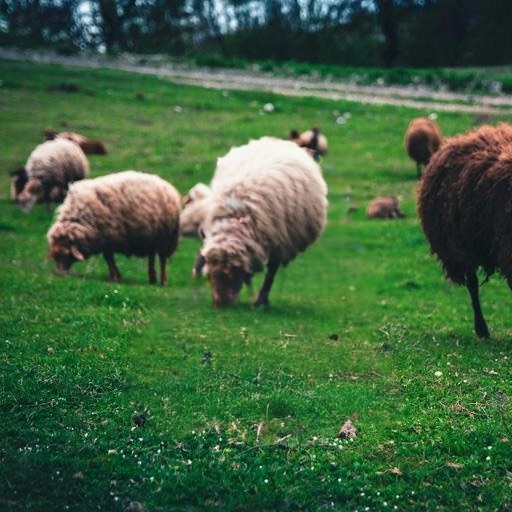} &
\includegraphics[width=0.089\textwidth]{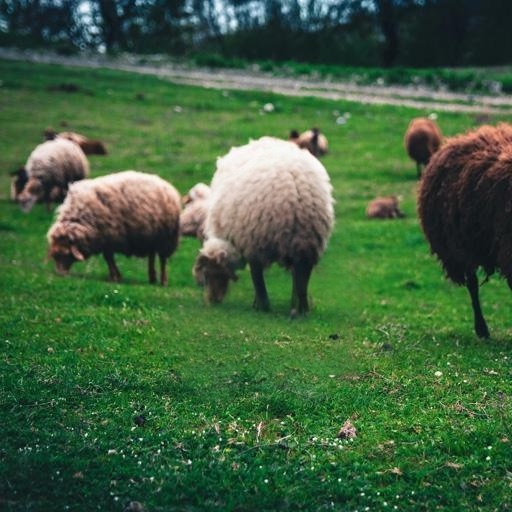} \\
\includegraphics[width=0.089\textwidth]{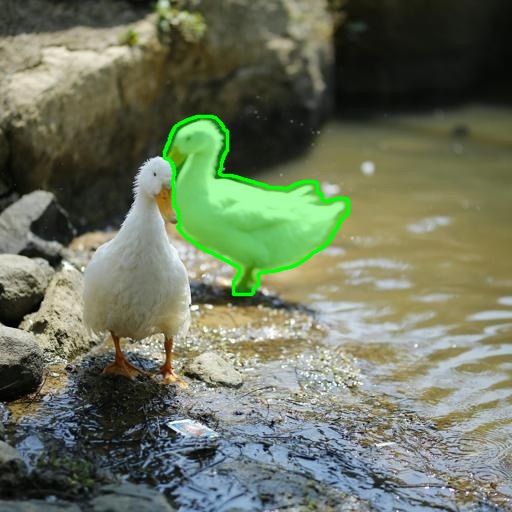} &
\includegraphics[width=0.089\textwidth]{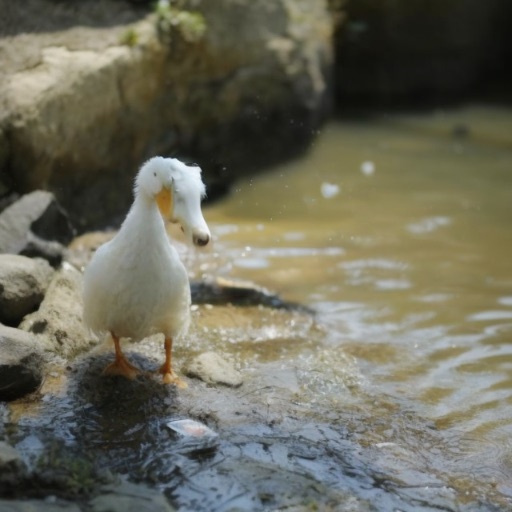} &
\includegraphics[width=0.089\textwidth]{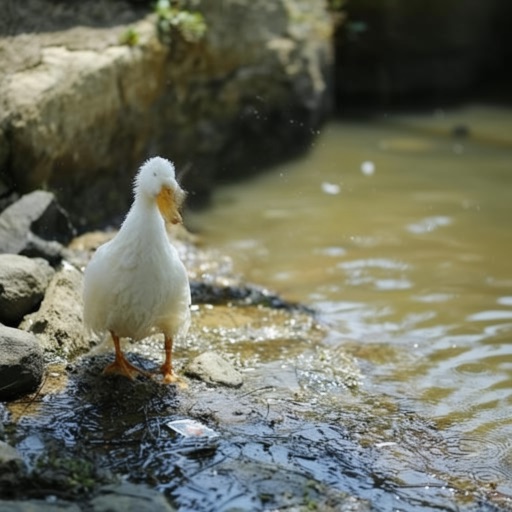} &
\includegraphics[width=0.089\textwidth]{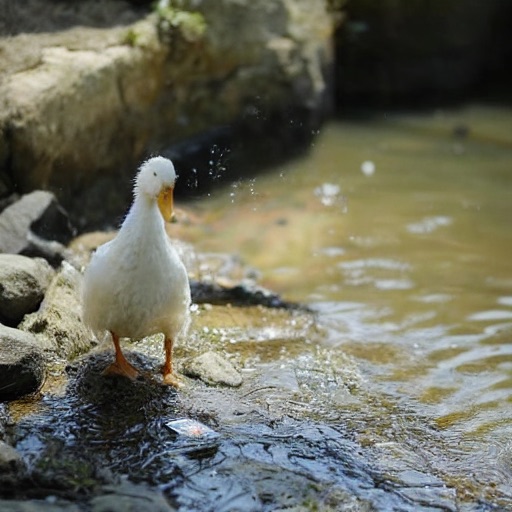} &
\includegraphics[width=0.089\textwidth]{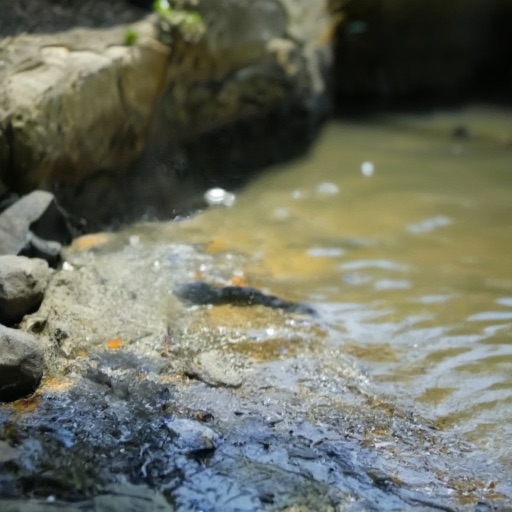} &
\includegraphics[width=0.089\textwidth]{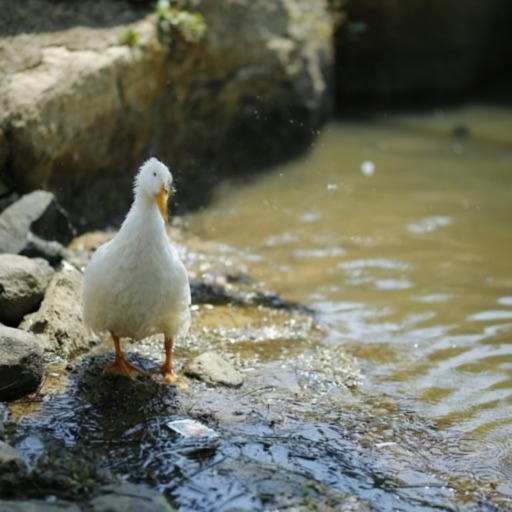} &
\includegraphics[width=0.089\textwidth]{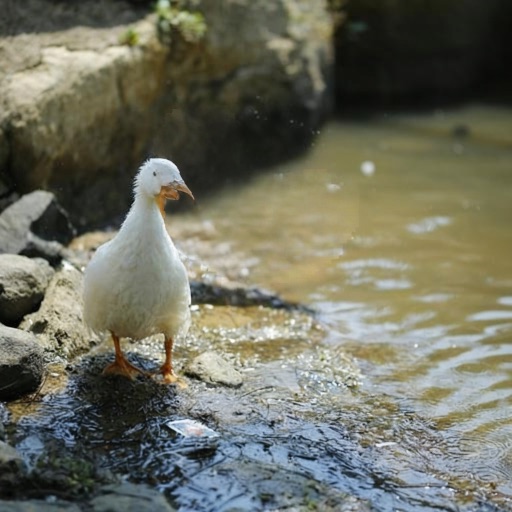} &
\includegraphics[width=0.089\textwidth]{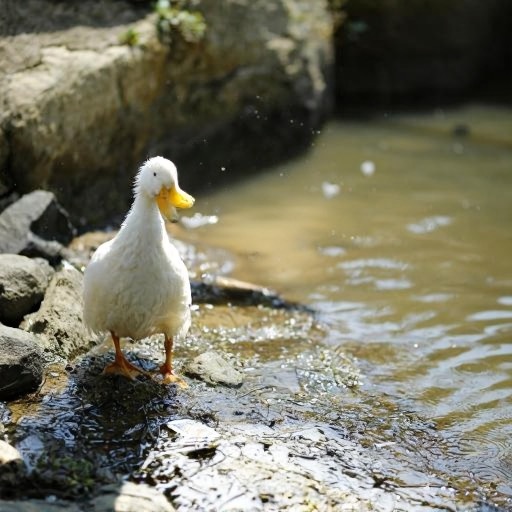} &
\includegraphics[width=0.089\textwidth]{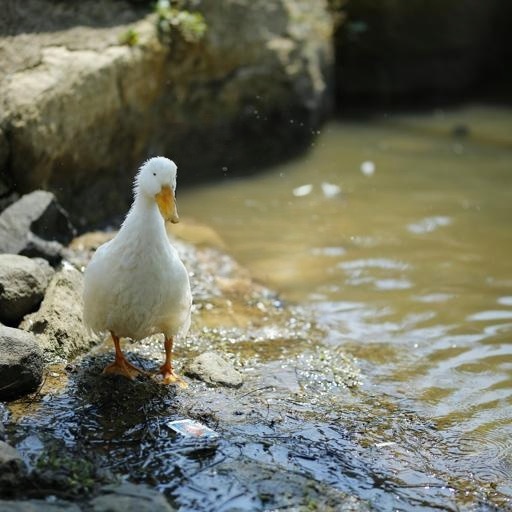} &
\includegraphics[width=0.089\textwidth]{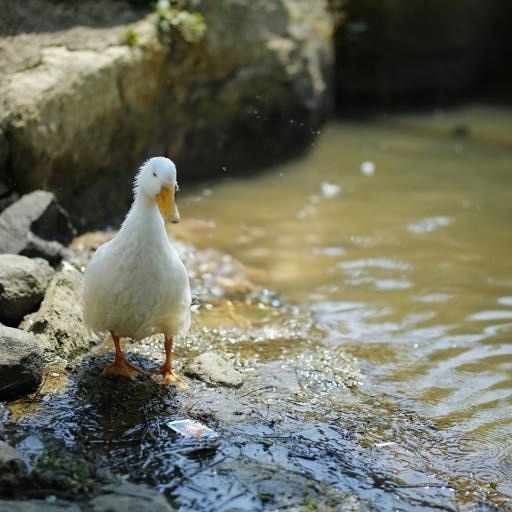} &
\includegraphics[width=0.089\textwidth]{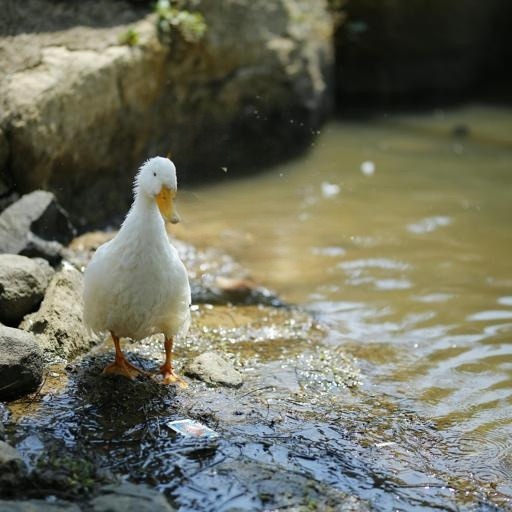} \\
\includegraphics[width=0.089\textwidth]{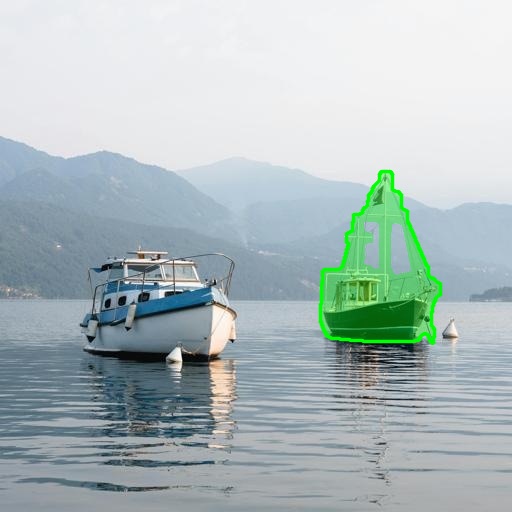} &
\includegraphics[width=0.089\textwidth]{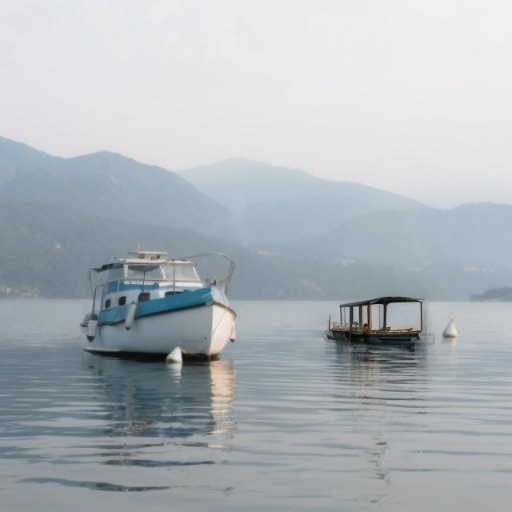} &
\includegraphics[width=0.089\textwidth]{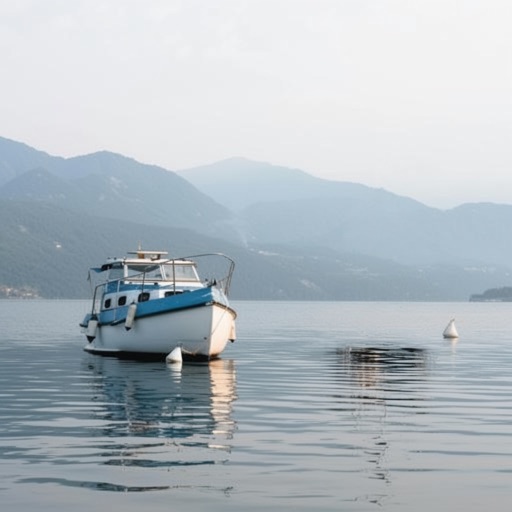} &
\includegraphics[width=0.089\textwidth]{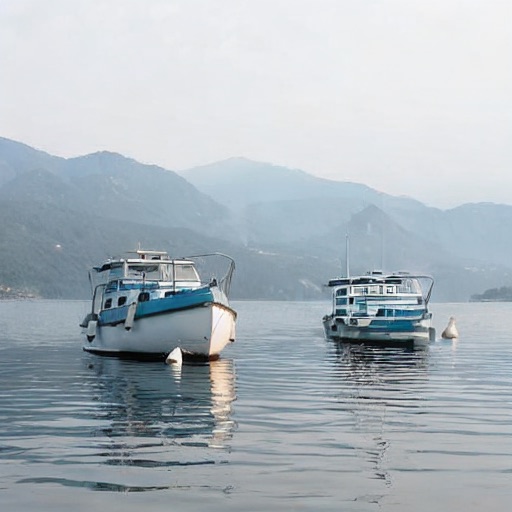} &
\includegraphics[width=0.089\textwidth]{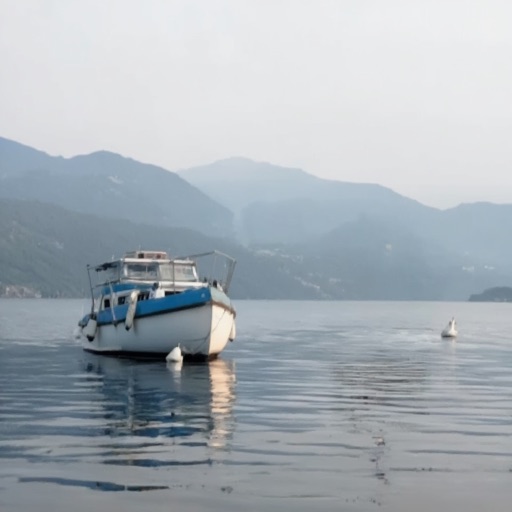} &
\includegraphics[width=0.089\textwidth]{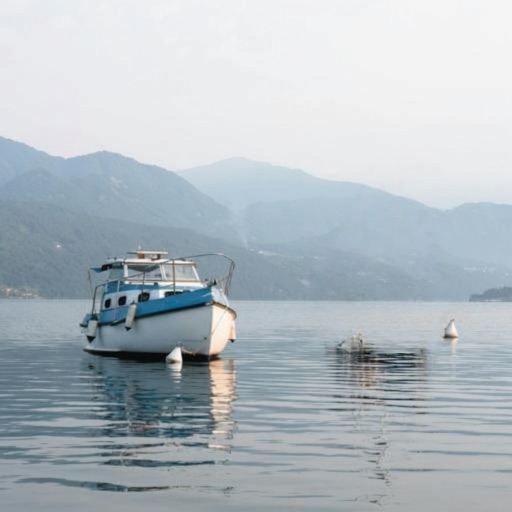} &
\includegraphics[width=0.089\textwidth]{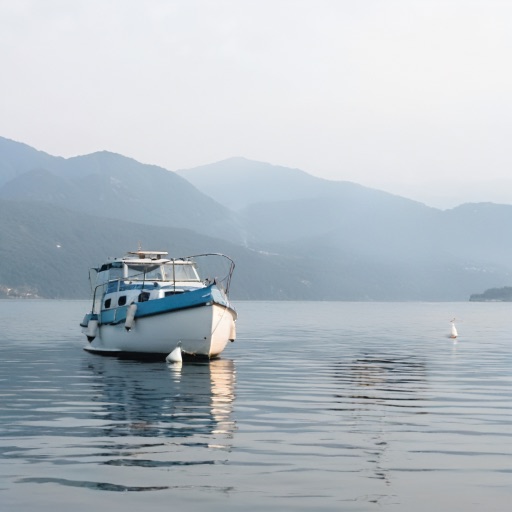} &
\includegraphics[width=0.089\textwidth]{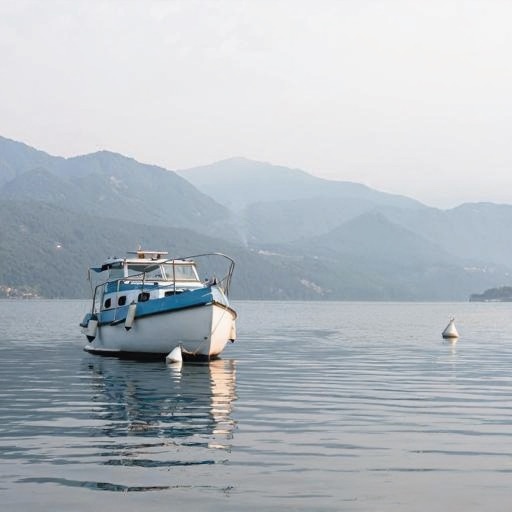} &
\includegraphics[width=0.089\textwidth]{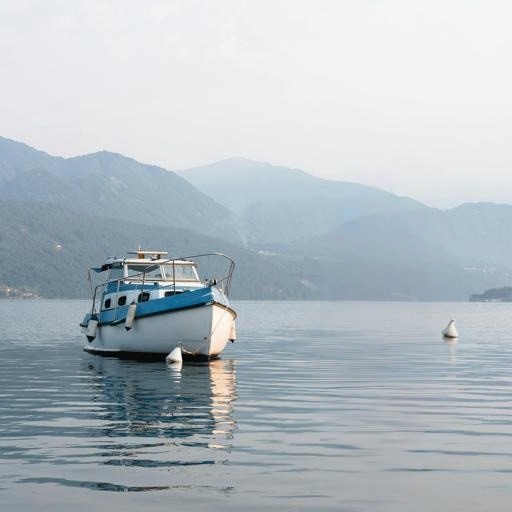} &
\includegraphics[width=0.089\textwidth]{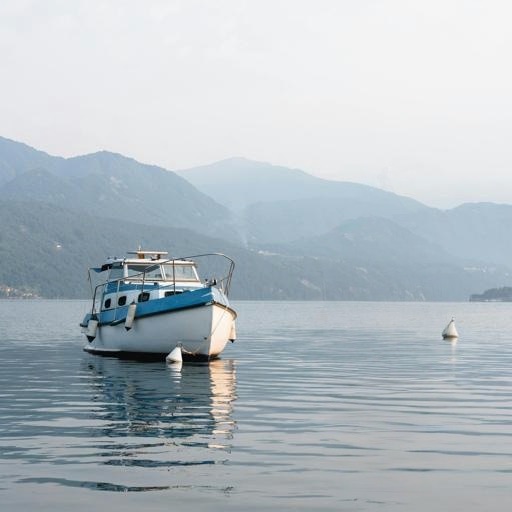} &
\includegraphics[width=0.089\textwidth]{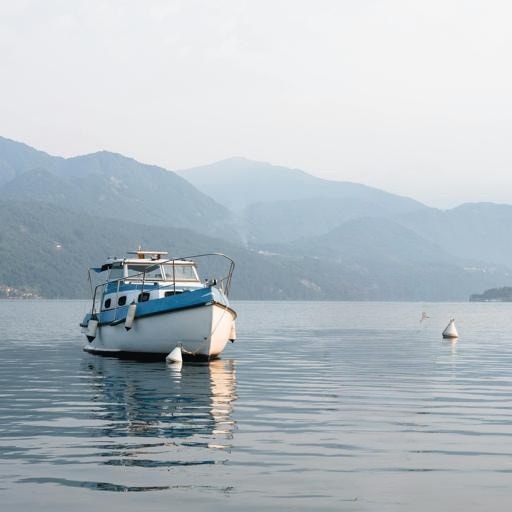} \\
\includegraphics[width=0.089\textwidth]{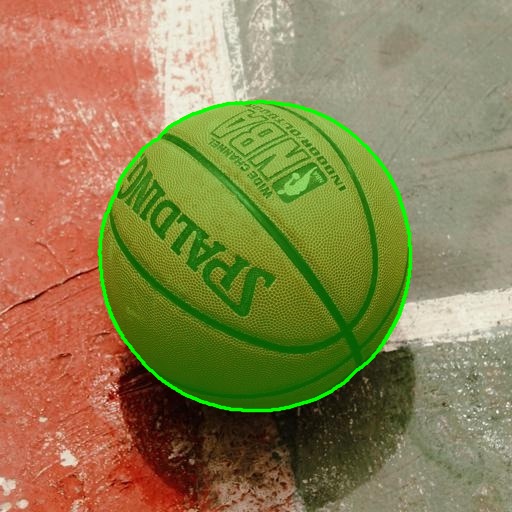} &
\includegraphics[width=0.089\textwidth]{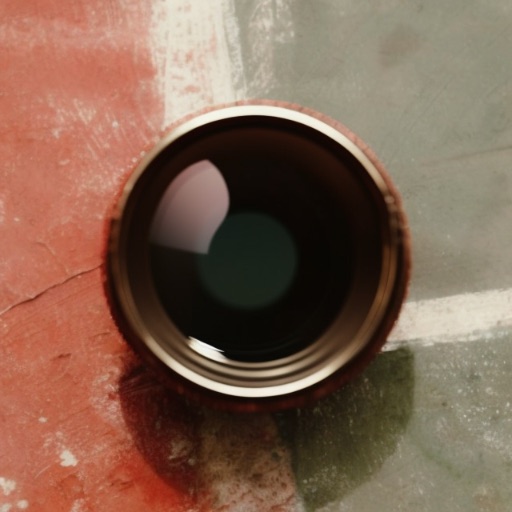} &
\includegraphics[width=0.089\textwidth]{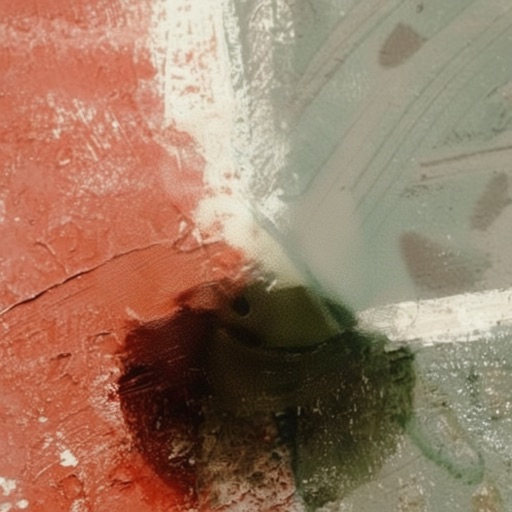} &
\includegraphics[width=0.089\textwidth]{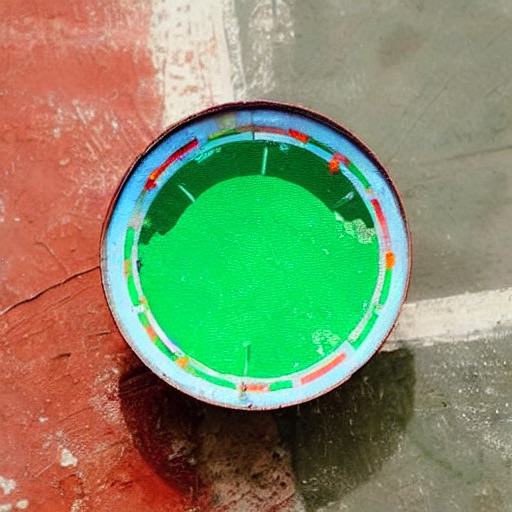} &
\includegraphics[width=0.089\textwidth]{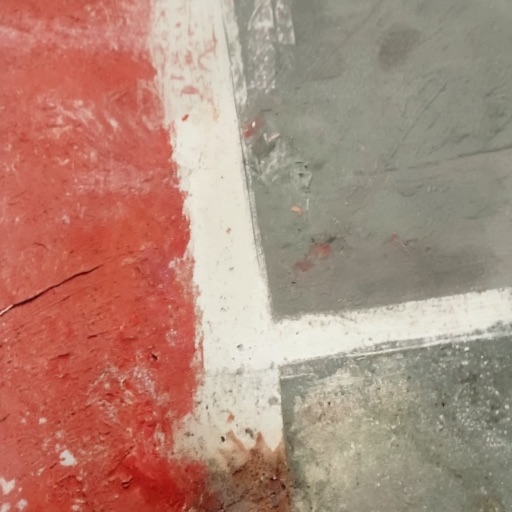} &
\includegraphics[width=0.089\textwidth]{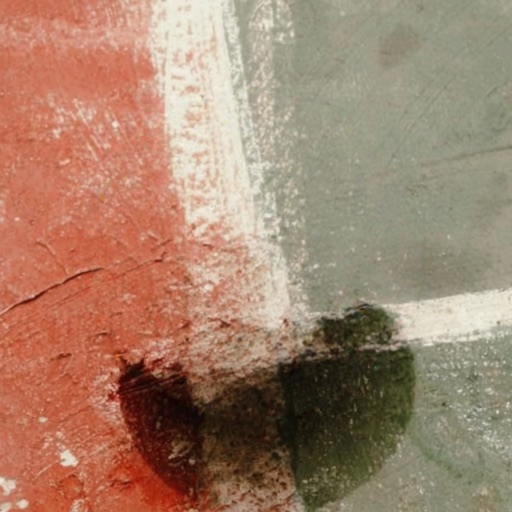} &
\includegraphics[width=0.089\textwidth]{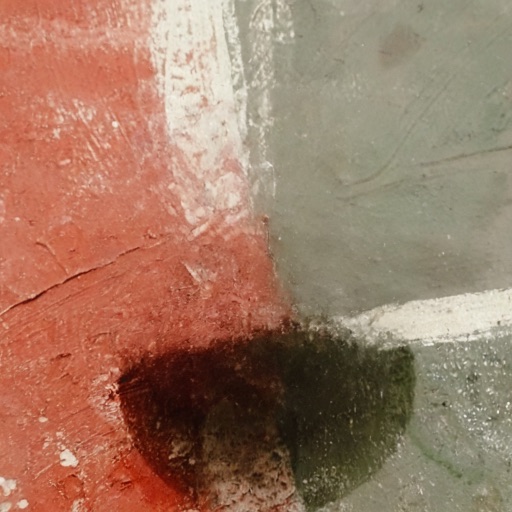} &
\includegraphics[width=0.089\textwidth]{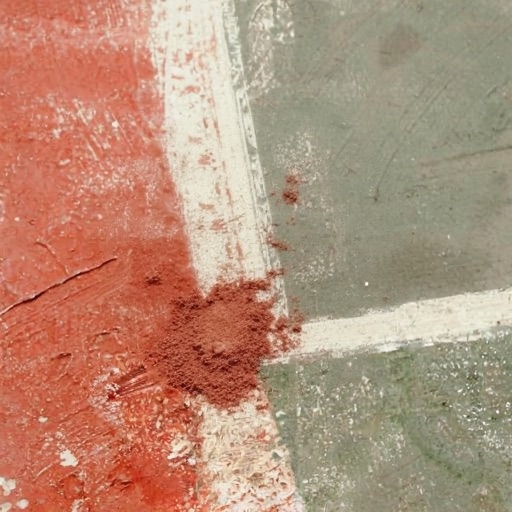} &
\includegraphics[width=0.089\textwidth]{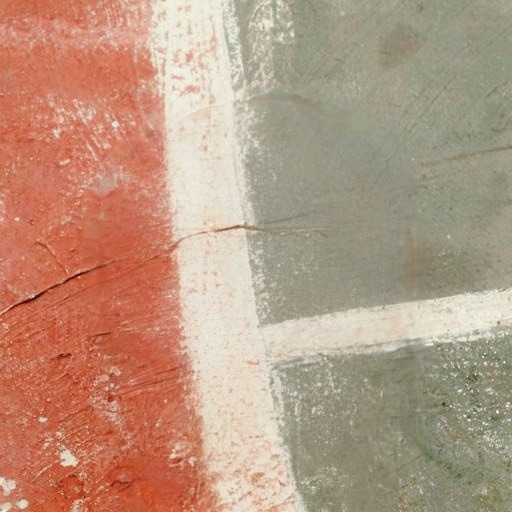} &
\includegraphics[width=0.089\textwidth]{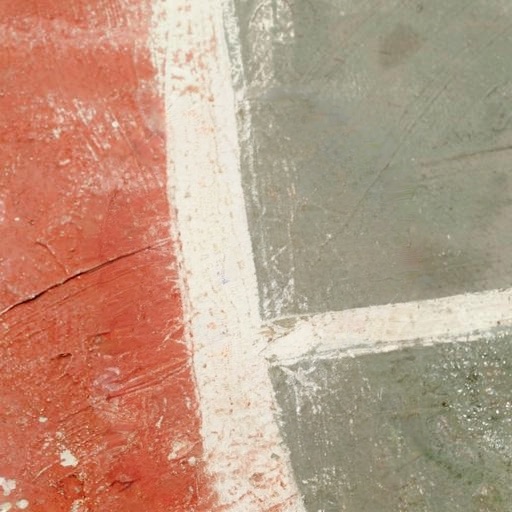} &
\includegraphics[width=0.089\textwidth]{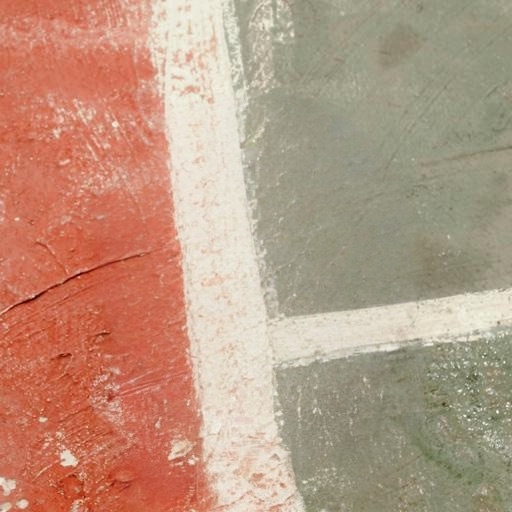} \\
\includegraphics[width=0.089\textwidth]{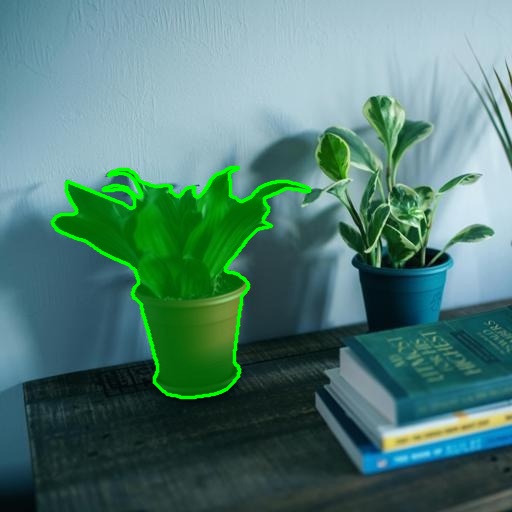} &
\includegraphics[width=0.089\textwidth]{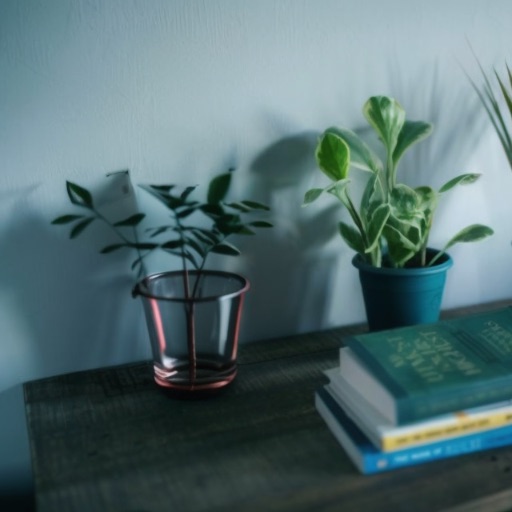} &
\includegraphics[width=0.089\textwidth]{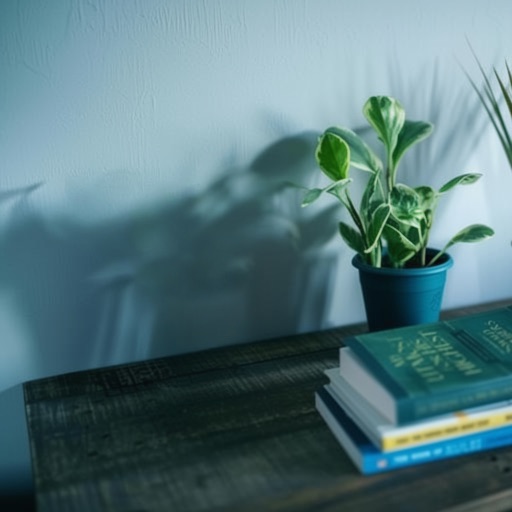} &
\includegraphics[width=0.089\textwidth]{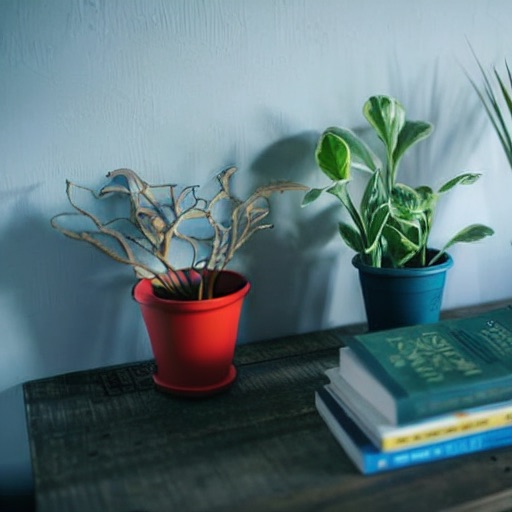} &
\includegraphics[width=0.089\textwidth]{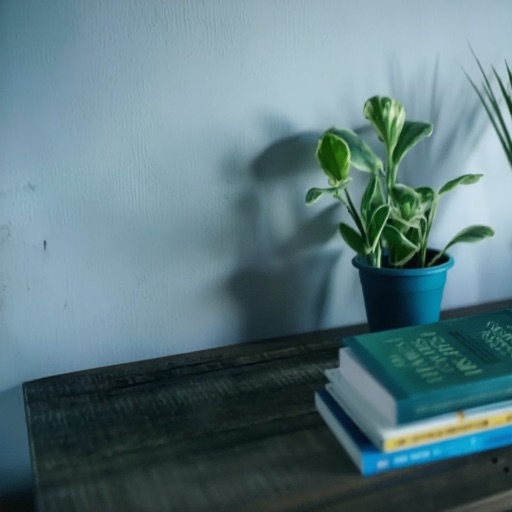} &
\includegraphics[width=0.089\textwidth]{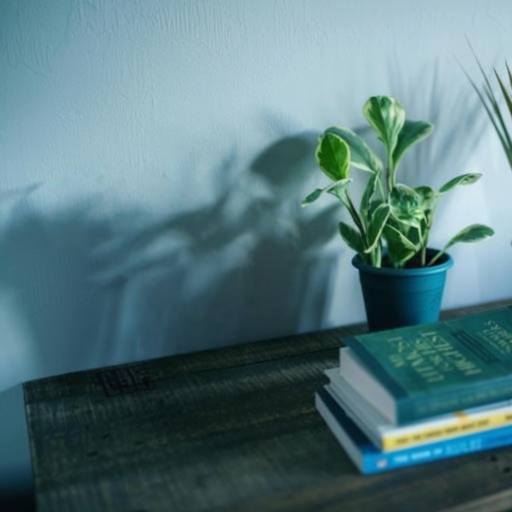} &
\includegraphics[width=0.089\textwidth]{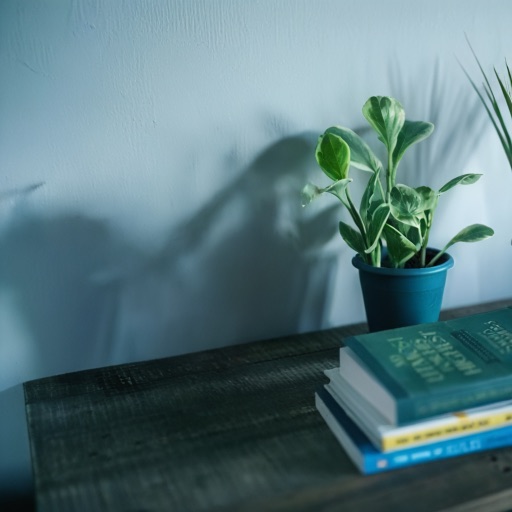} &
\includegraphics[width=0.089\textwidth]{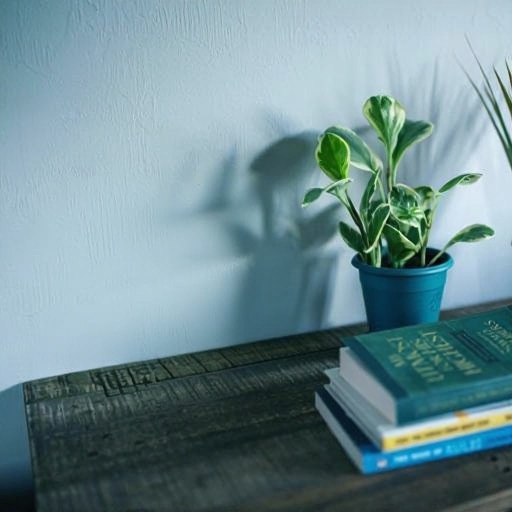} &
\includegraphics[width=0.089\textwidth]{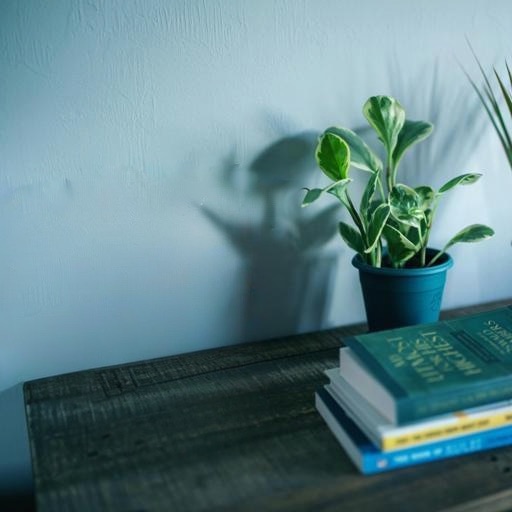} &
\includegraphics[width=0.089\textwidth]{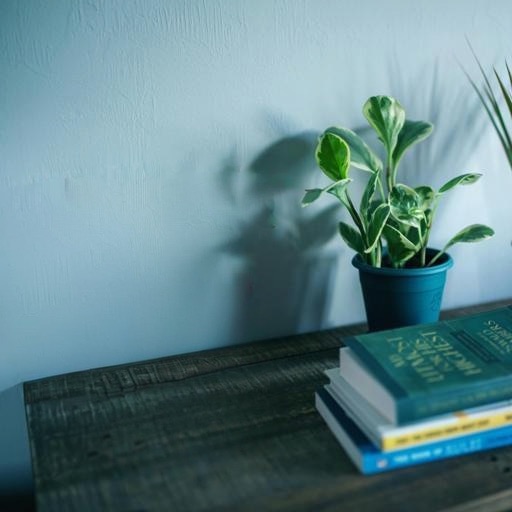} &
\includegraphics[width=0.089\textwidth]{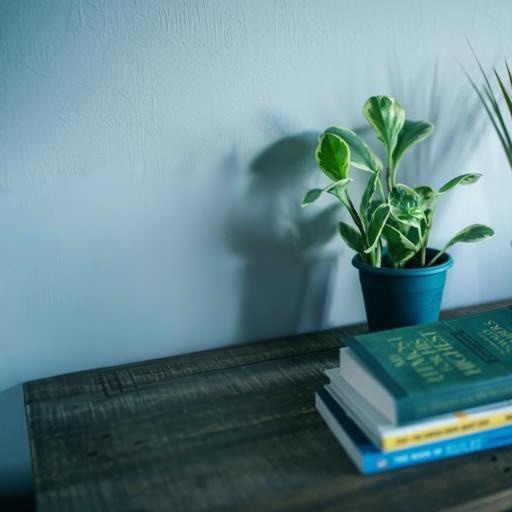} \\
\includegraphics[width=0.089\textwidth]{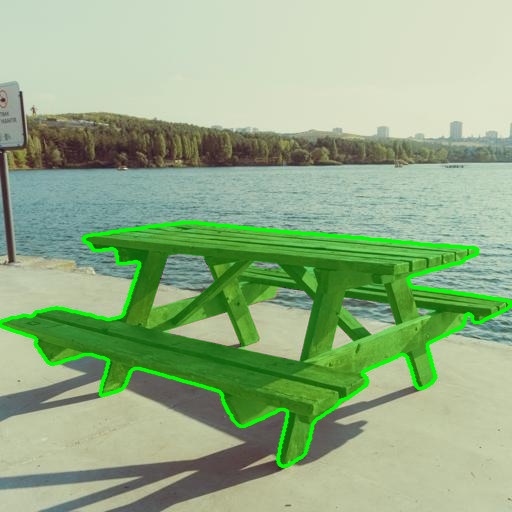} &
\includegraphics[width=0.089\textwidth]{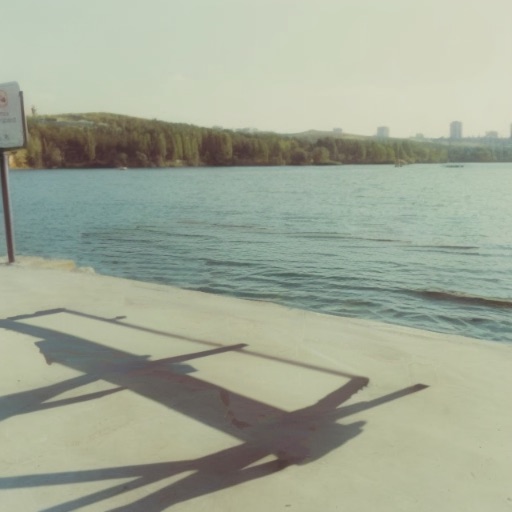} &
\includegraphics[width=0.089\textwidth]{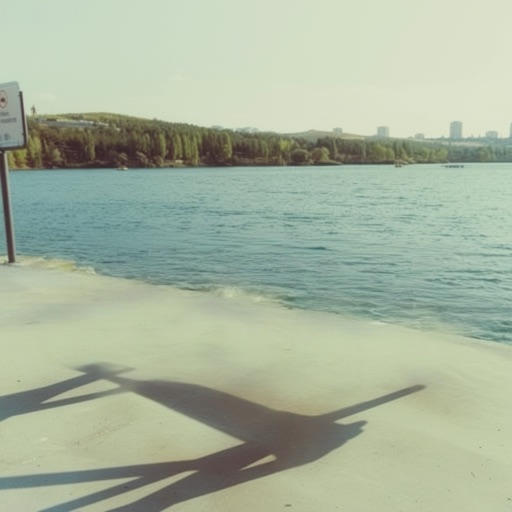} &
\includegraphics[width=0.089\textwidth]{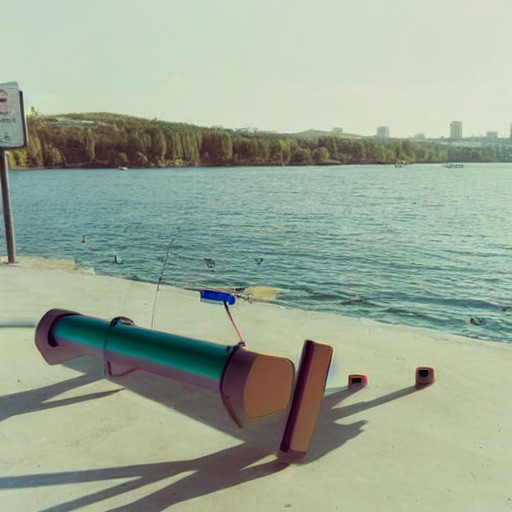} &
\includegraphics[width=0.089\textwidth]{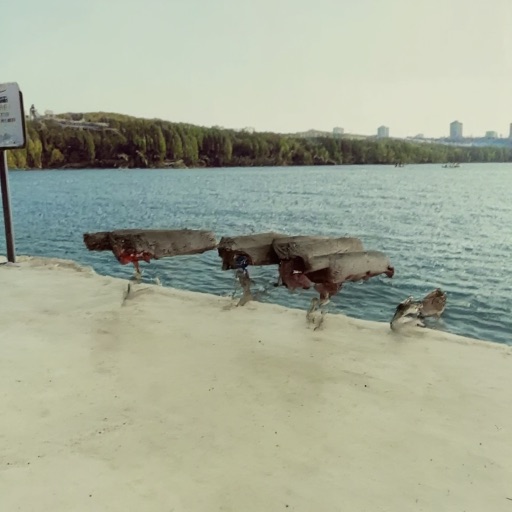} &
\includegraphics[width=0.089\textwidth]{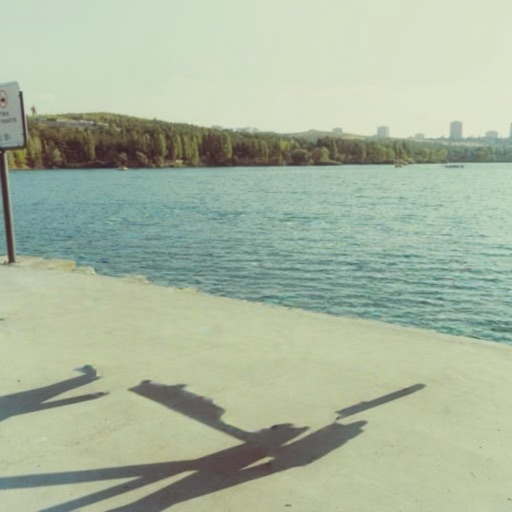} &
\includegraphics[width=0.089\textwidth]{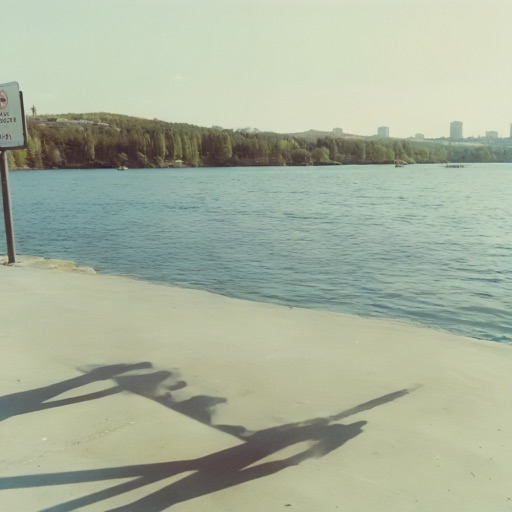} &
\includegraphics[width=0.089\textwidth]{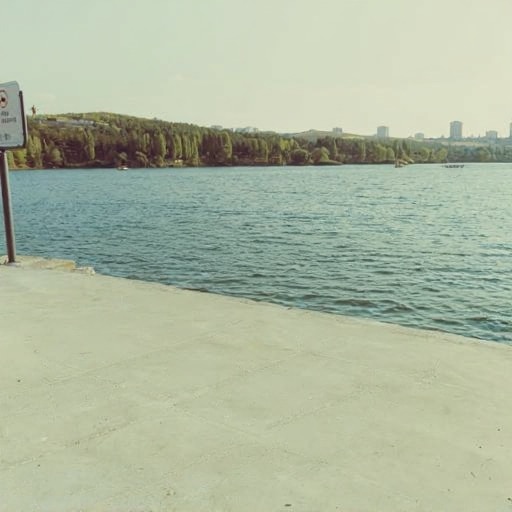} &
\includegraphics[width=0.089\textwidth]{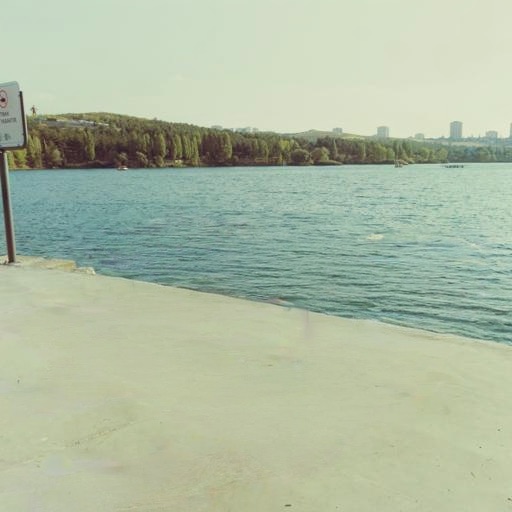} &
\includegraphics[width=0.089\textwidth]{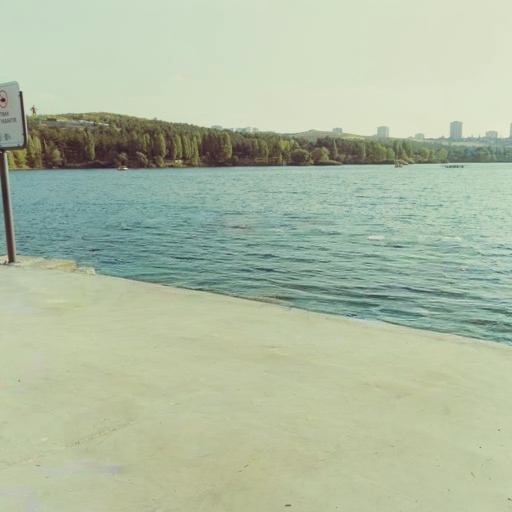} &
\includegraphics[width=0.089\textwidth]{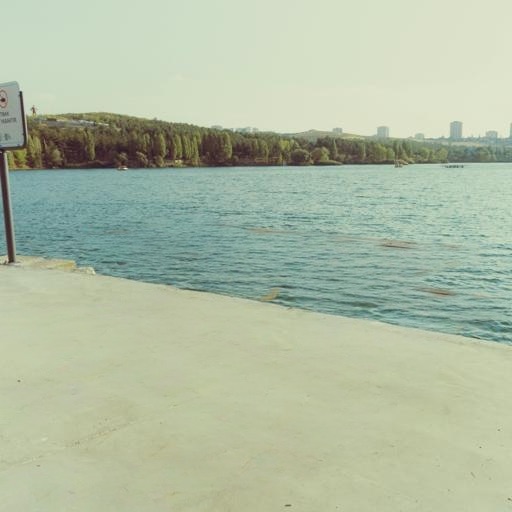} \\
\includegraphics[width=0.089\textwidth]{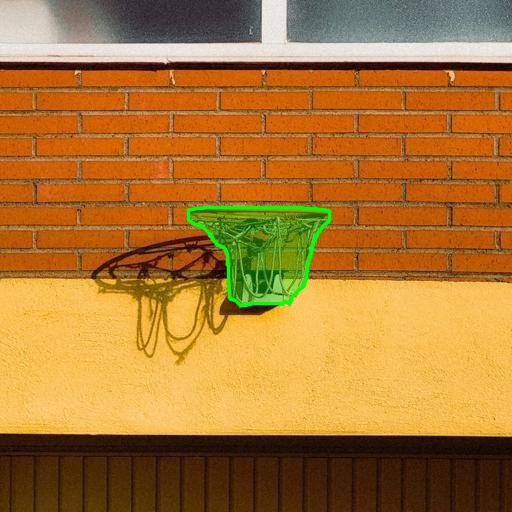} &
\includegraphics[width=0.089\textwidth]{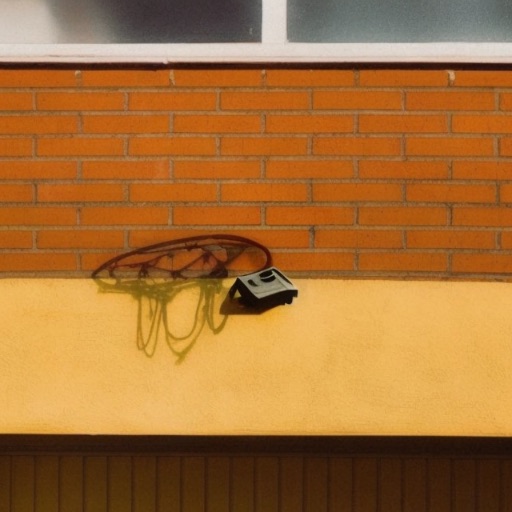} &
\includegraphics[width=0.089\textwidth]{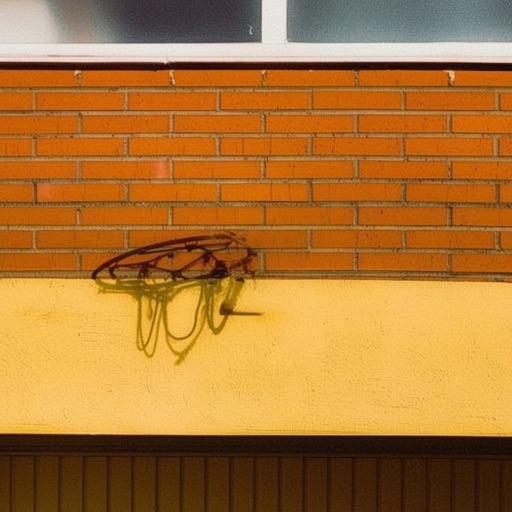} &
\includegraphics[width=0.089\textwidth]{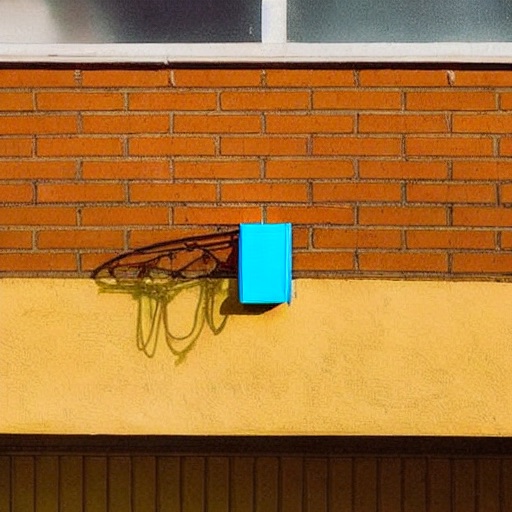} &
\includegraphics[width=0.089\textwidth]{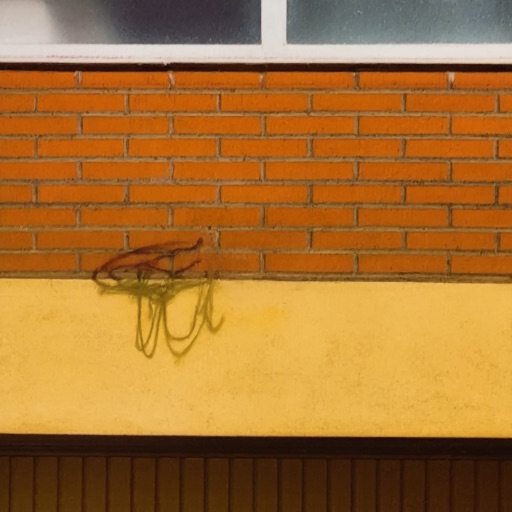} &
\includegraphics[width=0.089\textwidth]{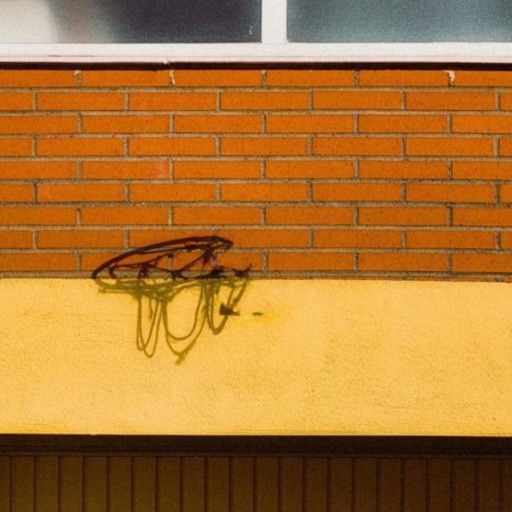} &
\includegraphics[width=0.089\textwidth]{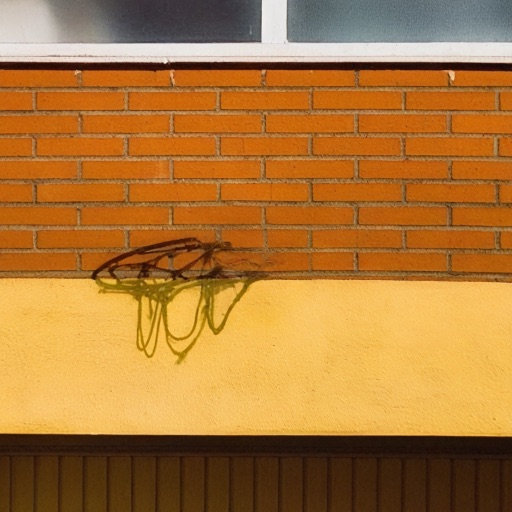} &
\includegraphics[width=0.089\textwidth]{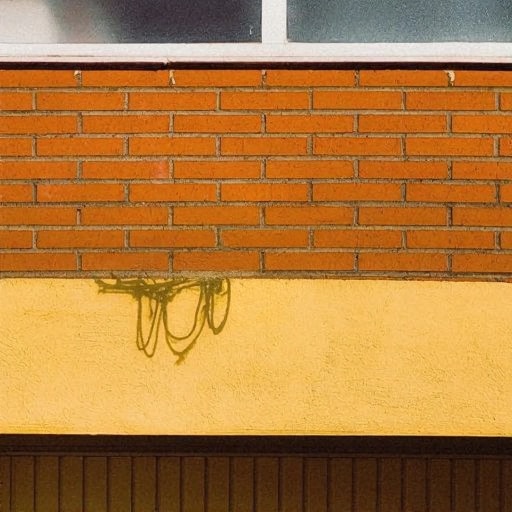} &
\includegraphics[width=0.089\textwidth]{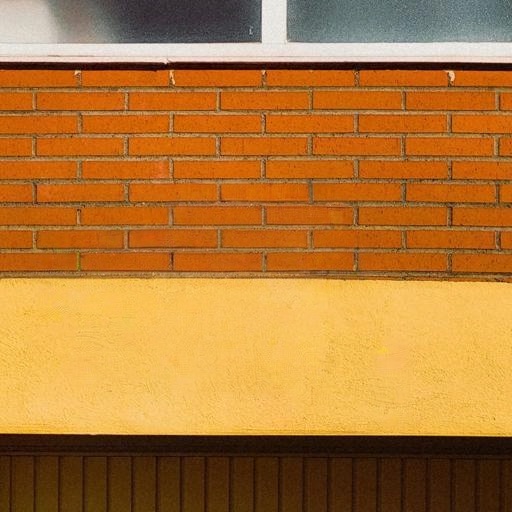} &
\includegraphics[width=0.089\textwidth]{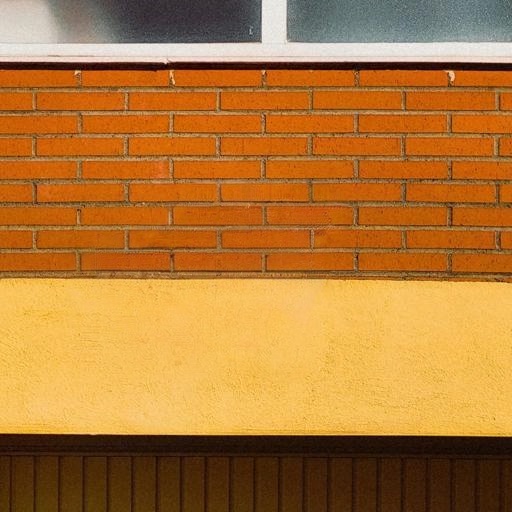} &
\includegraphics[width=0.089\textwidth]{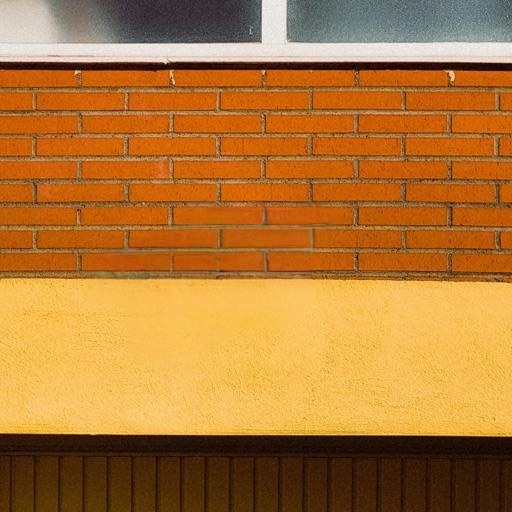} \\
\includegraphics[width=0.089\textwidth]{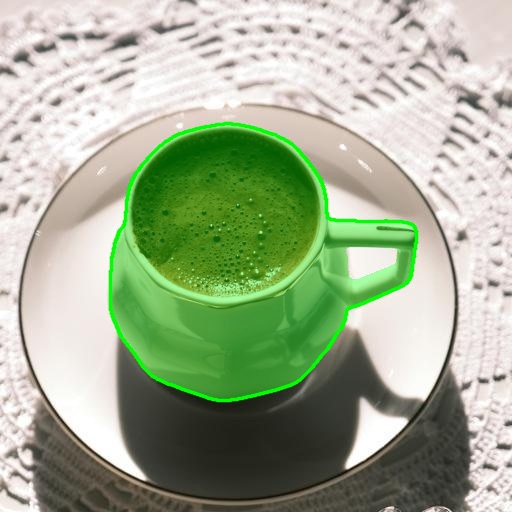} &
\includegraphics[width=0.089\textwidth]{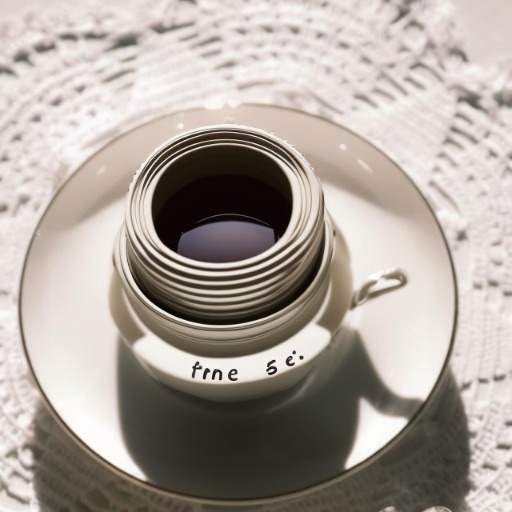} &
\includegraphics[width=0.089\textwidth]{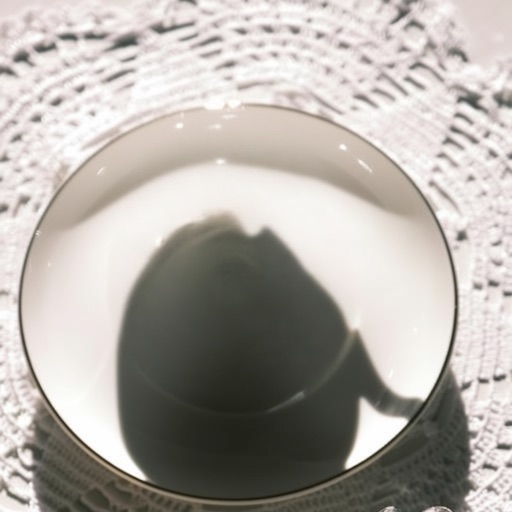} &
\includegraphics[width=0.089\textwidth]{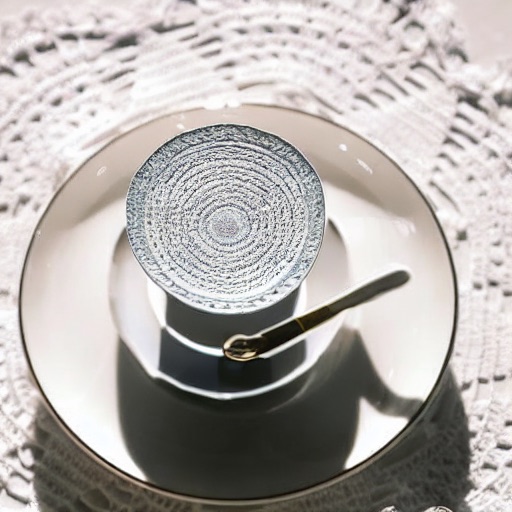} &
\includegraphics[width=0.089\textwidth]{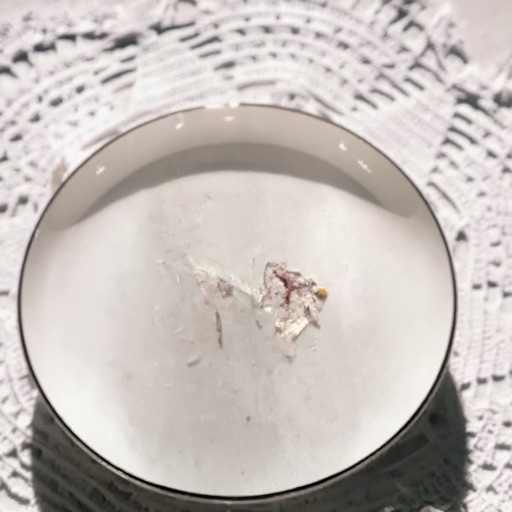} &
\includegraphics[width=0.089\textwidth]{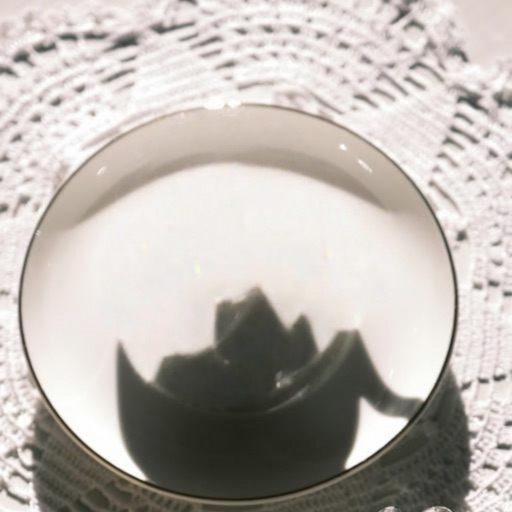} &
\includegraphics[width=0.089\textwidth]{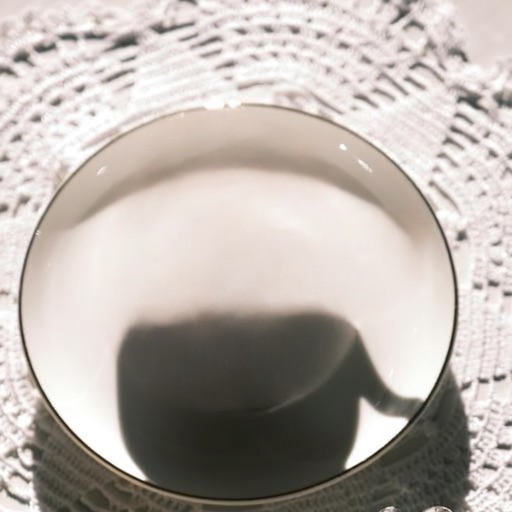} &
\includegraphics[width=0.089\textwidth]{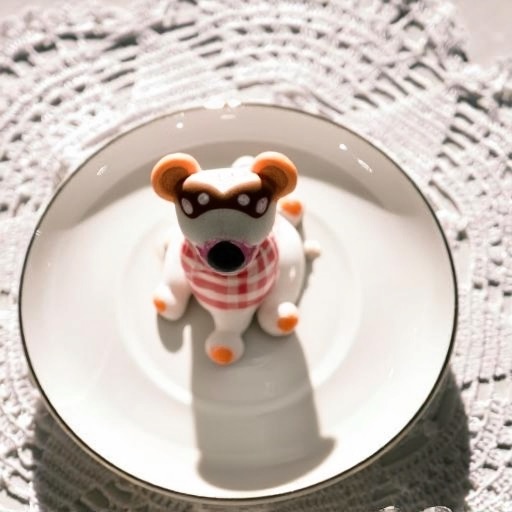} &
\includegraphics[width=0.089\textwidth]{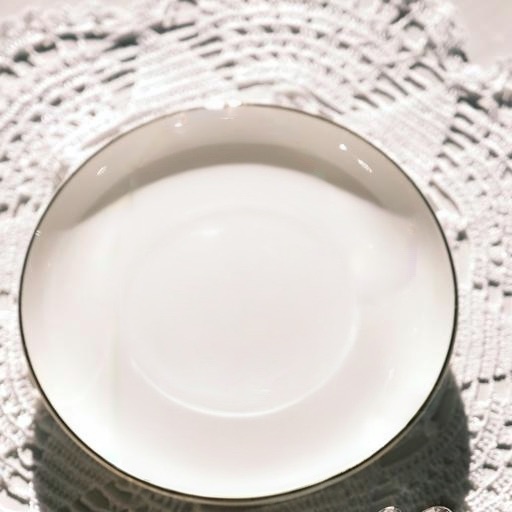} &
\includegraphics[width=0.089\textwidth]{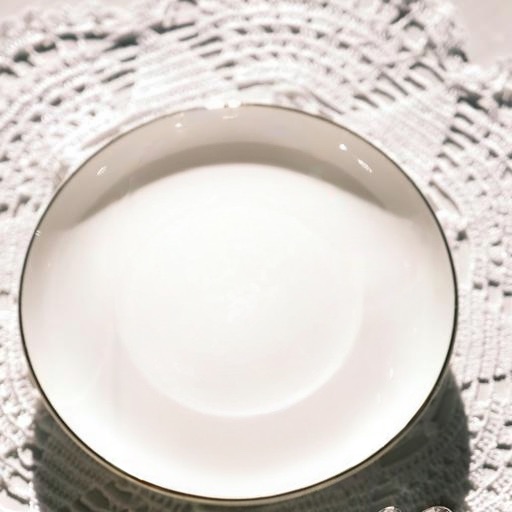} &
\includegraphics[width=0.089\textwidth]{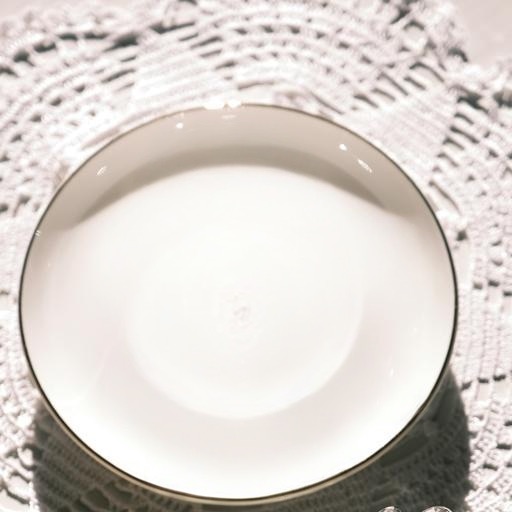} \\
\suppquallabels
\end{tabular}
\caption{Additional qualitative comparisons on eight OBER-Wild samples without ground-truth targets (part 1).}
\label{fig:more_results_a}
\end{figure*}

\begin{figure*}[t]
\centering
\setlength{\tabcolsep}{0pt}
\renewcommand{\arraystretch}{1.0}
\begin{tabular}{ccccccccccc}
\includegraphics[width=0.089\textwidth]{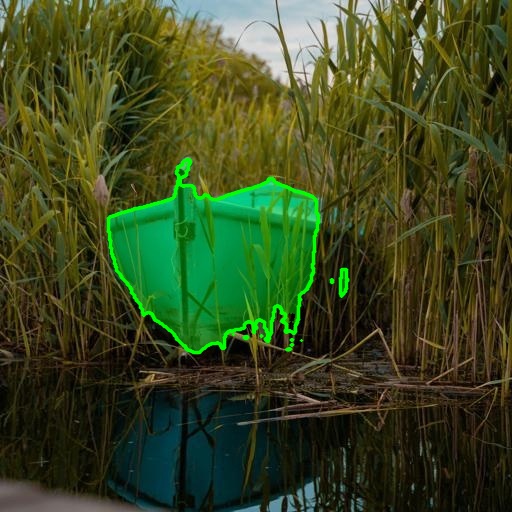} &
\includegraphics[width=0.089\textwidth]{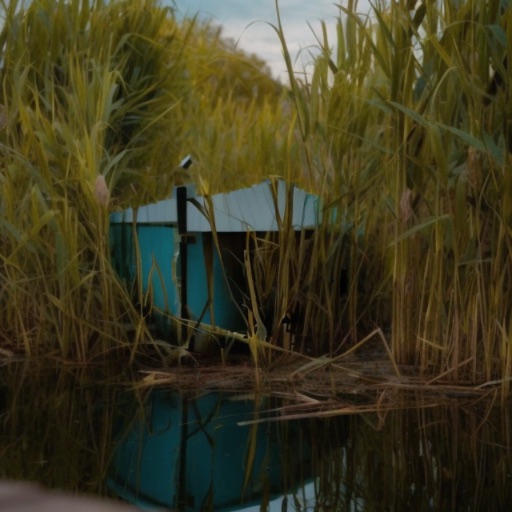} &
\includegraphics[width=0.089\textwidth]{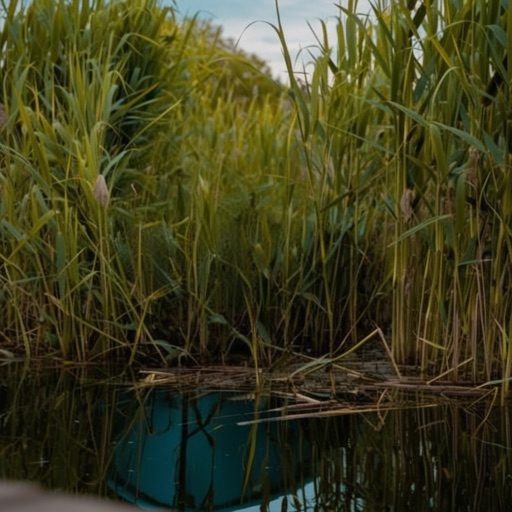} &
\includegraphics[width=0.089\textwidth]{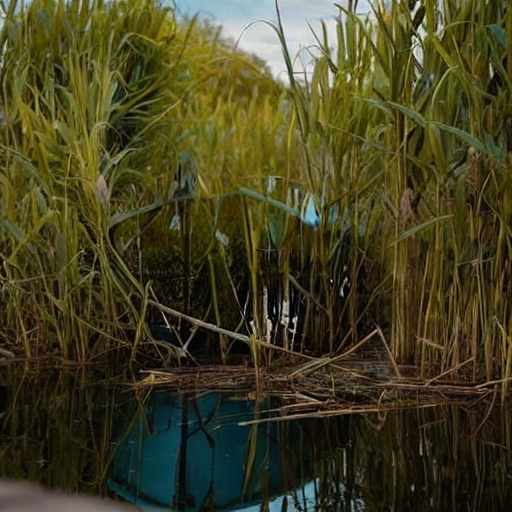} &
\includegraphics[width=0.089\textwidth]{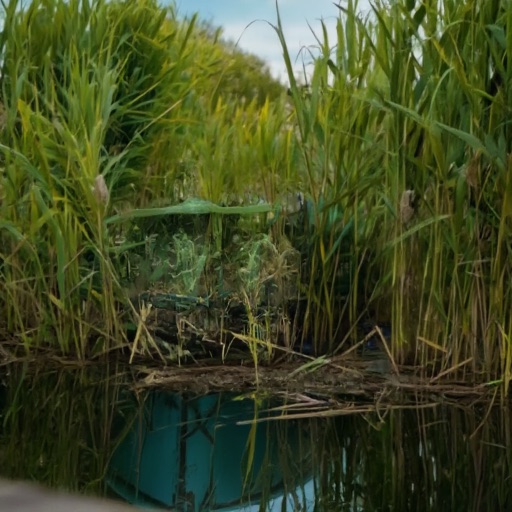} &
\includegraphics[width=0.089\textwidth]{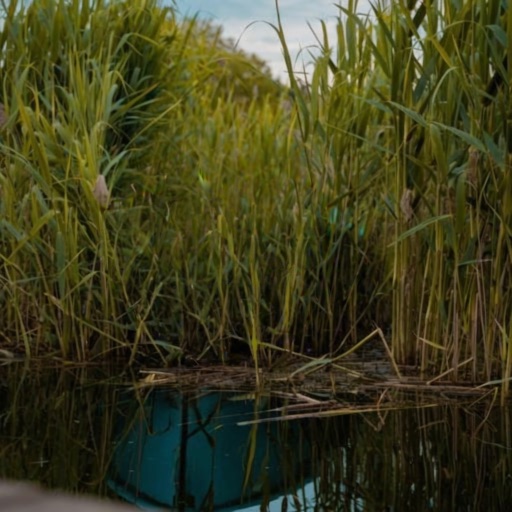} &
\includegraphics[width=0.089\textwidth]{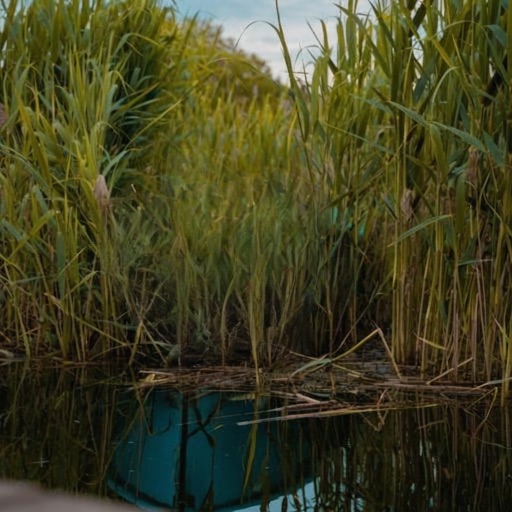} &
\includegraphics[width=0.089\textwidth]{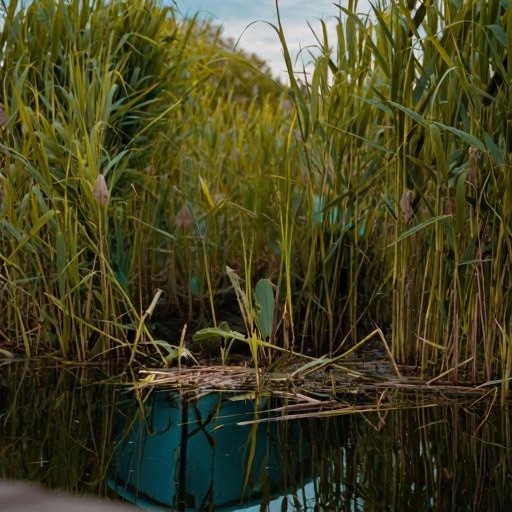} &
\includegraphics[width=0.089\textwidth]{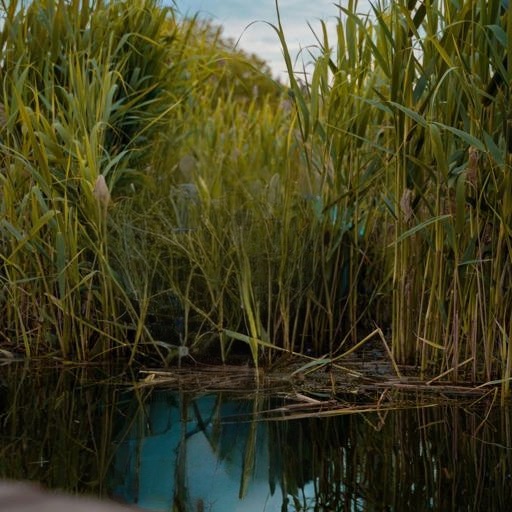} &
\includegraphics[width=0.089\textwidth]{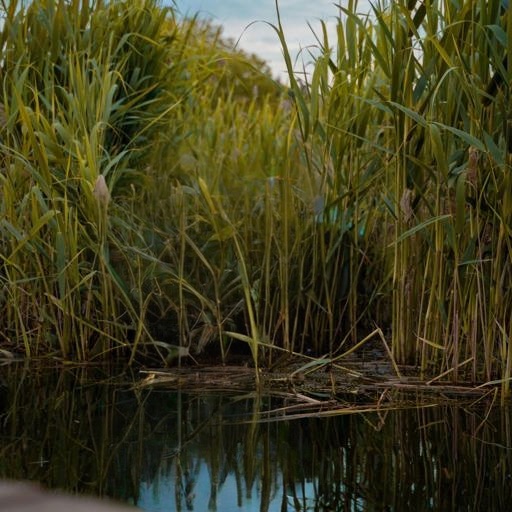} &
\includegraphics[width=0.089\textwidth]{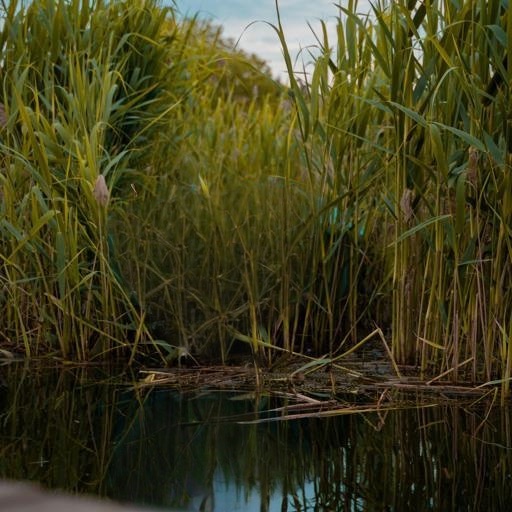} \\
\includegraphics[width=0.089\textwidth]{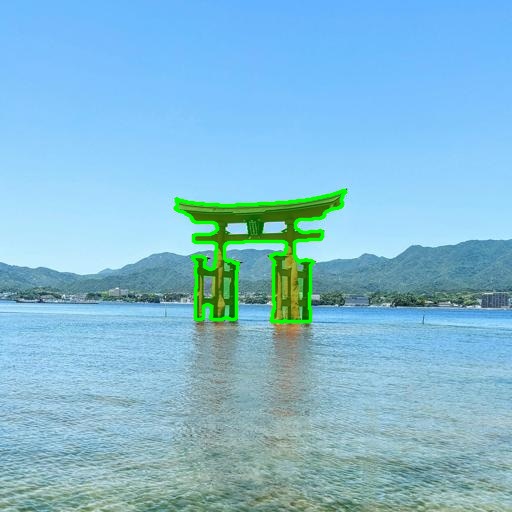} &
\includegraphics[width=0.089\textwidth]{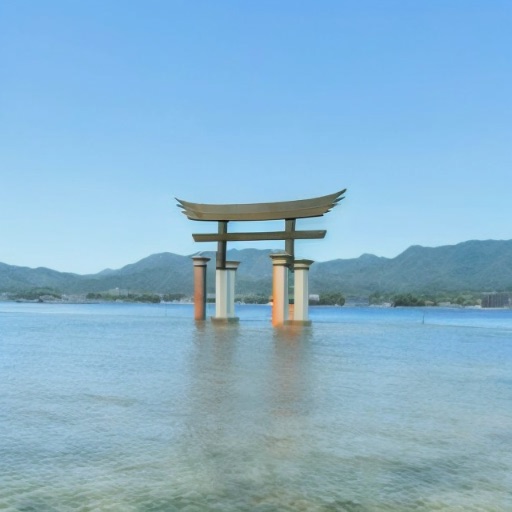} &
\includegraphics[width=0.089\textwidth]{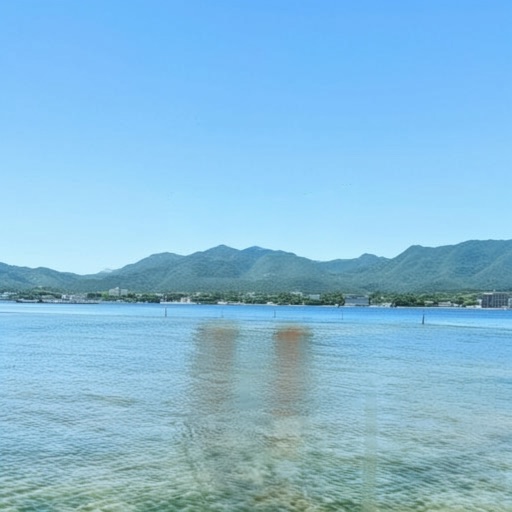} &
\includegraphics[width=0.089\textwidth]{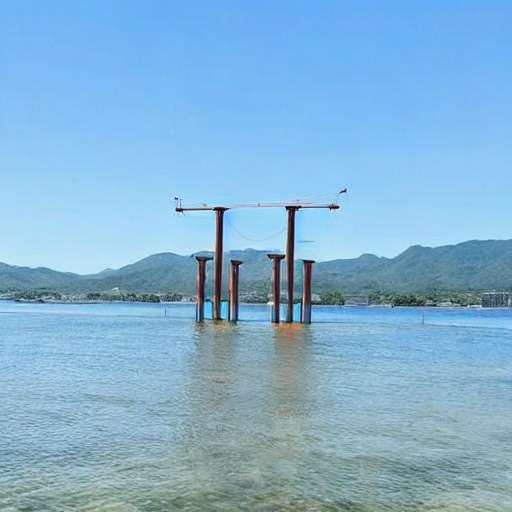} &
\includegraphics[width=0.089\textwidth]{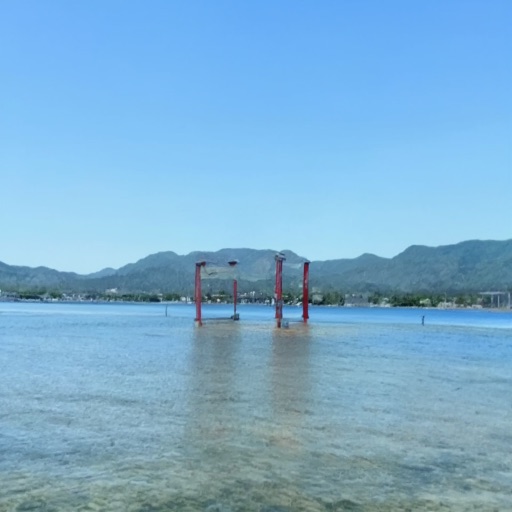} &
\includegraphics[width=0.089\textwidth]{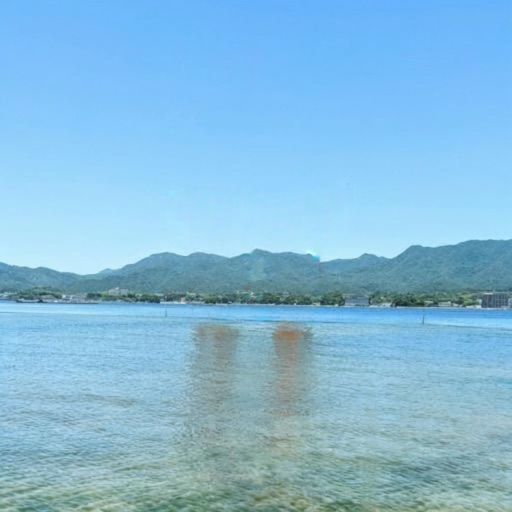} &
\includegraphics[width=0.089\textwidth]{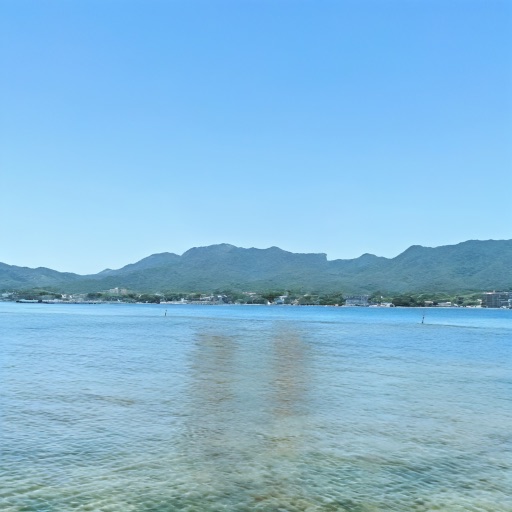} &
\includegraphics[width=0.089\textwidth]{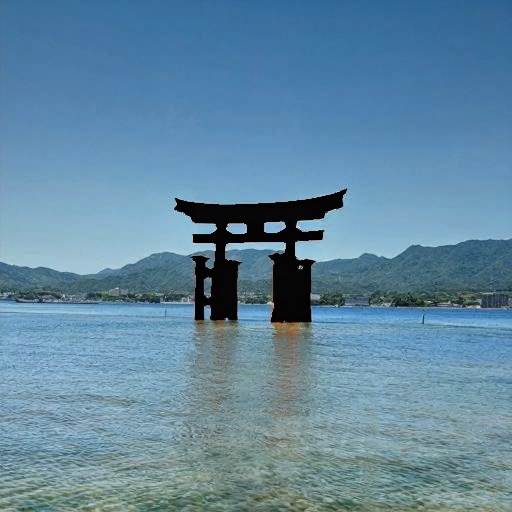} &
\includegraphics[width=0.089\textwidth]{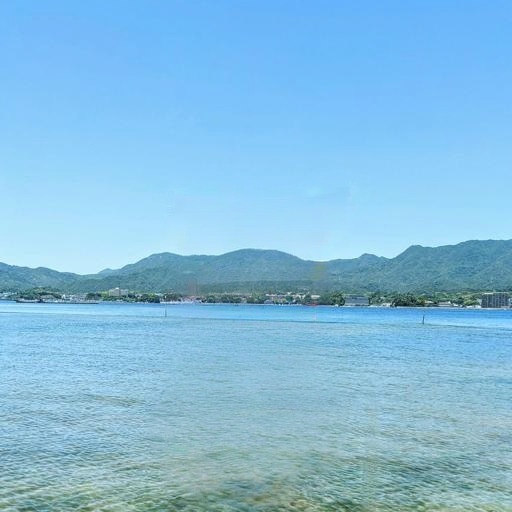} &
\includegraphics[width=0.089\textwidth]{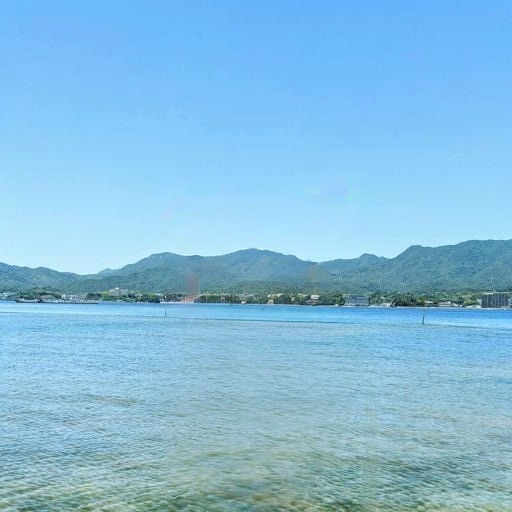} &
\includegraphics[width=0.089\textwidth]{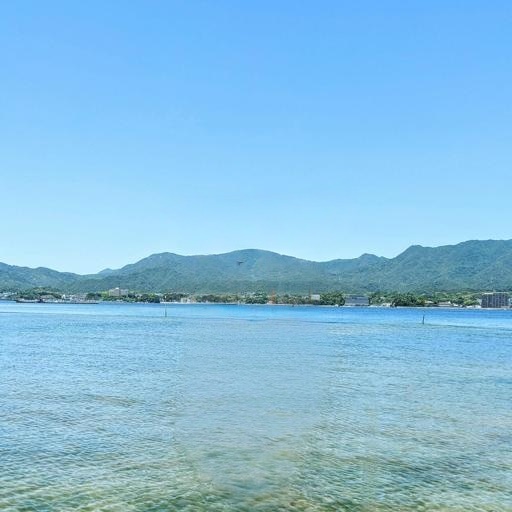} \\
\includegraphics[width=0.089\textwidth]{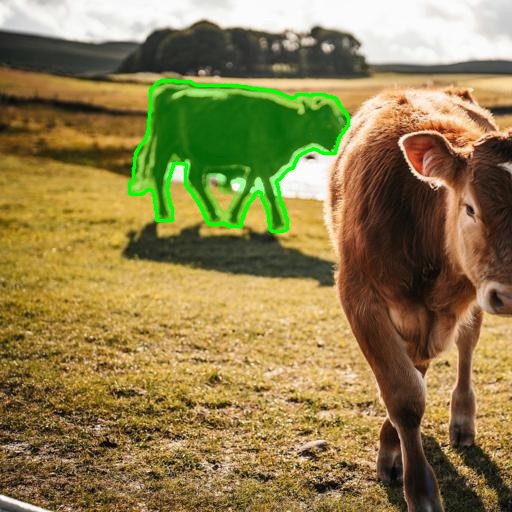} &
\includegraphics[width=0.089\textwidth]{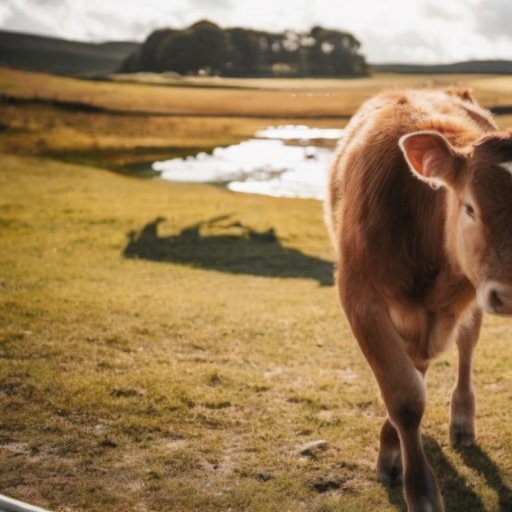} &
\includegraphics[width=0.089\textwidth]{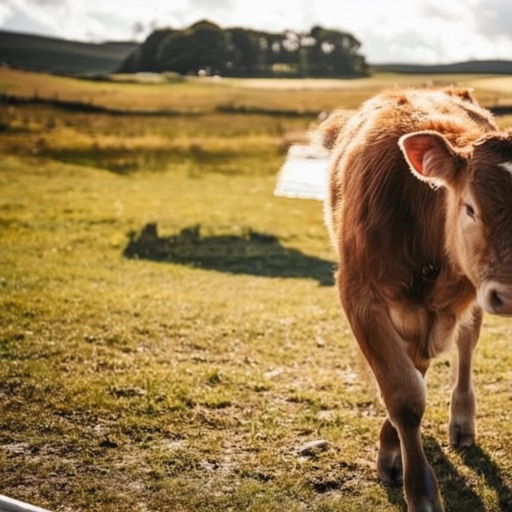} &
\includegraphics[width=0.089\textwidth]{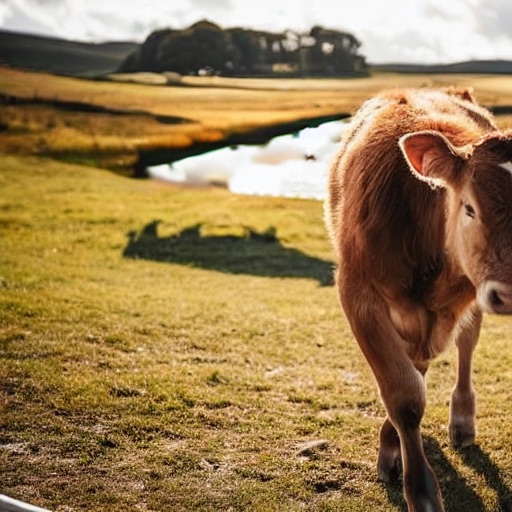} &
\includegraphics[width=0.089\textwidth]{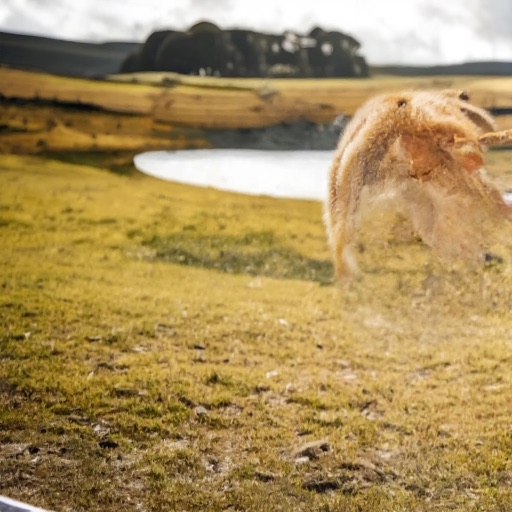} &
\includegraphics[width=0.089\textwidth]{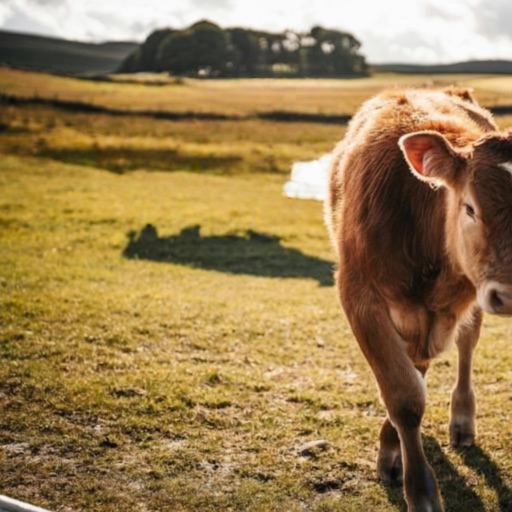} &
\includegraphics[width=0.089\textwidth]{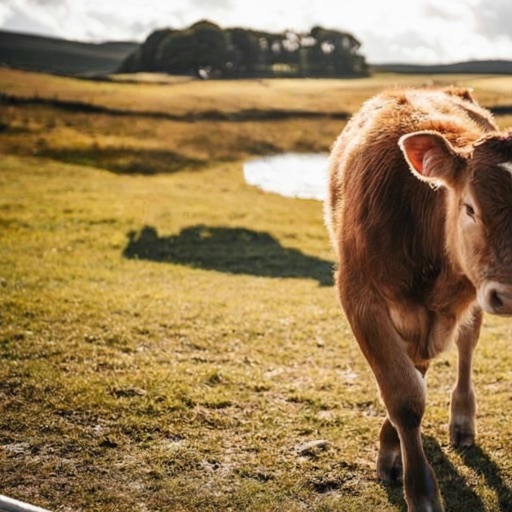} &
\includegraphics[width=0.089\textwidth]{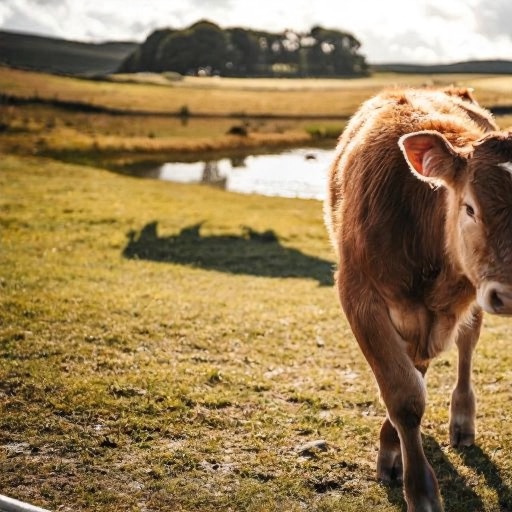} &
\includegraphics[width=0.089\textwidth]{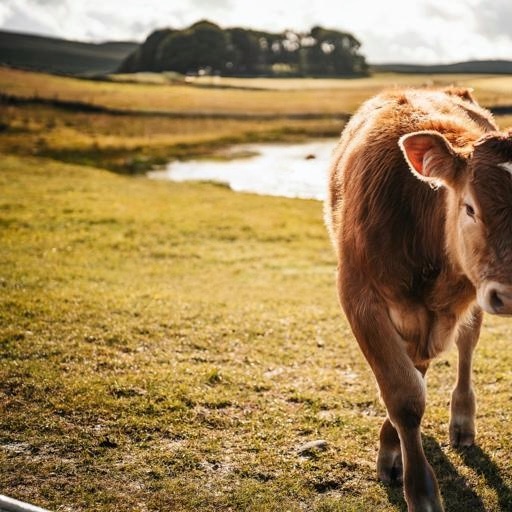} &
\includegraphics[width=0.089\textwidth]{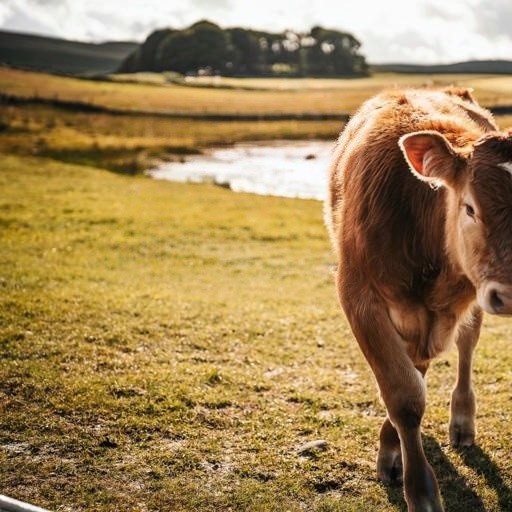} &
\includegraphics[width=0.089\textwidth]{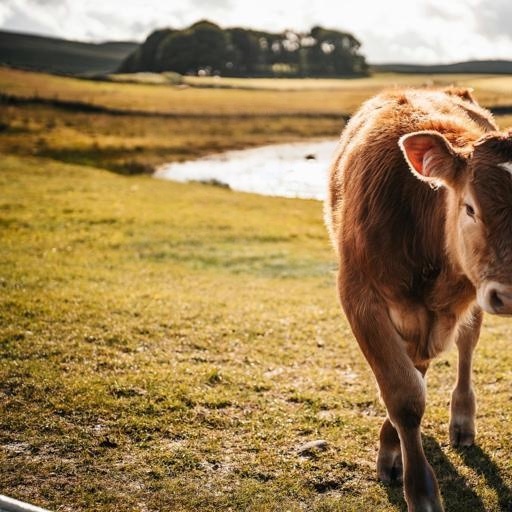} \\
\includegraphics[width=0.089\textwidth]{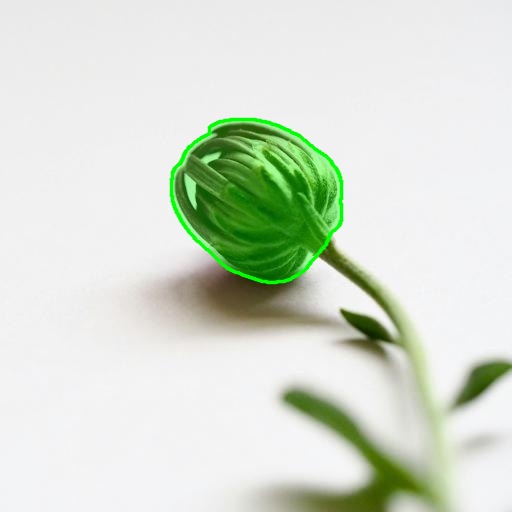} &
\includegraphics[width=0.089\textwidth]{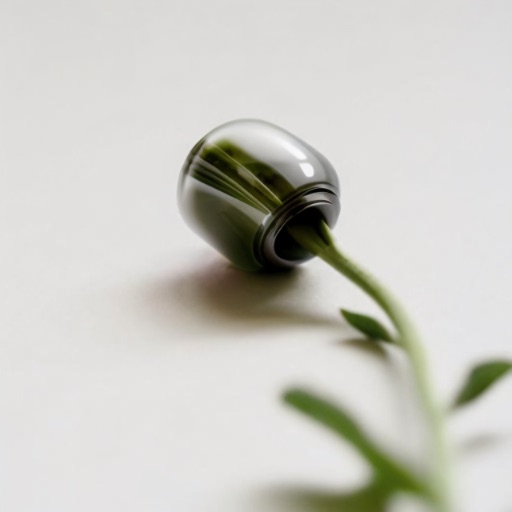} &
\includegraphics[width=0.089\textwidth]{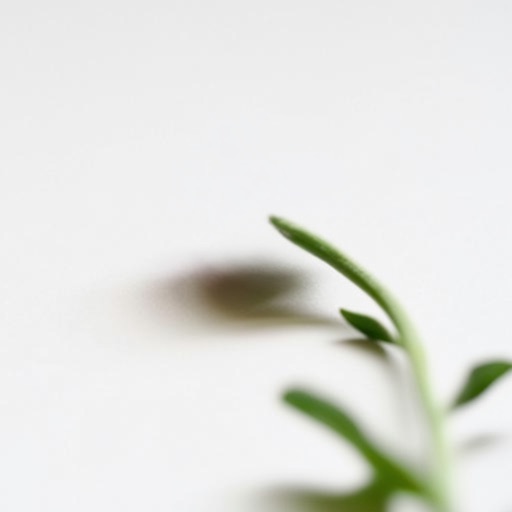} &
\includegraphics[width=0.089\textwidth]{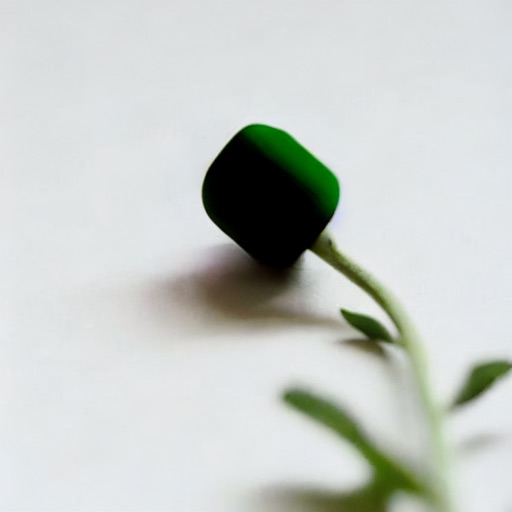} &
\includegraphics[width=0.089\textwidth]{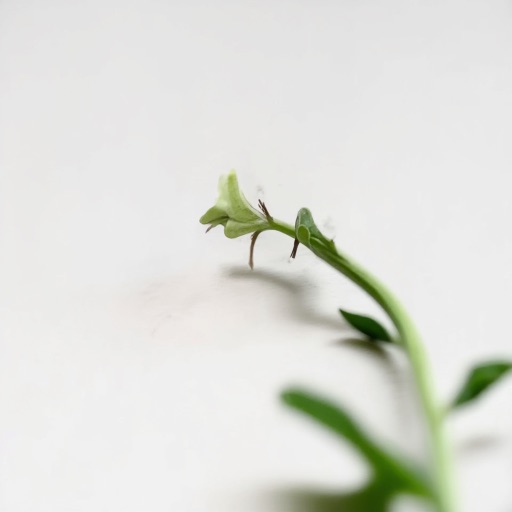} &
\includegraphics[width=0.089\textwidth]{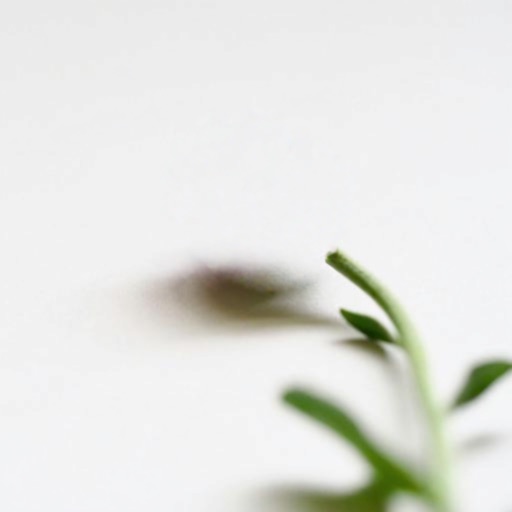} &
\includegraphics[width=0.089\textwidth]{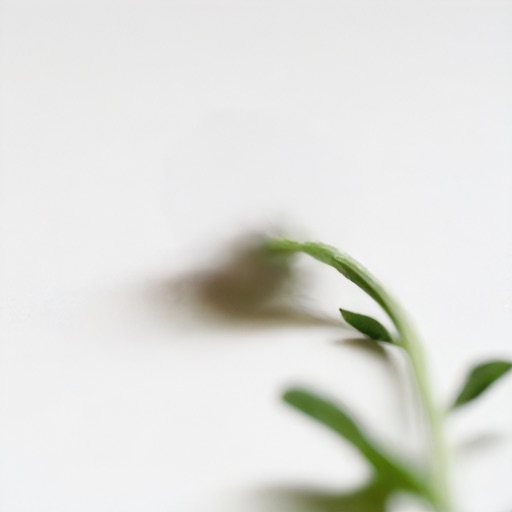} &
\includegraphics[width=0.089\textwidth]{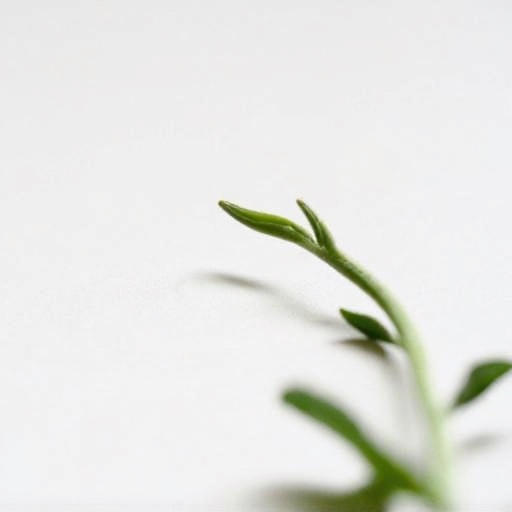} &
\includegraphics[width=0.089\textwidth]{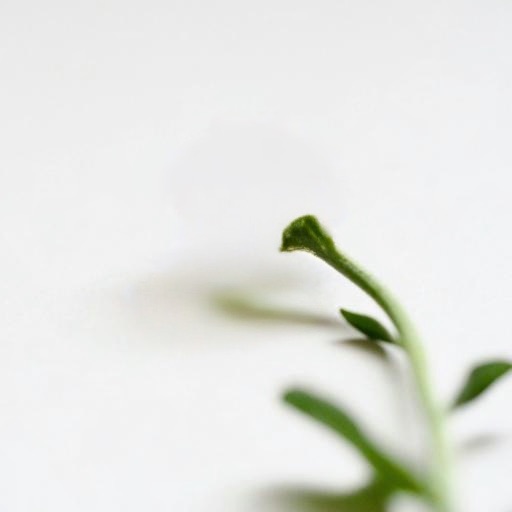} &
\includegraphics[width=0.089\textwidth]{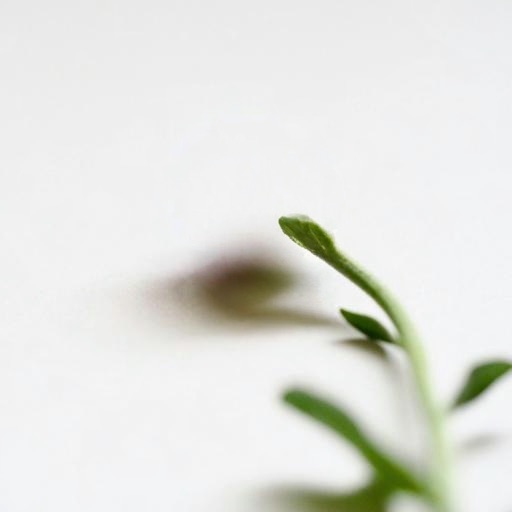} &
\includegraphics[width=0.089\textwidth]{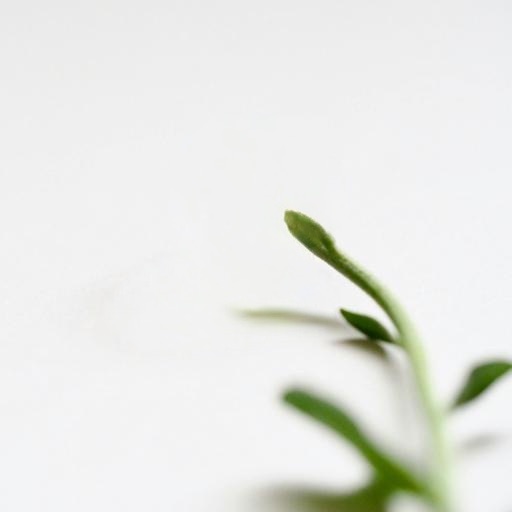} \\
\includegraphics[width=0.089\textwidth]{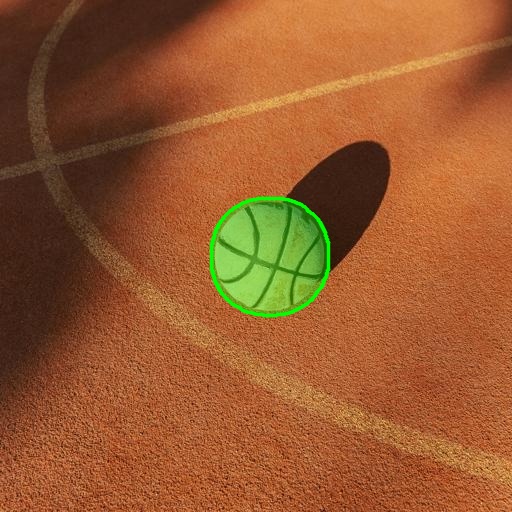} &
\includegraphics[width=0.089\textwidth]{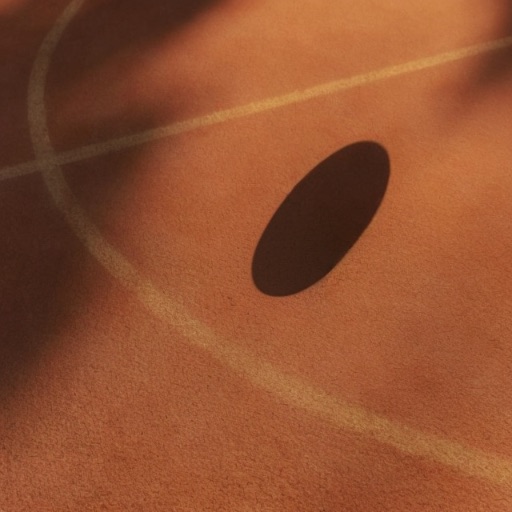} &
\includegraphics[width=0.089\textwidth]{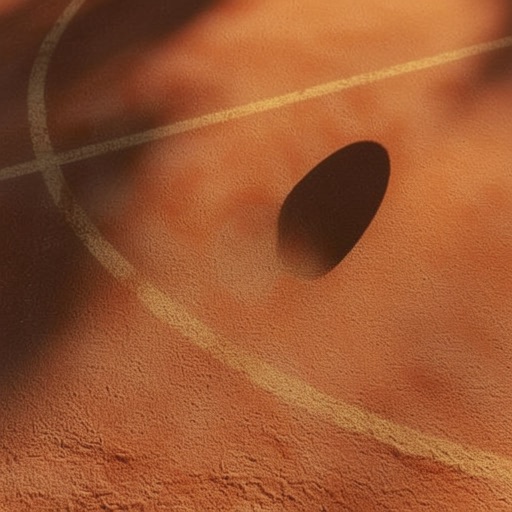} &
\includegraphics[width=0.089\textwidth]{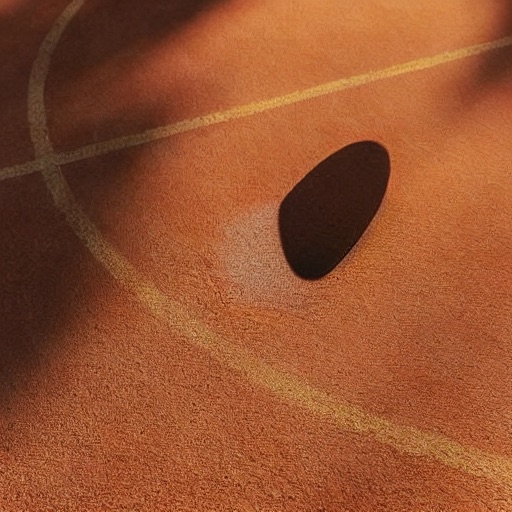} &
\includegraphics[width=0.089\textwidth]{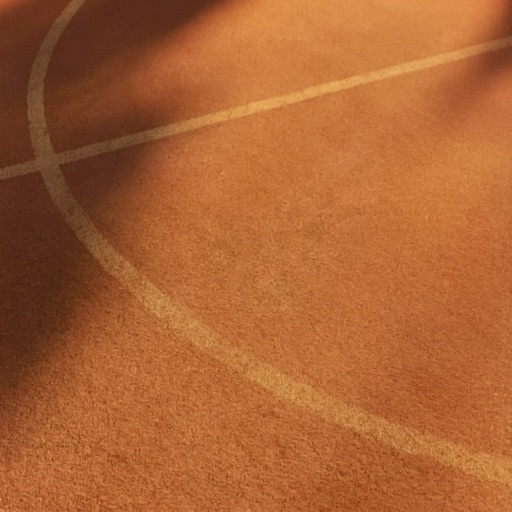} &
\includegraphics[width=0.089\textwidth]{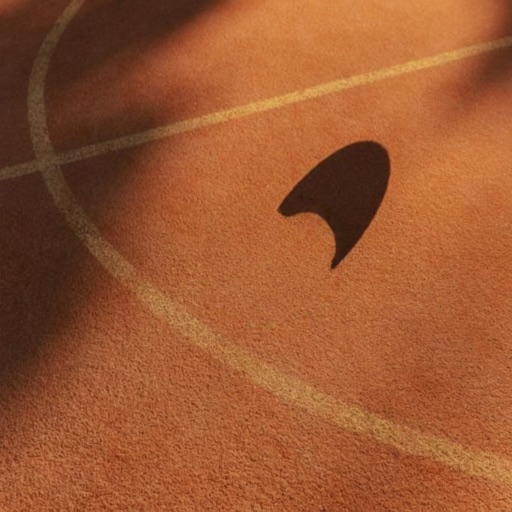} &
\includegraphics[width=0.089\textwidth]{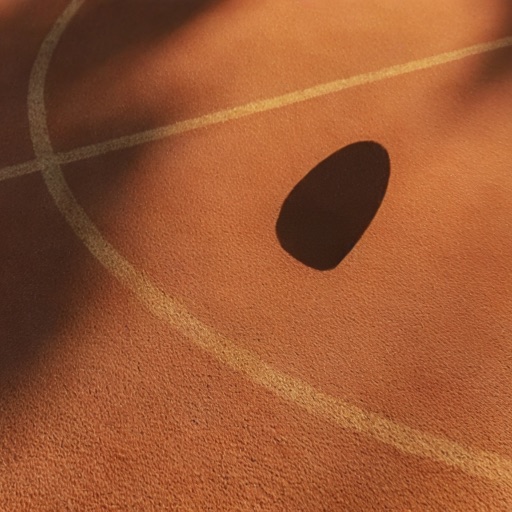} &
\includegraphics[width=0.089\textwidth]{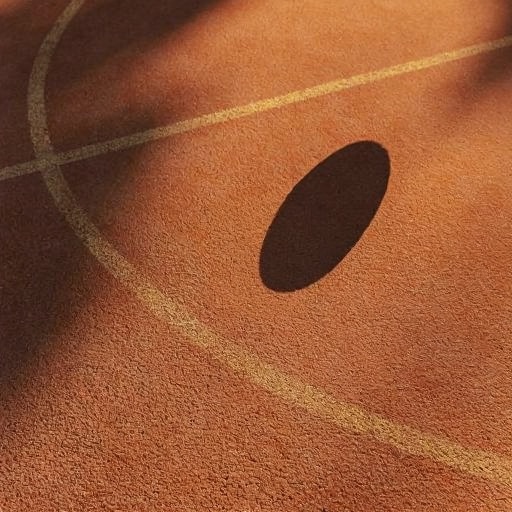} &
\includegraphics[width=0.089\textwidth]{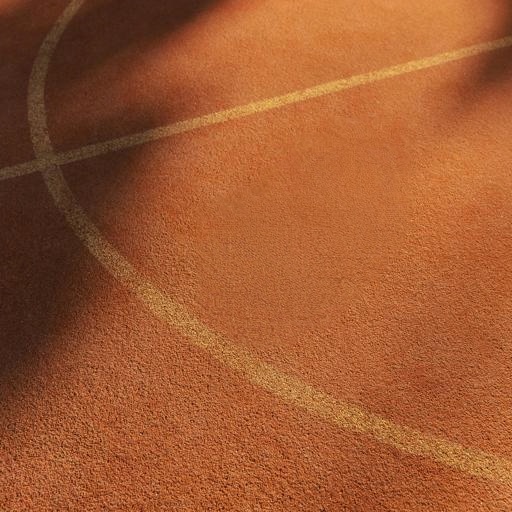} &
\includegraphics[width=0.089\textwidth]{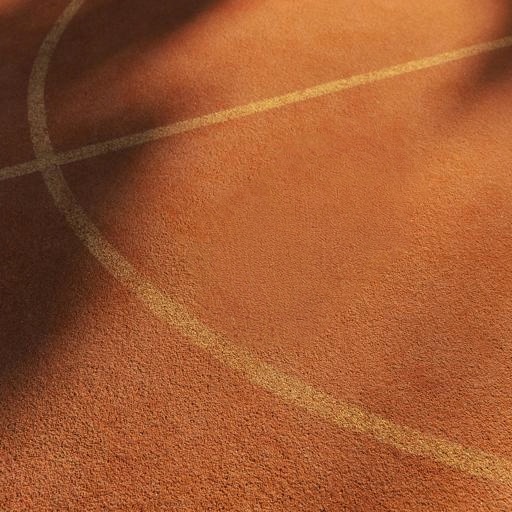} &
\includegraphics[width=0.089\textwidth]{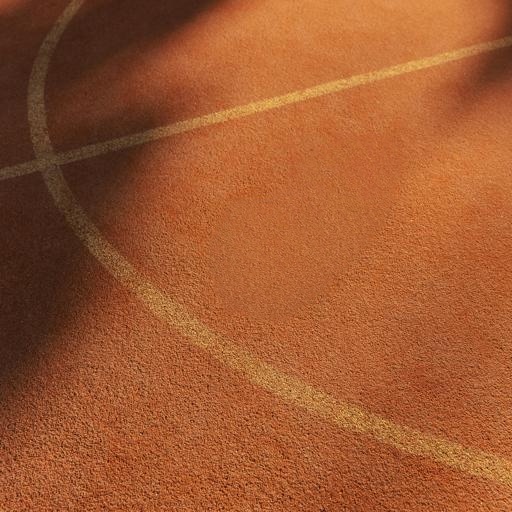} \\
\includegraphics[width=0.089\textwidth]{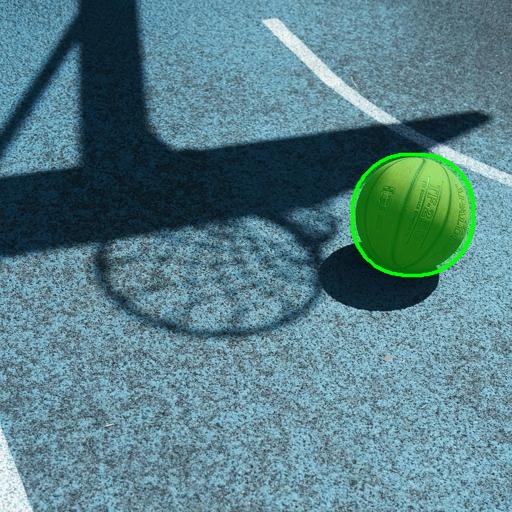} &
\includegraphics[width=0.089\textwidth]{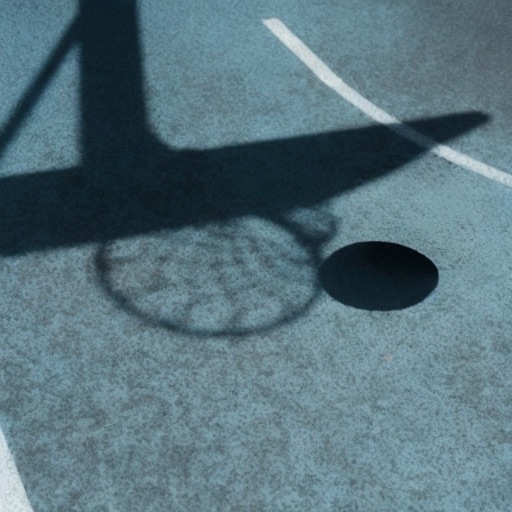} &
\includegraphics[width=0.089\textwidth]{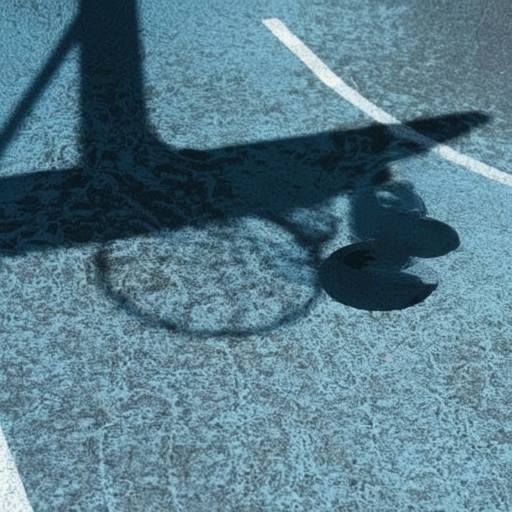} &
\includegraphics[width=0.089\textwidth]{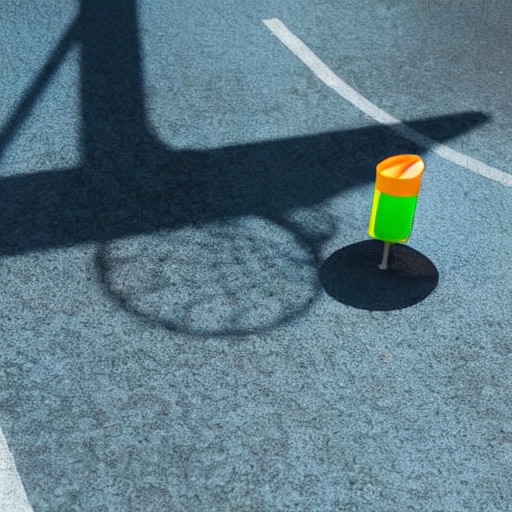} &
\includegraphics[width=0.089\textwidth]{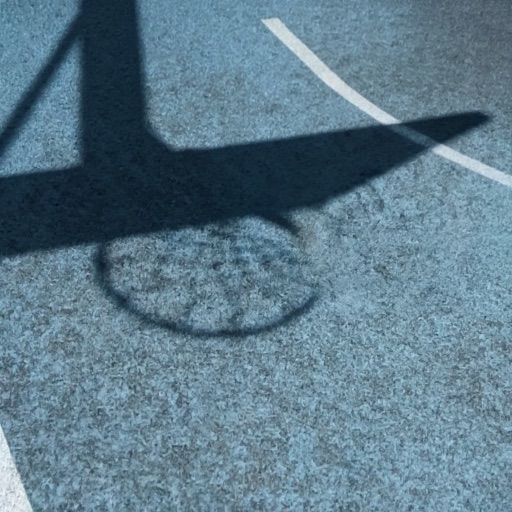} &
\includegraphics[width=0.089\textwidth]{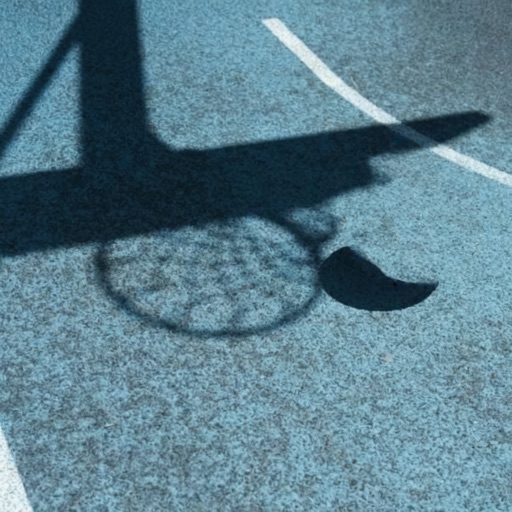} &
\includegraphics[width=0.089\textwidth]{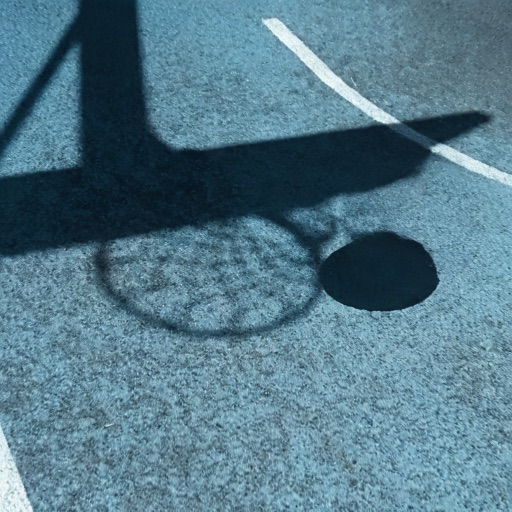} &
\includegraphics[width=0.089\textwidth]{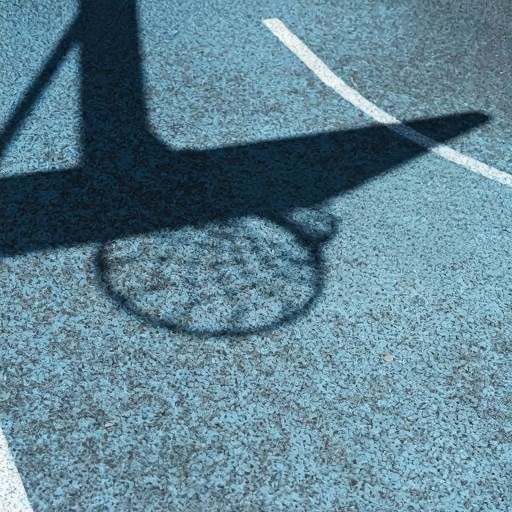} &
\includegraphics[width=0.089\textwidth]{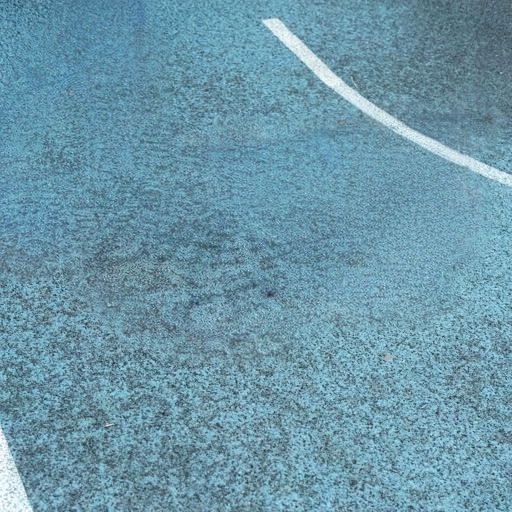} &
\includegraphics[width=0.089\textwidth]{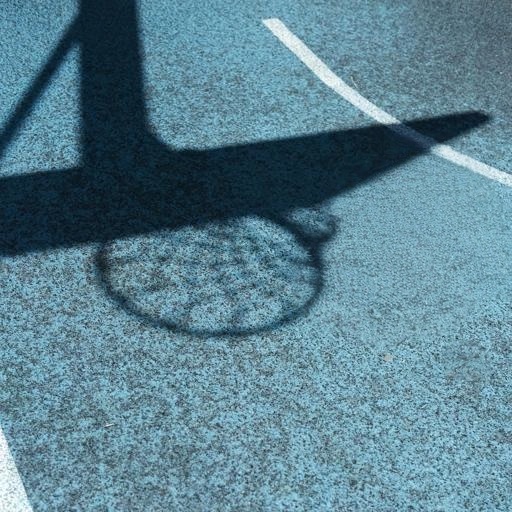} &
\includegraphics[width=0.089\textwidth]{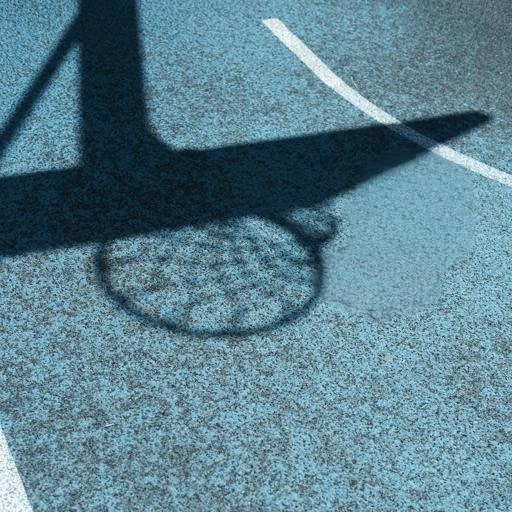} \\
\includegraphics[width=0.089\textwidth]{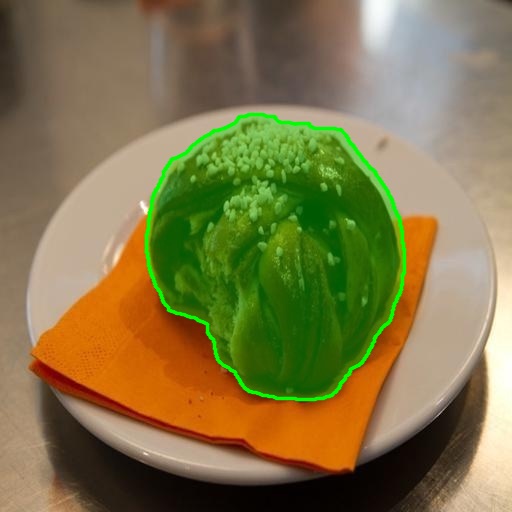} &
\includegraphics[width=0.089\textwidth]{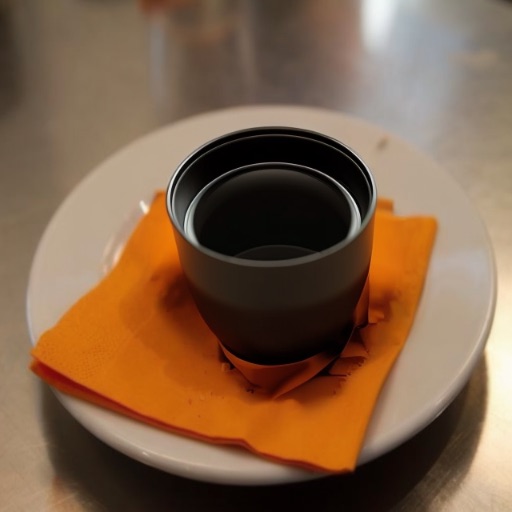} &
\includegraphics[width=0.089\textwidth]{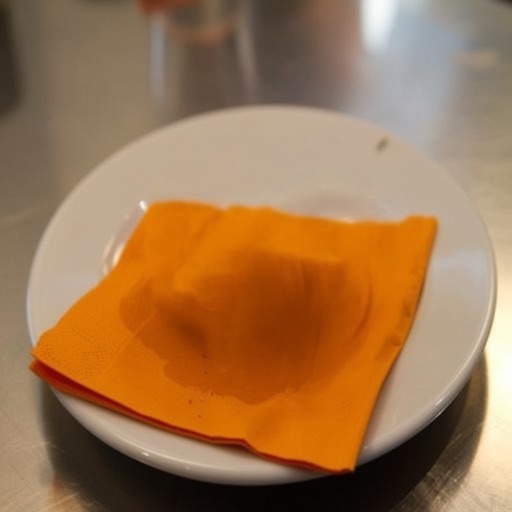} &
\includegraphics[width=0.089\textwidth]{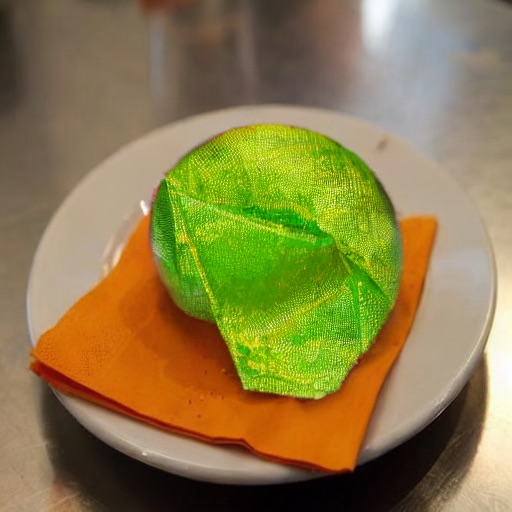} &
\includegraphics[width=0.089\textwidth]{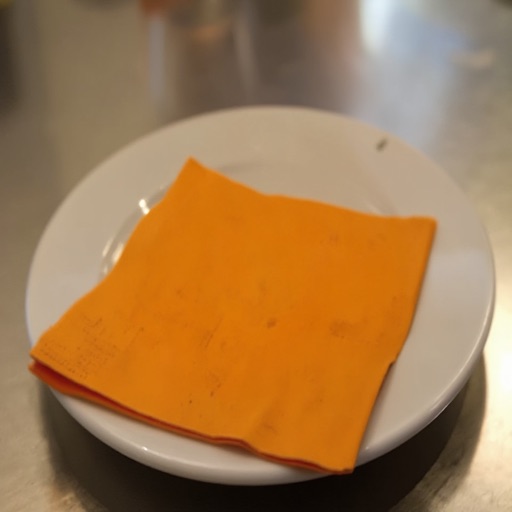} &
\includegraphics[width=0.089\textwidth]{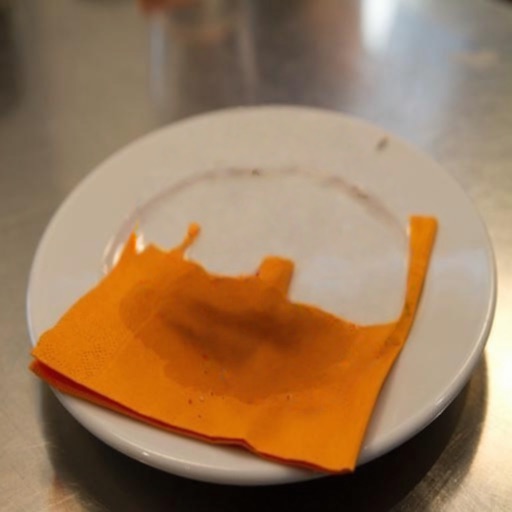} &
\includegraphics[width=0.089\textwidth]{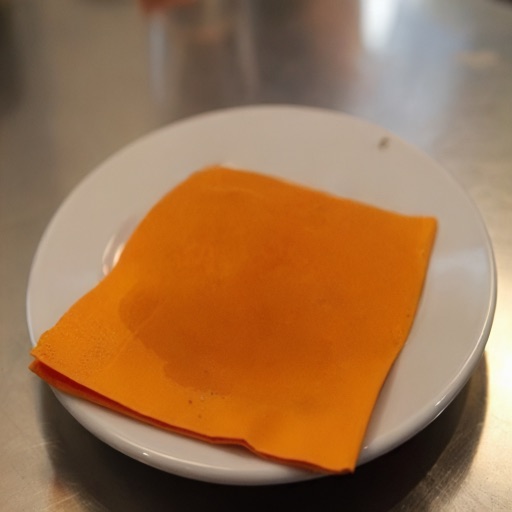} &
\includegraphics[width=0.089\textwidth]{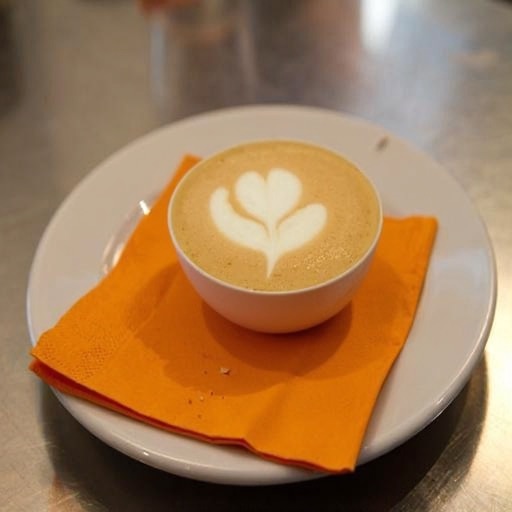} &
\includegraphics[width=0.089\textwidth]{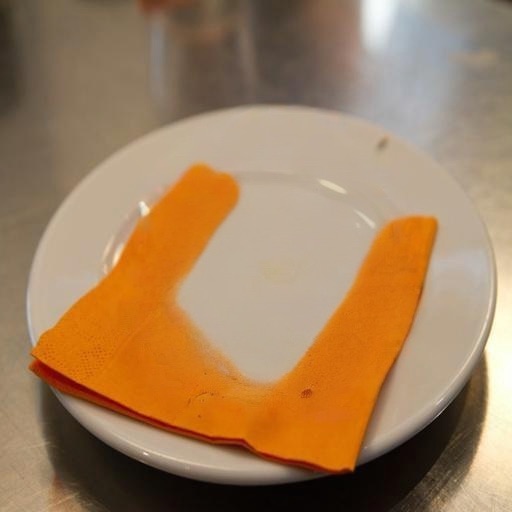} &
\includegraphics[width=0.089\textwidth]{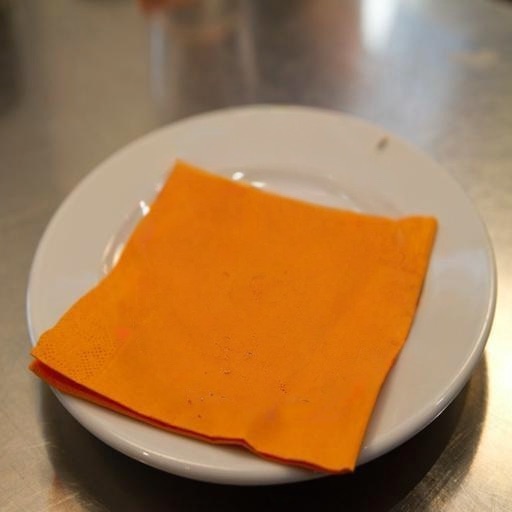} &
\includegraphics[width=0.089\textwidth]{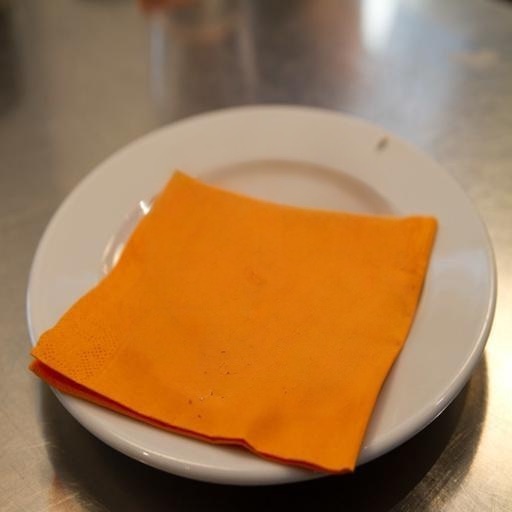} \\
\includegraphics[width=0.089\textwidth]{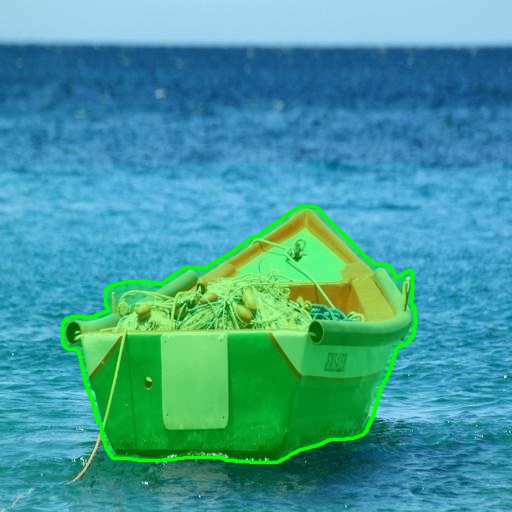} &
\includegraphics[width=0.089\textwidth]{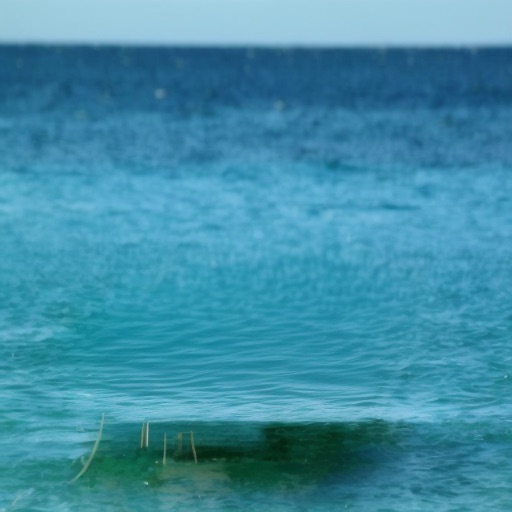} &
\includegraphics[width=0.089\textwidth]{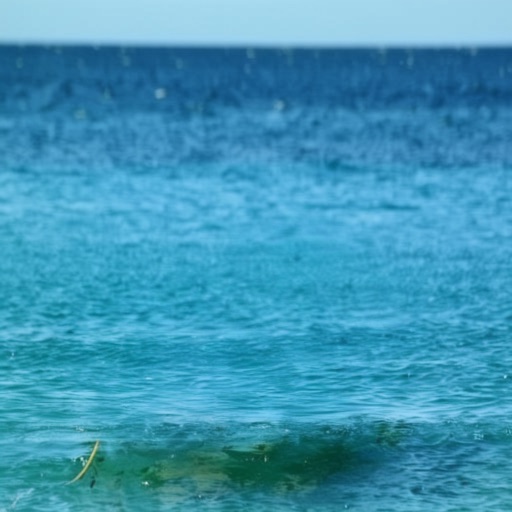} &
\includegraphics[width=0.089\textwidth]{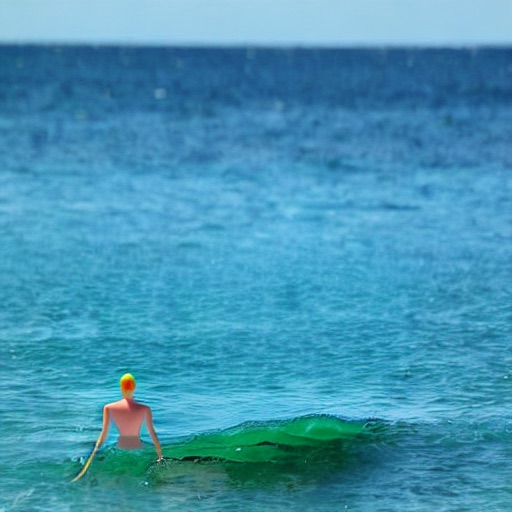} &
\includegraphics[width=0.089\textwidth]{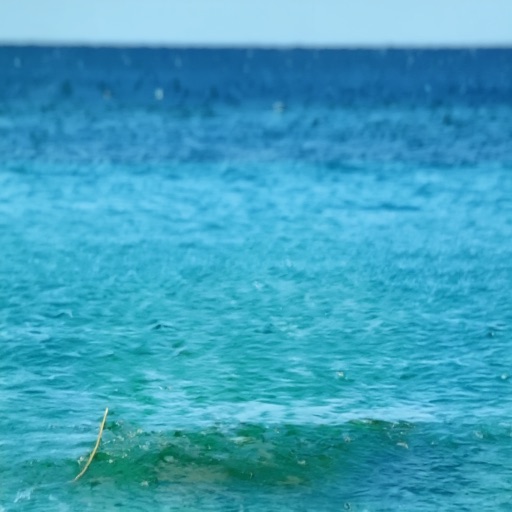} &
\includegraphics[width=0.089\textwidth]{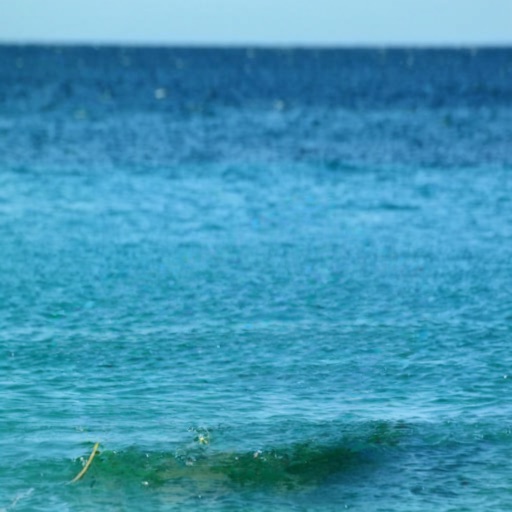} &
\includegraphics[width=0.089\textwidth]{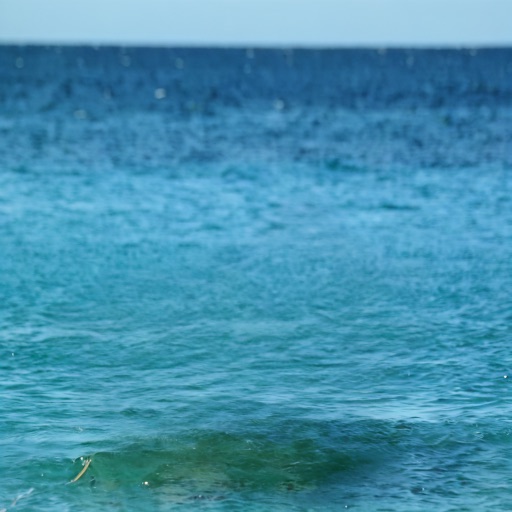} &
\includegraphics[width=0.089\textwidth]{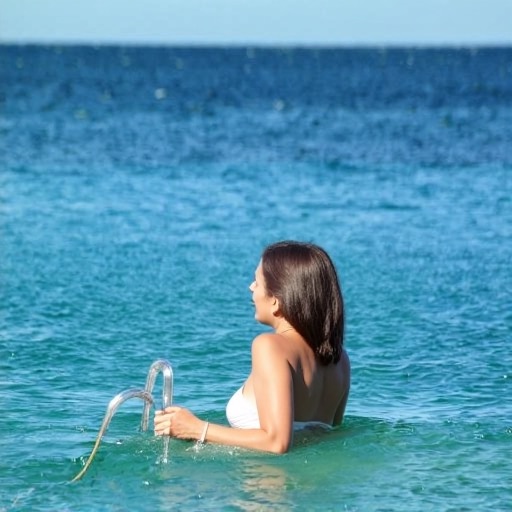} &
\includegraphics[width=0.089\textwidth]{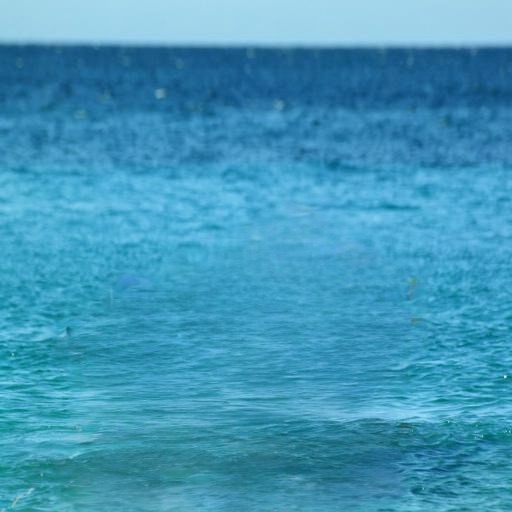} &
\includegraphics[width=0.089\textwidth]{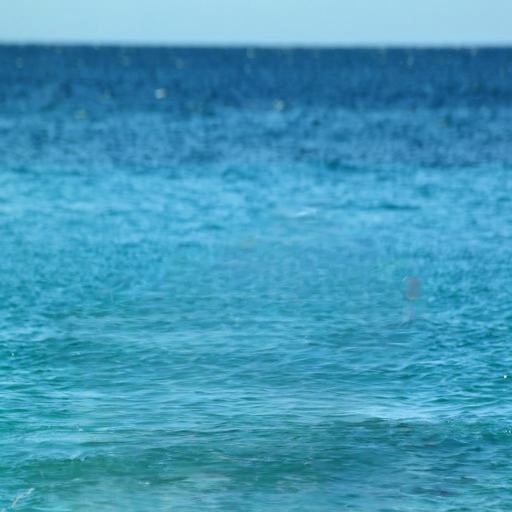} &
\includegraphics[width=0.089\textwidth]{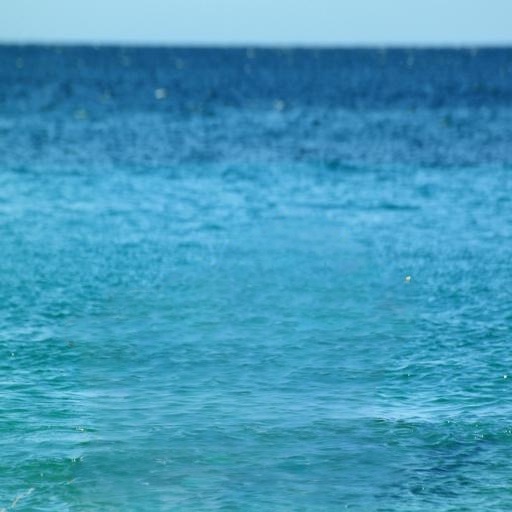} \\
\suppquallabels
\end{tabular}
\caption{Additional qualitative comparisons on eight OBER-Wild samples without ground-truth targets (part 2).}
\label{fig:more_results_b}
\end{figure*}

\section{Limitations and Future Work}

Despite its low denoising cost and latency, TurboClear is not yet lightweight in terms of memory footprint. Reducing the number of denoising steps decreases cumulative computation, but each inference still requires a full forward pass through the SDXL-based UNet and VAE, together with the LSF module. Under our single-A800 evaluation setup, the peak GPU memory consumption is 7,962 MiB. This requirement may limit deployment on memory-constrained consumer or edge devices. Model quantization, structured pruning, lightweight backbones, and memory-efficient inference are promising future directions.

TurboClear also relies on relatively strong supervision during training, including paired clean targets, object-effect masks, and a pretrained multi-step teacher. Although object-effect masks are not required at inference time, obtaining such annotations for new domains can be expensive. Reducing this dependency through weakly supervised or self-supervised region discovery would improve scalability.

Finally, TurboClear is designed for single-image object-effect removal and does not explicitly model temporal consistency. Applying it independently to video frames may lead to flickering or inconsistent background reconstruction, especially under large object or camera motion. Extending region-calibrated distillation and spatial fusion with temporal correspondence and cross-frame constraints is an important direction for future work.

\end{document}